\documentclass[a4paper,12pt,notitlepage]{report}
\usepackage[left=1in, right=1in, top=1in, bottom=1in,footskip=.25in]{geometry}
\usepackage{times}

\usepackage{multicol}
\usepackage[bookmarks=true,colorlinks=true,citecolor=blue,linkcolor=blue]{hyperref}
\usepackage{microtype}
\usepackage{amsmath}%
\usepackage{amssymb}%
\usepackage{graphics}%
\usepackage[tight]{subfigure}
\usepackage{tabularx}
\usepackage{booktabs}
\usepackage{multirow}
\usepackage{xspace} %
\usepackage{flushend}
\usepackage[ruled]{algorithm2e}
\usepackage{bm}
\usepackage{doi}
\usepackage{algorithmic}
\usepackage{wrapfig}
\usepackage[font=footnotesize]{caption}
\usepackage{setspace}

\usepackage{tikz}
\usetikzlibrary{calc}

\usepackage{acronym}
\acrodef{MDP}{Markov Decision Process}
\acrodef{POMDP}{Partially Observable Markov Decision Process}
\acrodef{DNN}{deep neural network}
\acrodef{CNN}{convolutional neural network}
\acrodef{RNN}{recursive neural network}
\acrodef{LSTM}{long short-term memory}
\acrodef{IRL}{inverse reinforcement learning}
\acrodef{IOC}{inverse optimal control}

\usepackage[backend=biber,natbib=true,style=alphabetic,citestyle=numeric,sorting=nty,maxbibnames=99,maxcitenames=1,mincitenames=1,bibstyle=ieee,dateabbrev=false]{biblatex}
\DefineBibliographyStrings{english}{
  andothers   = {et al\adddot}\addspace
}

\DeclareSortingTemplate{ymdt}{
  \sort{
    \field{presort}
  }
  \sort[final]{
    \field{sortkey}
  }
  \sort[direction=descending]{
    \field{sortyear}
    \field{year}
    \literal{9999}
  }
  \sort[direction=descending]{
    \field[padside=left,padwidth=2,padchar=0]{month}
    \literal{99}
  }
  \sort{
    \field{sorttitle}
  }
  \sort[direction=descending]{
    \field[padside=left,padwidth=4,padchar=0]{volume}
    \literal{9999}
  }
}

\usepackage{color}
\usepackage[textsize=scriptsize]{todonotes} %
\usepackage{url}
\usepackage{subfiles}

\begin{document}

\onehalfspacing

\pagenumbering{roman}

\definecolor{ttic_blue}{RGB}{55,93,137}
\begin{tikzpicture}[overlay,remember picture]
    \draw [line width=1mm] [ttic_blue]
        ($ (current page.north west) + (0.9in,-0.9in) $)
        rectangle
        ($ (current page.south east) + (-0.9in,0.9in) $);
\end{tikzpicture}

\begin{center}
\Large \MakeUppercase{\textbf{Neural Approaches to}}\\
\Large \MakeUppercase{\textbf{Co-Optimization in Robotics}}\\
\normalsize
\vspace{0.25in}
by\\Charles Schaff\\
\vspace{1in}
A thesis submitted\\
in partial fulfillment of the requirements for\\
the degree of\\
\vspace{0.3in}
Doctor of Philosophy in Computer Science\\
\vspace{0.3in}
at the\\
\vspace{0.3in}
\MakeUppercase{Toyota Technological Institute at Chicago}\\
Chicago, Illinois\\
\vspace{0.3in}
August, 2022\\
\vspace{1.0in}
Thesis Committee:\\
Matthew R.\ Walter (Thesis Advisor)\\
Ayan Chakrabarti\\
Audrey Sedal\\
David McAllester\\
\end{center}
\thispagestyle{empty} \clearpage

\chapter*{Acknowledgements}
\addcontentsline{toc}{chapter}{Acknowledgements}
The Ph.D. journey is filled with long periods of confusion, striking moments of clarity, and lots of hard work in between. Many people have helped me throughout this process and I am grateful for their efforts.

\vspace{1em}

First and foremost, I would like to thank my advisor Matthew R.\ Walter for your guidance and insight over these years.
You were knowledgeable about every project, dove head first into the details with me, and was patient as I learned.
The work in this thesis would not be possible without your efforts.
Thank you to Ayan Chakrabarti and Audrey Sedal. I learned a lot from your expertise and suggestions, and working with you was a lot of fun.
Thank you to Roman Garnett and Kilian Weinberger for introducing me to machine learning and guiding me through my first steps into research.
Thank you to my committee for generously providing your time and feedback.

\vspace{1em}

Thank you to all the TTIC students, faculty, and staff for creating a welcoming and fun place to learn and do research.
Thank you to David Yunis and Arthur MacKeith, working with you was always enjoyable.
And thank you to all of the robotics students for the fun MSI hackathons and research discussions, especially Andrea Daniele for maintaining the robots in the lab and providing near constant technical support.

\vspace{1em}

Thank you to the Salonica breakfast club for the laughs and fun over the years.

\vspace{1em}

Finally, thank you to my parents and the rest of my family. I am grateful for your love and support. \clearpage

\tableofcontents
\clearpage

\listoffigures
\clearpage

\listoftables
\clearpage

\begin{abstract}\addcontentsline{toc}{chapter}{\abstractname}
Robots and intelligent systems that sense or interact with the world are increasingly being used to automate a wide array of tasks.
The ability of these systems to complete these tasks depends on a large range of technologies such as the mechanical and electrical parts that make up the physical body of the robot and its sensors,
perception algorithms to perceive the environment, and planning and control algorithms to produce meaningful actions. 
These components have strong dependencies between them.
For example, robots will perform better when their bodies admit dynamics that are well suited for the control problems that they regularly encounter, and perception systems perform better with appropriate sensor design and placement.
Therefore, it is often necessary to consider the interactions between these components when designing an embodied system.

This thesis explores work on the task-driven co-optimization of robotics systems in an end-to-end manner, simultaneously optimizing the physical components of the system with inference or control algorithms directly for task performance.
Through the study of specific problems, such as beacon-based localization and legged locomotion, we develop a learning-based framework to co-optimize all aspects of robotics systems.
In this way, this thesis makes strides towards an efficient and automated approach to the design of robotics systems tailored to a specific application, which has the potential to both improve the performance of robotics systems and reduce the cost and barrier to entry of robot design.

We start by considering the problem of optimizing a beacon-based localization system directly for localization accuracy.
Beacon-based localization is a popular approach in environments where GPS is unavailable, such as underwater, underground, or indoors.
Designing such a system involves placing beacons throughout the environment and inferring location from sensor readings. 
The space of algorithms to automatically design these systems is relatively unexplored and past work often optimizes placement and inference separately.
In our work, we develop a deep learning approach to optimize both beacon placement and location inference directly for localization accuracy. In simulated experiments, our approach significantly outperforms strategies that consider beacon placement and location inference separately.

We then turn our attention to the related problem of task-driven optimization of robots and their controllers.
Approaches that automate the design of robots have a long history and include several techniques such as evolutionary algorithms, trajectory optimization, and nonlinear programming.
Reinforcement learning has proven successful at solving complex control problems but, at the start of our work, it was largely unexplored for co-optimization.
Therefore, we start by proposing a data-efficient algorithm based on multi-task reinforcement learning. Our approach efficiently optimizes both physical design and control parameters directly for task performance by leveraging a design-conditioned controller capable of generalizing over the space of physical designs.
We then follow this up with an extension to allow for the optimization over discrete morphological parameters such as the number and configuration of limbs.
Finally, we conclude by exploring the fabrication and deployment of optimized robots.
In this work we extend our previous algorithm to allow for the co-optimization of soft crawling robots, develop techniques for speeding up finite element simulations, and successfully fabricate and transfer the optimized robot from simulation to the real world. \end{abstract}
\clearpage

\pagenumbering{arabic}

\newcommand{\nlimb}{\textsc{N-Limb}\xspace}
\newcommand{\mS}{\mathcal{S}}
\newcommand{\mA}{\mathcal{A}}
\newcommand{\mP}{\mathcal{P}}
\newcommand{\mR}{\mathcal{R}}
\newcommand{\mE}{\mathcal{E}}
\newcommand{\mM}{\mathcal{M}}

\chapter{Introduction}

Intelligent machines that sense or interact with their environment are comprised of both physical components, such as cameras, lidars, joints, limbs, etc., and computational components, such as inference, perception, control, planners, etc.
The design and configuration of physical components define an interface between software and the environment, providing both the input (sensors) and output (actuators) by which computational components make inferences about or control the world around them.
Therefore, the success of inference or control is fundamentally linked with, and limited by, hardware design.
Inference problems are easier when sensors measure data containing information about the quantity being inferred and control problems are easier when physical components allow for stable and efficient motions for the task at hand.
Therefore, when designing intelligent systems, it is necessary to jointly consider the hardware of the system with its computational processes.

Traditionally, this design process has required experts with extensive domain knowledge.
For example, in order to design the gripper of a robot, one must consider the type of gripper (e.g., pinch, suction, humanoid, etc.), how to actuate it (e.g., joints or tendons), what types of sensors to use (e.g., vision or touch) and where to place them, and how all of these choices affect the grasping dynamics of the resulting system to enable control algorithms that efficiently and robustly solve the distribution of manipulation problems relevant to the specific application.
Naturally, the design of a gripper intended to perform pick-and-place tasks in an Amazon warehouse will be very different from a gripper designed to sort recycling, tighten bolts on an assembly line, or serve as a general purpose solution to a variety of tasks.
Adapting a manipulator to different tasks often requires redesigning the system for each specific application, which is costly and laborious.

The goal of this thesis is to develop co-optimization algorithms that automate and improve upon this costly and laborious process by encompassing all aspects of the design of embodied systems. Throughout this thesis, we will use the term \textit{co-optimization} to refer to optimization problems and algorithms that aim to optimize multiple components of an embodied system (e.g., physical parameters such as limb lengths along with control parameters). We use the term co-optimization in place of the more popular \textit{joint optimization} to avoid confusion when optimizing the physical joints of a robot.
The co-optimization algorithms we develop in this thesis aim to produce optimized systems by considering the physics of sensors, actuators, and the environment, fabrication constraints, and the distribution of scenarios relevant for a particular application.
If successful, these algorithms will improve the quality of embodied systems, reduce the barrier to entry to robot design, and allow designers to focus on \textit{what} they want the system to accomplish and not \textit{how} it should be accomplished.

\section{A Brief History of Co-optimization in Robotics}

\begin{figure}[t]
    \centering
    \subfigure[Designed examples of genotype graphs and corresponding creature morphologies.]{\includegraphics[width=0.45\textwidth]{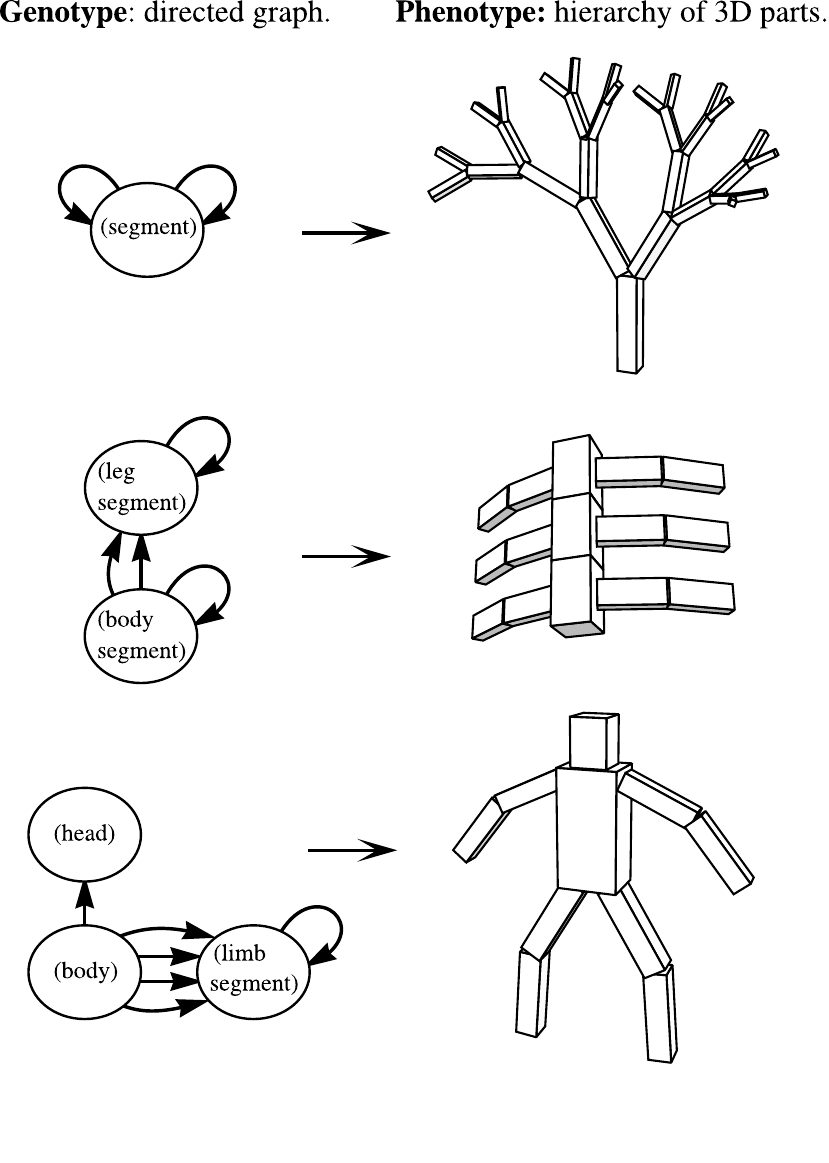}\label{fig:sims_genotype}}
    \subfigure[Evolved creatures for various tasks.]{\includegraphics[width=0.45\textwidth]{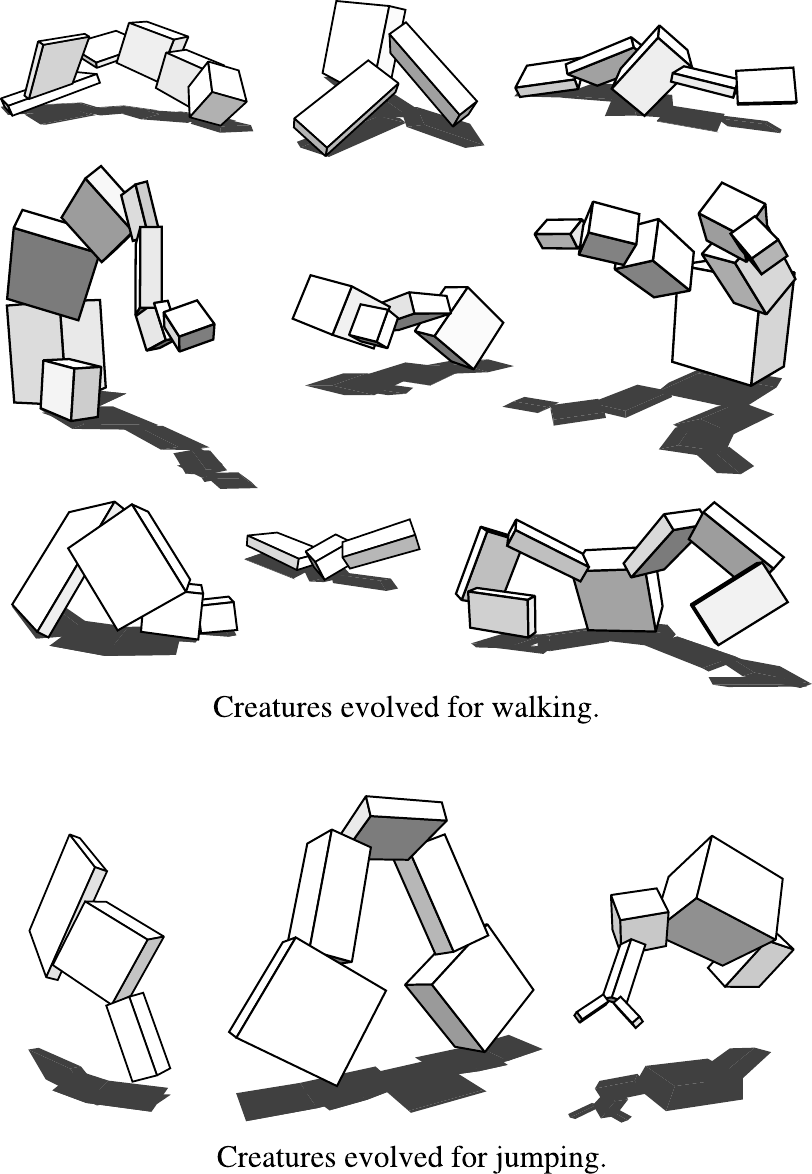}\label{fig:sims_results}}
    \caption{Karl Sims' seminal work on the evolution of virtual creatures~\cite{sims1994}. In this work, Sims utilized evolutionary algorithms to maximize the fitness of various creatures built from box-shaped parts. The morphology, sensors, and controls were all evolved together for simple tasks such as walking, jumping, and swimming.}
    \label{fig:sims}
\end{figure}

The goal of this section is to provide context for the work in this thesis, and provide a brief introduction to the major ideas that have shaped the field of co-optimization in robotics over the last thirty years.

\subsection{Early Approaches to Co-optimization}
Early approaches to co-optimization generally fall into two main classes: evolutionary algorithms and nonlinear programming.

Many approaches to the co-optimization of robots can be traced back to Karl Sims' seminal work in 1994~\cite{sims1994}. In this work (Figure \ref{fig:sims}), Sims described an evolutionary approach to optimizing morphology, sensor placement, and control inside several virtual environments. Their approach takes inspiration from evolutionary biology and maintains a population of agents, $\{x_i\}_i$, that are defined by a ``genotype" (represented as a directed graph), a collection of optical and contact sensors, and control parameters defining a sequence of operations that map sensor observations to actions.
At each iteration, the population is evaluated under a fitness function $f(x)$. The most fit agents in the population are then crossed and mutated to form the population at the next iteration.
Sims found that under this approach, complex morphologies and behaviors automatically emerged to solve the given tasks. Figure~\ref{fig:sims_results} shows examples of optimized designs from their work.

\begin{wrapfigure}{r}{0.5\textwidth}
    \centering
    \includegraphics[width=0.5\textwidth]{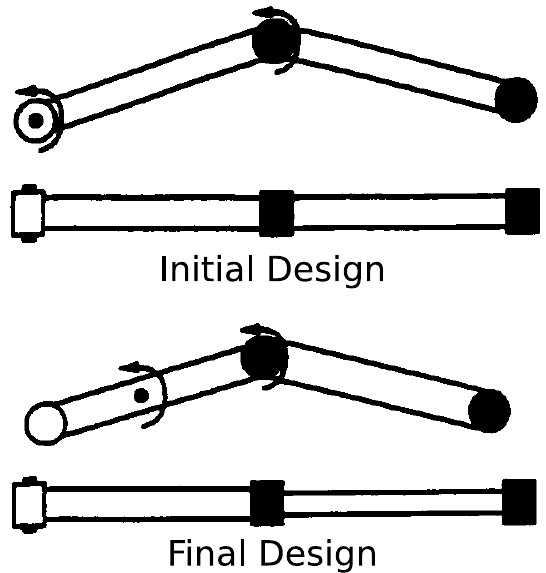}
    \caption{The optimization of a two-bar linkage system by \citet{park94} for high speed positioning.}
    \label{fig:park}
\end{wrapfigure}

Around the same time as the work by~\citet{sims1994}, \citet{park94} proposed an alternative approach to co-optimization based on nonlinear programming. In their work, they optimized a two-bar linkage system together with a PD controller for high speed positioning (e.g., for high throughput pick and place tasks). Given the simple closed-form dynamics of the system, the authors were able to derive near-optimal PD gains for each proposed arm, as well as a closed-form objective related to the settling time of the arm and controller. Then, the authors carried out the optimization with a form of projected gradient descent to minimize the settling time of specific motions while maintaining fabrication constraints on the arm's inertia, workspace, and vertical deflections due to gravity. The end result of the optimization was an arm able to move twice as quickly without sacrificing control accuracy.

These two approaches present a trade-off in the efficiency and generality of co-optimization. Evolutionary approaches provide a general approach that makes no assumptions about dynamics of the system or the objective function. This allows evolutionary methods to be applied with little to no modification to a large set of problems. However, this comes at the cost of low sample efficiency and often poor optimization of the objective function. On the other hand, methods based on nonlinear programming techniques can efficiently and effectively optimize the objective, but require explicit knowledge of how the dynamics of the system relate to the physical parameters. This often constrains these approaches to problems with simple differentiable dynamics, which can make it challenging to incorporate complex phenomena such as contact with the environment.

\subsection{Evolutionary Approaches to Co-optimization}

Both classes of methods continued to develop largely independent of each other. However, while there are several successful works that use nonlinear programming~\cite{li2001design, ravichandran06, allison2014special, pil1996integrated, reyer02, yan2009integrated}, the field of ``Evolutionary Robotics''~\cite{floreano2008bio, bongard2013evolutionary} has generated a lot more attention. For a comprehensive review of evolutionary approaches to robot design, see the survey article by~\citet{prabhu2018survey}.
Of those works, one interesting line of research relevant to Chapter~\ref{sec:soft_design} of this thesis is the optimization of soft robots through evolutionary algorithms~\cite{hiller2011automatic, cheney2014unshackling, cheney2018scalable, joachimczak2016artificial}. These works consider the design and fabrication of pneumatically actuated, voxel-based soft robots for locomotion tasks.
However, despite the popularity of evolutionary methods for co-optimization, there has been only marginal improvements over the work of \citet{sims1994} when considering the scale and complexity of the problems that could be solved~\cite{cheney2018scalable, lipson2016difficulty, joachimczak2016artificial}.
One major symptom noted by~\citet{lipson2016difficulty} is the premature convergence of morphological structure during the optimization, leading to poor performance and a strong dependence on initial seeds.
Many solutions for this problem were proposed, including better evolutionary algorithms~\cite{hornby2003generative, lehman2011evolving, mouret2015illuminating}, better genetic encodings~\cite{hornby2001body, bongard2003evolving, cheney2014unshackling}, more complex tasks to encourage complex bodies~\cite{auerbach2014environmental, cheney2015evolving}, and methods reconsidering the joint structure of the optimization~\cite{cheney2018scalable}.
While all of these solutions play a part, restructuring the optimization to include an outer design optimization loop and an inner control optimization loop---as discussed in~\citet{cheney2018scalable}---appears to significantly simplify the optimization landscape.

In the work of~\citet{sims1994} and those that follow, design and control parameters are mutated simultaneously and once per generation. This can lead to design parameters getting stuck in local optima as mutated controllers will often perform poorly with mutated designs.
The under-optimization of the controller for new morphologies could explain the premature convergence of these algorithms, and~\citet{cheney2018scalable} confirm this by showing that additional evolution of control parameters leads to improved optimization of morphologies in the case of soft robot locomotion. This represents a shift in which the co-optimization problem can be viewed as a nested optimization problem, with the evaluation of physical design parameters requiring an inner control optimization loop.
While this view of co-optimization as nested optimization of physical design and control is not new---for example, the early work of~\citet{park94} solve for a near optimal PD controller for each two-link arm considered during the optimization---it solidifies the view that optimizing control parameters in an inner loop improves the performance of co-optimization methods.
Indeed, as research on this problem has progressed, nearly all recent works have converged on this paradigm of nested optimization of design and control.

\subsection{Neural Approaches to Co-optimization}

As the popularity of deep learning, and especially deep reinforcement learning, grew within the robot learning community, attention within the co-optimization field began to shift from evolutionary approaches to deep learning.
In recent years, several approaches leverage deep reinforcement learning in various ways~\cite{schaff_jointly_2019, ha_reinforcement_2019, luck_data-efficient_2019, chen_hardware_2020, spielberg2019learning, schaff2022soft, pathak2019learning, schaff2022n}.
A common theme shared by methods is addressing the question of how to produce optimized controls for new designs (or estimate the performance of such a controller) in an efficient way.
Initial approaches by~\citet{schaff_jointly_2019} and~\citet{ha_reinforcement_2019} accomplish this by optimizing and leveraging a design-conditioned control policy.
\citet{luck_data-efficient_2019} leverage a design-conditioned Q-function to estimate the performance of the optimal controller for each design and optimize the physical design using an evolutionary strategy (CMA-ES) to maximize this Q-function.
\citet{chen_hardware_2020} model the design space as a differentiable computational graph, which allows them to extend policy gradient methods to optimize design parameters.  
\citet{yuan2021transform2act} represent the design generation process as part of the environment and train a dual-purpose design generation and control policy.
Other approaches leverage deep learning in other ways.
For example, \citet{spielberg2019learning} propose an autoencoder-based method that jointly optimizes the placement and control of a large number deformable voxels for soft-body locomotion tasks. \citet{zhao2020robogrammar} use a model predictive control algorithm to solve for the controls of each design and train a neural network to predict robot performance to aid graph-based search over designs.

It is in this context that the work in this thesis is situated. Chapter~\ref{sec:beacons} and Chapter~\ref{sec:design_and_control} present early work on applying deep learning to both the co-optimization of localization systems and robots, respectively.
We then extend the work of Chapter~\ref{sec:design_and_control} to allow for the optimization over large sets of morphologies (Chapter~\ref{sec:nlimb_paper}) as well as develop a framework for the optimization and fabrication of soft robots (Chapter~\ref{sec:soft_design}). 
\section{Overview of the Thesis}

Deep learning has shown to be adept at making accurate inferences about the world, as well as discovering complex control strategies for many problems in robotics.
Motivated by this success, this thesis seeks to extend deep learning methods to create an algorithmic framework for solving task-driven co-optimization problems in robotics.
To develop this framework, this thesis will focus on two specific problem instances.
The first is the co-optimization of sensors and inference in the context of beacon-based localization (Chapter~\ref{sec:beacons}). This work employs a differentiable physics model and optimizes sensor placement and location inference in an end-to-end fashion.
The second is the co-optimization of the physical body of a robot and its controller in the context of legged locomotion. This work employs multi-task reinforcement learning to learn a control policy that can be used to quickly evaluate and search over a space of physical designs.
This problem was initially considered in the context of optimizing design parameters within a single morphology (Chapter~\ref{sec:design_and_control}) and then later extended to optimize over large sets of morphologies (Chapter~\ref{sec:nlimb_paper}).
These findings were then applied to the challenging area of soft robotics, in which novel simulation techniques and algorithmic improvements were combined to create high-performing crawling robots that were subsequently fabricated and deployed in the real world (Chapter~\ref{sec:soft_design}).

\begin{figure}[t]
    \centering
    \includegraphics[width=0.9\textwidth]{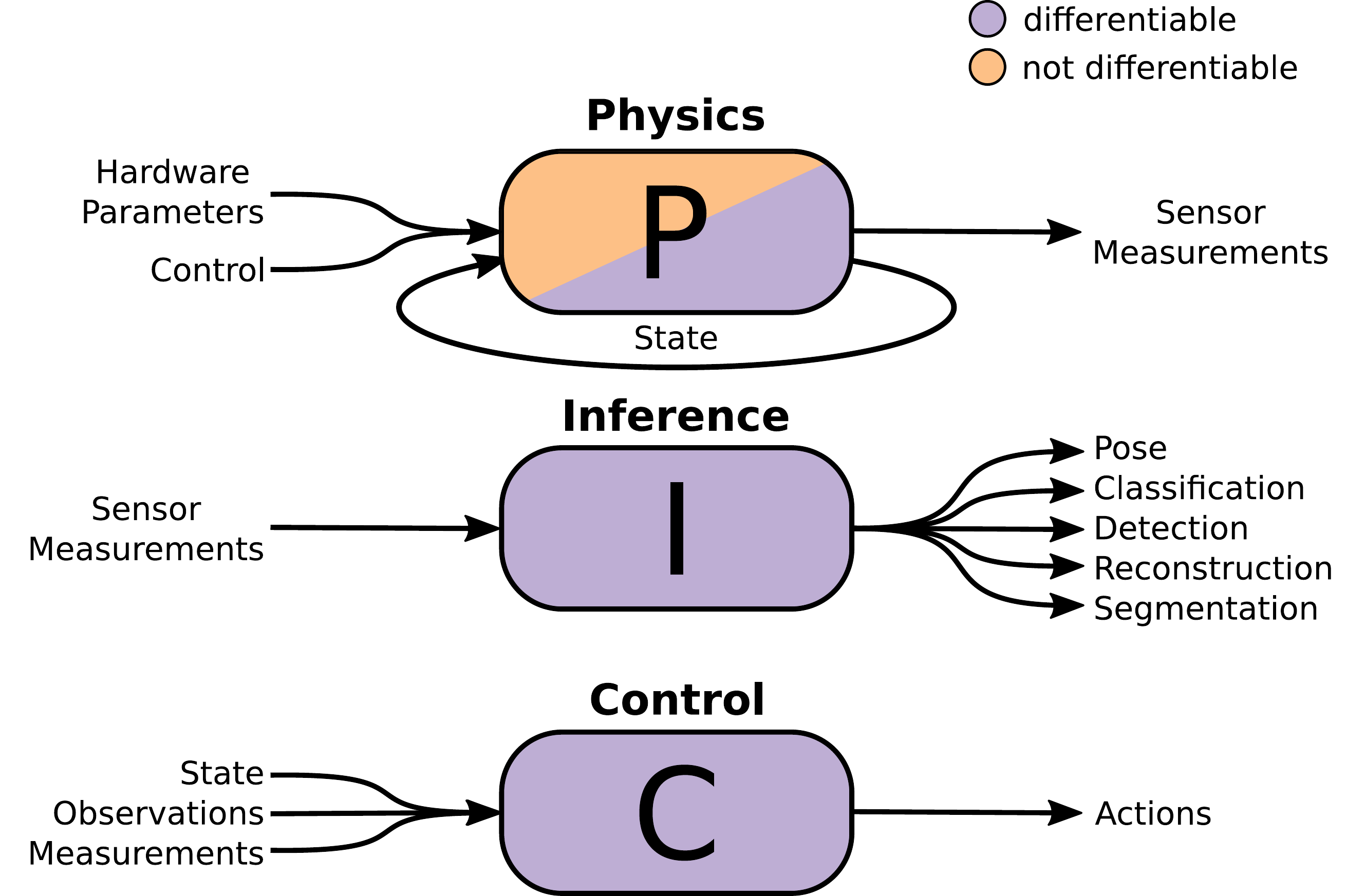}
    \caption{Abstract components for physics, inference, and control. By implementing inference and control algorithms as deep neural networks, both can be optimized together in an end-to-end fashion. Additionally, if physics is modelled as a differentiable function, hardware parameters can also be optimized end-to-end to improve task performance.}
    \label{fig:components}
\end{figure}

The ultimate goal for co-optimization algorithms in the context of robotics is to be able to optimize entire systems, including the physical design, sensors, inference, and control algorithms, in an end-to-end fashion based only on the task definition and knowledge about the distribution of scenarios the system is likely to encounter.
To enable end-to-end optimization, a central theme in this thesis is to view an embodied system and its environment as a unified computational graph that considers physical parameters, physics, inference, and control as a cascade of operations for which the parameters can be optimized in an end-to-end fashion (Figure~\ref{fig:components}).
Depending on the problem instance, physics, inference, and control blocks with varying structure can be stitched together to model the environment and the processes by which the desired task will be solved.
This view encourages the end-to-end optimization of embodied systems directly for task objectives.
Depending on the differentiability of the physics model used with respect to hardware parameters---which may or may not be a reasonable assumption based on the task at hand---all parameters of the system can be optimized end-to-end with deep learning techniques, or multiple optimization problems can be carried out simultaneously.

The following sections briefly outline the work presented in this thesis and summarize key contributions.

\subsection{Optimizing Sensor Design and Inference for Beacon-based Localization}

\begin{figure}[t]
    \centering
    \includegraphics[width=0.9\textwidth]{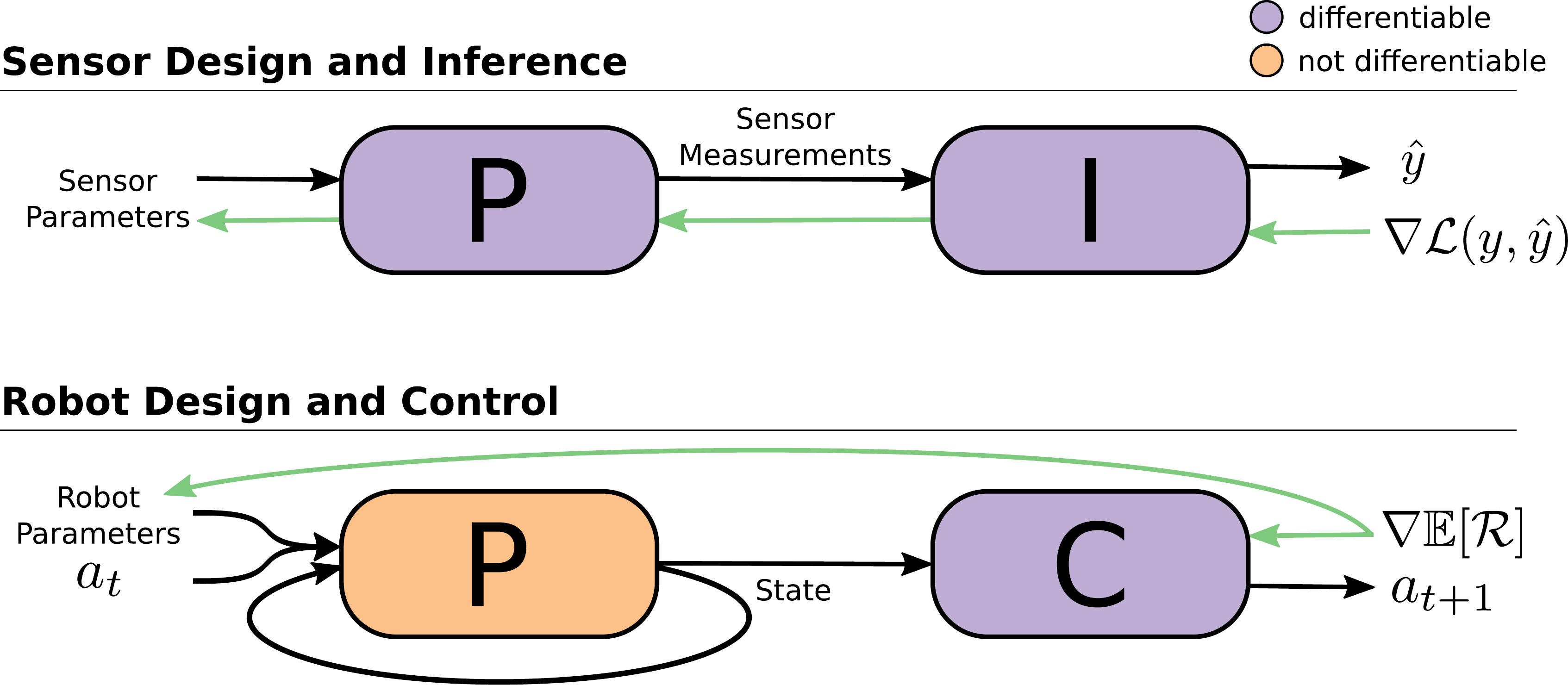}
    \caption{Our work focus on two problem instances: jointly optimizing sensors and inference (with a differentiable physics model), and jointly optimizing the physical design and control of robots (with a black-box physics model). While our proposed algorithms are developed within the context of these problems, they can be combined for other instances (e.g., when optimizing physical parameters, sensors, inference, and control simultaneously).}
    \label{fig:problems}
\end{figure}

Localization is a critical task for human and robot navigation, resource discovery, asset tracking, logistical operations, and resource allocation~\cite{teller03}. 
It settings in which GPS is not available (i.e., indoors, underground, or underwater), measurements of transmissions from fixed beacons~\cite{kurth03, newman03, olson04, djugash06, isler06, detweiler08, amundson09, huang11, kennedy12, derenick13} provide an attractive alternative. Designing a system for beacon-based localization requires simultaneously deciding (a) how the beacons should be distributed (e.g., spatially and across transmission channels); and (b) how location should be determined based on measurements of signals received from these beacons.
Note that these decisions are inherently coupled. The placement of beacons and their channel allocation influence the nature of the ambiguity in measurements at different locations, and therefore which inference strategy is optimal.

Different approaches to this problem consider placement and inference independently and jointly. 
When it is assumed that a beacon network already exists, one common approach to this problem is to infer location from a pre-collected database of received signal strength (RSS) and location pairs~\cite{prasithsangaree02, haeberlen04, park10}.
Similar to our approach, other works use a neural network to predict location from RSS signatures~\cite{sala10, altini10}.
When focusing on beacon placement, heuristics are often used, such as coverage~\cite{agarwal09, kang13}, line-of-sight constraints~\cite{gonzales-banos01}, or signal-to-noise ratio~\cite{fang10}.

In this work, we propose a deep learning algorithm to optimize both beacon placement and location inference. In our approach,
\begin{itemize}
    \item we leverage a differentiable model for signal propagation, which allows phenomena such as noise, interference, and attenuation to directly influence the localization system;
    \item we use a neural network inference strategy that, combined with the differentiable propagation model, allows us to optimize for location accuracy in an end-to-end manner;
    \item using a regularizer on the number of beacons, we allow for explicit trade-offs between accuracy and the cost of placing sensors.
\end{itemize}
For more details, see Chapter~\ref{sec:beacons}.

\subsection{Optimizing Design and Control for Legged Locomotion}

A robot's ability to successfully interact with its environment depends both on its physical design as well as its proficiency at control, which are inherently coupled. Therefore, designing a robot requires reasoning both over the mechanical elements that make up its physical structure as well the control algorithm that regulates its motion.

The co-optimization problem that attempts to automate this design process has a long history in robotics research.
One category of approaches are model-based~\cite{park94,pil96,de02,reyer02,paul06,villarreal13}, which exploit full knowledge about the dynamics of the system. These approaches are often effective but restricted to scenarios with simple dynamics (e.g., two or three link systems).
One well-studied model-based approach formulates this problem as trajectory optimization~\cite{spielberg17}. These methods only require that the physics model be differentiable, but they are limited in their usefulness as they require good initial guesses at a motion trajectory and only produce a single trajectory (as opposed to a state-conditioned policy).
More flexible approaches, which require no knowledge of the dynamics, employ evolutionary methods to optimize both design and control~\cite{sims1994,lipson00,paul01,bongard11}, but they suffer from high sample complexity and tend to yield poor design-control pairs.

In this thesis, we seek a flexible approach that requires no knowledge of dynamics but can outperform evolutionary methods. 
To accomplish this, we jointly optimize design and control using deep reinforcement learning, which has proven effective on a large array of tasks in robotics~\citep{tobin17,rajeswaran17,levine18}.
In our approach, 
\begin{itemize}
    \item we leverage multi-task deep reinforcement learning to efficiently train a design-conditioned control policy on a distribution of designs that is capable of zero-shot generalization to unseen designs; 
    \item we leverage this policy to provide fast evaluations of arbitrary robot designs and use those evaluations to update the design distribution towards better designs;
    \item we treat the physics model as a black-box, allowing for complex and realistic physics simulators to be used with our approach (such as the Bullet physics engine~\cite{coumans2021}).
\end{itemize}
The result is a flexible, fast, and data-driven approach to co-optimization of design and control that leverages recent advances in deep RL. For more details, see Chapter~\ref{sec:design_and_control}.

\subsection{Optimizing Design and Control across Morphologies}

The above work was successful at co-optimizing design and control for a targeted application, but it was restricted to only optimizing the continuous design parameters within a single, user specified morphology.
Chapter~\ref{sec:nlimb_paper} of the thesis extends the above approach to remove this assumption and the need for users to specify a specific morphology for the task. The result is an algorithm that is able to optimize over any number of kinematic or dynamic design parameters such as the number and arrangement of limbs or the position and type of the controllable degrees of freedom.

The search over discrete morphologies and their controllers significantly complicates the approach described above. 
First, it introduces a search over a prohibitively large, discrete, and combinatorial space of morphologies.
And second, state and action spaces can vary across morphologies, making the multi-task reinforcement learning problem more complicated and the design-conditioned control policy challenging to implement.

Our work~\cite{schaff2022n} introduces Neural Limb Optimization (\nlimb) to overcome these challenges and makes the following contributions:
\begin{itemize}
    \item Our approach represents the space of morphologies with a context-free graph grammar. This has two benefits: it allows for the easy incorporation of fabrication constraints and inductive biases into the symbols and expansion rules of the grammar; and it provides a way to iteratively generate designs by sampling expansion rules.
    \item The grammar allows us to define complex, multi-modal distributions over the space of morphologies via a novel autoregressive model that recursively applies expansion rules of the grammar until a complete graph has been formed.
    \item We develop a morphologically-aware transformer architecture to parameterize the design-conditioned control policy. The flexible nature of transformers~\cite{vaswani2017attention} allows the model to easily adapt to changes in the state and action spaces of different morphologies.
\end{itemize}
For more details, see Chapter~\ref{sec:nlimb_paper}.

\subsection{Optimizing Design and Control for Soft Robotics}

The deformable nature soft robots enables designs that respond to contact or control inputs in sophisticated ways, with behaviors that have proven effective across a variety of domains.
This property tightly couples design and control, and ``good'' designs will allow for control algorithms that exploit the deformability of the robot in useful ways.
This coupling makes soft robotics a natural choice for the co-optimization of design and control.

However, the simulation of soft robots is still an open problem. Deformable material simulations are often computationally intensive, and it is challenging to create a simulator that is both accurate enough for sim-to-real transfer and fast enough to sufficiently explore the joint design-control space.
Several simulation techniques have been proposed in recent years that trade off between speed and physical accuracy~\cite{Hu19, Hiller2014, huang2020dynamic, Coevoet17}.
However, recent co-optimization methods for soft robots~\citep{talamini2019evolutionary, cheney2014unshackling, kriegman2019automated, ma2021diffaqua, deimel2017automated} rely on faster simulations, at the expense of accuracy, and rarely attempt to transfer their optimized robots from simulation to reality.

In this work, we propose a complete framework for the simulation, co-optimization, and sim-to-real transfer of the design and control of soft legged robots.
In our approach,
\begin{itemize}
    \item we propose a \textit{design-reconfigurable} model order reduction technique that enables fast and accurate finite element simulation across a space of physical robots;
    \item we extend our previous work on co-optimization of design and control to optimize over a discrete set of designs; 
    \item we build optimized, pneumatically actuated soft robots and transfer their learned controllers to reality, showing that our method produces systems that outperform a standard expert-designed crawling robot in the real world.
\end{itemize}
For more details, see Chapter~\ref{sec:soft_design}.

\chapter{Optimizing Sensor Design and Inference for Beacon-based Localization} \label{sec:beacons}
\newcommand{\ti}{\tilde{I}}
\newcommand{\dd}{\mathcal{D}}
\newcommand{\te}{\tilde{\mathcal{E}}}
\newcommand{\ee}{\mathcal{E}}
\newcommand{\R}{\mathbb{R}}

Measurements obtained through a distributed network of
sensors or beacons can be an effective means of monitoring
location, or the spatial distribution of other phenomena.
The measurements themselves only provide indirect or noisy
information regarding the physical properties of interest, and
so additional computational processing is required for inference. Such inference must be designed to take into account
possible degradation in the measurements (e.g., spurious observations, sensor noise, or missing measurements), and exploit prior statistical knowledge of the environment. However, the success of inference, in the end, is limited by how the
sensors are physically distributed in the first place.

Consider location-awareness, which is critical to human and robot navigation, resource discovery, asset tracking, logistical operations, and resource allocation~\cite{teller03}. In situations for which GPS is unavailable (indoors, underground, or underwater) or impractical, measurements of transmissions from fixed beacons~\cite{kurth03, newman03, olson04, djugash06, isler06, detweiler08, amundson09, huang11, kennedy12, derenick13} provide an attractive alternative. Designing a system for beacon-based location-awareness requires simultaneously deciding (a) how the beacons should be distributed (e.g., spatially and across transmission channels); and (b) how location should be determined based on measurements of signals received from these beacons.

Note that these decisions are inherently coupled. The placement of beacons and their channel allocation influence the nature of the ambiguity in measurements at different locations, and therefore which inference strategy is optimal. Therefore, one should ideally search over the space of both beacon allocation---which includes the number of beacons, and their placement and channel assignment---and inference strategies to find a pair that is jointly optimal. Motivated by the recent work of~\citet{chakrabarti16}, we propose a new learning-based approach to designing the beacon distribution (across space and channels) and inference algorithm \emph{jointly} for the task of localization based on raw signal transmissions. We instantiate the inference method as a neural network, and encode beacon allocation as a differentiable neural layer. We then describe an approach to jointly training the beacon layer and inference network, with respect to a given signal propagation model and environment, to enable accurate location-awareness in that environment.

\begin{figure}[!t]
    \centering
    \includegraphics[width=\linewidth]{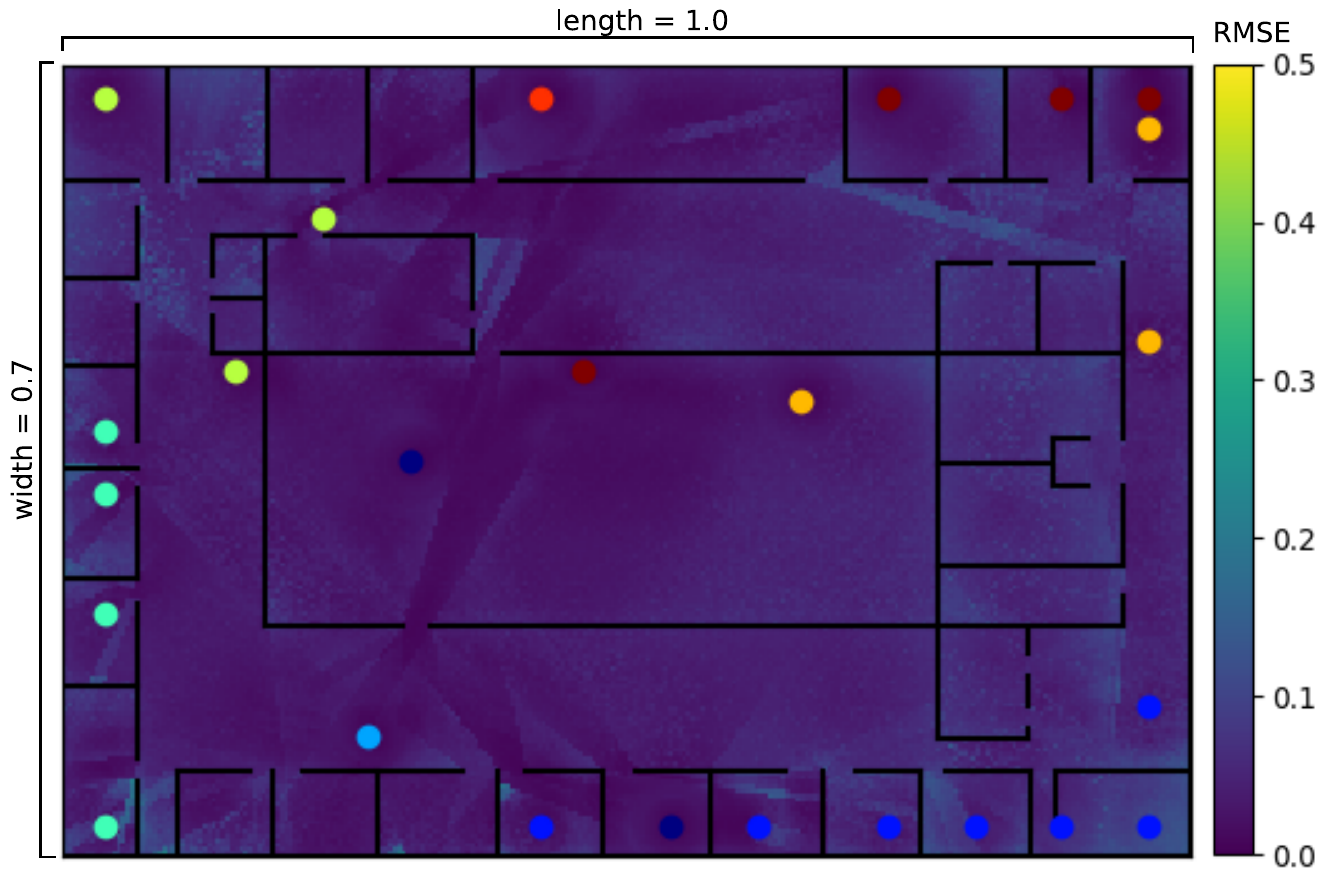}
    \caption{\small A visualization of the jointly optimal beacon allocation and inference model parameters produced by our framework for beacon-based localization for a given floor plan. Beacon placement and channel assignments (colored circles) are optimized with a neural network localization function to maximize location accuracy. The heatmap denotes the root mean squared error (RMSE) in the location estimates of our system throughout the environment.}
    \label{fig:beacons}
\end{figure}
We carry out evaluations under a variety of conditions---with different environment layouts, different signal propagation parameters, different numbers of transmission channels, and different desired trade-offs against the number of beacons. In all cases, we find that our approach automatically discovers designs--each different and adapted to its environment and settings---that enable high localization accuracy (Figure~\ref{fig:beacons}). Therefore, our method provides a way to consistently create optimized location-awareness systems for arbitrary environments and signal types, without expert supervision.

\section{Related Work}
Networks of sensors and beacons have been widely used for localization, tracking, and measuring other spatial phenomena. Many of the design challenges in sensor and beacon are related, since they involve problems that are duals of one another---based on whether the localization target is transmitting or receiving. In our work, we focus on localization with beacons, i.e., where an agent estimates its location based on transmissions received from fixed, known landmarks.

Most of the effort in localization is typically devoted to finding accurate inference methods, assuming the distribution and location of beacons in the environment are given. One setting for these methods is where sensor measurements can be assumed to provide direct, albeit possibly noisy, measurements of relative range or bearing from beacons---an assumption that is typically based on a simple model for signal propagation. Location estimation then proceeds by using these relative range and/or bearing estimates and knowledge of beacon locations. For example, acoustic long baseline (LBL) networks are frequently used to localize underwater vehicles~\cite{newman03,olson04}, while a number of low-cost systems exist that use radio frequency (RF) and ultrasound to measure range~\cite{priyantha05, gu09}.

\citet{moore04} propose an algorithm for estimating location based upon noise-corrupted range measurements, formulating the problem as one of realizing a two-dimensional graph whose structure is consistent with the observed ranges. \citet{detweiler08} describe a geometric technique that estimates a robot's location as it navigates a network of fixed beacons using either range or bearing observations. \citet{kennedy12} employ spectral methods to localize camera networks using relative angular measurements, and \citet{shareef08} use feed-forward and recurrent neural networks for localization based on noisy range measurements.

Another approach, common for RF beacon and WiFi-based networks, is to infer location directly from received signal strength (RSS) signatures. One way to do this is by matching RSS readings against a database of RSS-location pairs~\cite{prasithsangaree02}. This database is typically generated manually via a site-survey, or ``organically'' during operation~\cite{haeberlen04, park10}. \citet{sala10} and \cite{altini10} adopt a different approach, training neural networks to predict a receiver's location within an existing beacon network %
based upon received signal strength.

The above methods deal with optimal ways to infer location given an existing network of beacons. The decision of how to distribute these beacons, however, is often made manually based on expert intuition. Automated placement methods are used rarely, and for very specific settings, such as RSS fingerprint-based localization~\cite{fang10}. The most common of these is to ensure full coverage---i.e., to ensure that all locations are within an ``acceptable range'' of at least one beacon, assuming this condition is sufficient to guarantee accurate localization.

One common instance of optimizing placement for coverage is the standard art-gallery visibility problem~\cite{gonzales-banos01} that seeks placements that ensure that all locations have line-of-sight to at least one beacon. This problem assumes a polygonal environment and that the sensors have an unlimited field-of-view, subject to occlusions by walls (e.g., in the case of cameras). Related, \citet{agarwal09} propose a greedy landmark-based method that solves for the placement of the minimum number of beacons (within a $\log$ factor) necessary to cover all but a fraction of a given polygonal environment. Note that these methods treat occlusions as absolute, while in practice, obstructions often only cause partial attenuation---with points that are close but obstructed observing similar signal strengths as those that are farther away. \citet{kang13} provide an interesting alternative, and like us, use backpropagation to place WiFi access points---but again, only for the objective of maximizing coverage. \citet{fang10} consider localization accuracy for placing wireless access points to maximize receiver signal-to-noise ratios.

The above methods address spatial placement but not transmission channel assignments, and associated issues with interference. Automatic channel assignment methods have been considered previously, but only for optimizing communication throughput~\cite{hills02,ling06}---i.e., to minimize interference from two beacons in the same channel at any location. Note that this is a very different and simpler objective than one of enabling accurate localization, where the goal is to ensure
that there is a unique mapping from every RSS signature (with or without interference) to location.

Our approach provides a way to trade off localization accuracy with the number of beacons, similar to the performance-cost trade-offs considered by the more general problem of sensor selection~\cite{cameron90,isler05,joshi09}.
Some selection strategies are designed with specific inference strategies in mind. \citet{shewry87} and \citet{cressie93} use a greedy entropy-based approach to place sensors, tied to a Gaussian process (GP) model that is used for inference. However, this approach does not model the accuracy of the predictions at the selected locations. \citet{krause08} choose locations for a fixed sensor network that maximize mutual information using a GP model for phenomena over which inference is performed (e.g., temperature).
They show that this problem is NP-complete and propose a polynomial-time approximation algorithm that exploits the submodularity of mutual information to provide placements that are within a constant-factor of the optimal locations.
However, these formulations require that the phenomena be modeled as a GP, and are thus not suitable for the task of beacon-based localization. 
Meanwhile, \citet{cameron90} take a Bayesian decision-theoretic approach to determine sensor placement for localization and recognition tasks, however their formulation is designed for online (i.e., next-best-view) directed search.

\section{Approach}

We formalize the problem of designing a location-awareness system as
that of determining an optimal allocation of beacons $\dd$ and an
inference function $f(\cdot)$, given an environment model $\ee$. For a given
set of $L$ possible locations for beacons, we parameterize the
allocation $\dd = \{I_l\}$ as an assignment $I_l$ to each location
$l \in \{1,\ldots, L\}$, where $I_l\in\{0,1\}^{C+1}, \lvert I_l\rvert=1$ is a
$(C+1)$-dimensional $0$-$1$ vector with all but one entry equal to $0$.
This vector denotes whether to place a beacon at location $l$ and in which of the $C$ possible configurations, i.e., setting $I_l^{c+1}=1$ indicates the presence of a beacon in configuration $c$ whereas $I_l^{c+1}=1$ denotes the absence of a beacon.
In our experiments, a beacon's configuration corresponds to the channel on which it broadcasts.

\begin{figure}[!t]
    \centering
    \includegraphics[width=0.95\textwidth]{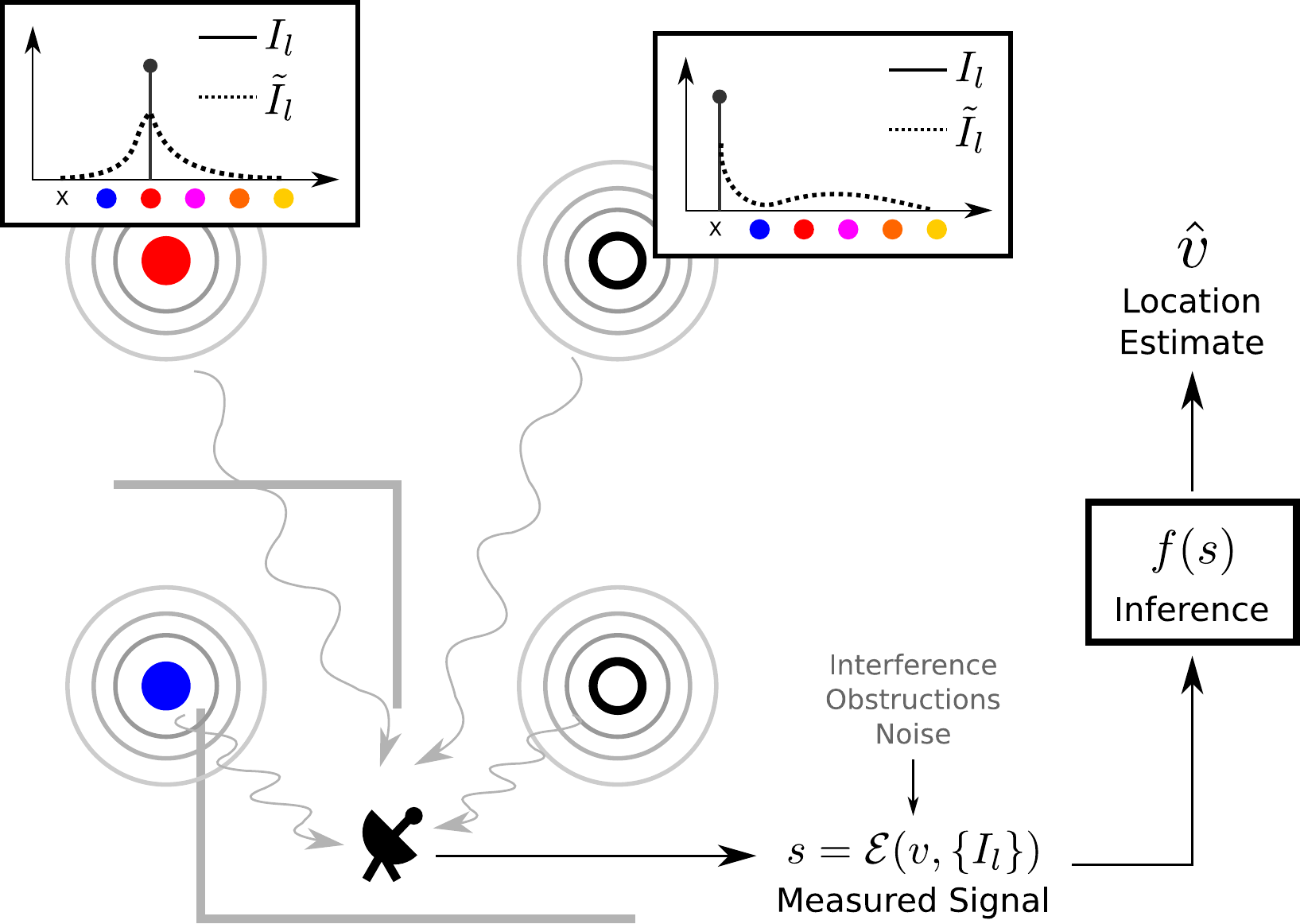}
    \caption{\small A schematic of our proposed approach. Given a set of beacon locations, our approach maintains a distribution over beacon allocations. Using a differential model for signal propagation $\ee(\cdot)$, noisy signal measurements $s$ are produced for location $v$. The sensor measurements are passed to an inference function $f(s)$ to produce a location estimate $\hat{v}$. Then, the beacon placement and parameters of $f(\cdot)$ are jointly optimized for location accuracy.}
    \label{fig:beacon_schema}
\end{figure}
We parameterize the environment in terms of a function $s=\ee(v,\{I_l\})$
that takes as input a location $v\in\mathbb{R}^2$ and a allocation of
beacons $\dd = \{I_l\}$, and produces a vector $s\in\R^m$ of measurements that
an agent is likely to make at that location (Figure~\ref{fig:beacon_schema}). Note that the environment
$\ee(\cdot)$ need not
be a deterministic function. In the case of probabilistic phenomena such
as noise and interference, $\ee(\cdot)$ will produce a sample from the
distribution of possible measurements. The inference function
$f(\cdot)$ is then tasked with computing a accurate estimate of
the location given these measurements.
Our goal is to jointly optimize the beacon allocation $\dd=\{I_l\}$ and inference function $f(\cdot)$ such that
$f(\ee(v,\{I_l\})) \approx v$ for a distribution of possible locations
where the agent may visit. 
To additionally control the beacon allocation, we add a regularizer to our objective that allows for trade-offs between accuracy and the number of beacons placed. 

Unfortunately, the problem as stated above involves a combinatorial
search, since the space of possible beacon allocations is discrete
with $(C+1)^{L}$ elements. We make the optimization tractable by
adopting an approach similar to that of \citet{chakrabarti16}. We
relax the assignment vectors $I_l$ to be real-valued and positive as
$\ti_l \in \R_+^{C+1}, \lvert \ti_l \rvert = 1$. The vector $\ti_l$ can be interpreted as expressing a probability distribution over the possible assignments at location $l$.

Instead of optimizing over distribution vectors $\ti_l$ directly, we
learn a weight vector $w_l\in\R^{C+1}$ such that
\begin{equation}
  \label{eq:wtoi}
  \ti_l = \mbox{SoftMax}(\alpha w_l)\qquad\ti_l^c = \frac{\exp(\alpha w_l^c)}{\sum_{c'}\exp(\alpha w_l^{c'})},
\end{equation}
where $\alpha$ is a positive scalar parameter. Since our goal is to
arrive at values of $\ti_l$ that correspond to hard assignments, we
begin with a small value of $\alpha$ and increase it during the course
of optimization according to an annealing schedule. Small values of
$\alpha$ in initial iterations allow gradients to be backpropagated
across Equation~\ref{eq:wtoi} to update $\{w_l\}$. As optimization
progresses, increasing $\alpha$ causes the distributions $\{I_l\}$ to
get ``peakier'', until they converge to hard assignments.

We also define a distributional version of the environment mapping
$\te(v,\{\ti_l\})$ that operates on these distributions instead of
hard assignments. This mapping can be interpreted as producing the
\emph{expectation} of the signal vector $s$ at location $v$, where the
expectation is taken over the distributions $\{\ti_l\}$. We require
that this mapping be differentiable with respect to the distribution
vectors $\{\ti_l\}$. In the next sub-section, we describe an
example of an environment mapping and its distributional version
that satisfies this requirement.

Next, we simply choose the inference function $f(\cdot)$ to be
differentiable and have some parametric form
(e.g., a neural network), and learn its parameters
jointly with the weights $\{w_l\}$ of the beacon distribution as the
minimizers of the loss function:
\begin{equation}
  \label{eq:lossdef}
  L(\{w_l\},\Theta) = R\left(\{\ti_l\}\right) + \frac{1}{\|\mathcal{V}\|}\sum_{v\in\mathcal{V}} \mathbb{E}_s%
   \left\|v-f\left(\te\left(v,\{\ti_l\}\right); \Theta \right)\right\|^2,
\end{equation}
where $\mathcal{V}$ is the set of possible agent locations, $\Theta$
are the parameters of the inference function $f(\cdot)$,
$\ti_l=$SoftMax$(\alpha w_l)$, as $\alpha\rightarrow\infty$, and $R(\cdot)$
is a regularizer. Note that the inner expectation in the second term
of Equation~\ref{eq:lossdef} is with respect to the distribution of possible
signal vectors for a fixed location and beacon distribution, and
captures the variance in measurements due to noise, interference, etc.

Since we require $f(\cdot)$ and $\te(\cdot)$ to be differentiable, we can optimize
both $\Theta$ and $\{w_l\}$ by minimizing Equation~\ref{eq:lossdef}
with stochastic gradient descent (SGD),
computing gradients over a small batch of locations
$v \in \mathcal{V}$, with a single sample of $s$ per location. We
find that the quadratic schedule for $\alpha$ used by
\citet{chakrabarti16} works well, i.e., at iteration $t$ we set $alpha$ in the following way:
\begin{equation} \label{eq:reg_schedule}
    \alpha=\alpha_0(1+\gamma t^2),
\end{equation}
where $\alpha_0$ defines the initial value of the regularization constant and $\gamma$ determines the rate at which it increases.

\section{Application to RF-based Localization}

To give a concrete example of an application of this framework, we
consider the following candidate setting of localization using RF
beacons. We assume that each beacon transmits a sinusoidal signal at
one of $C$ frequencies (channels). The amplitude of this signal is
assumed to be fixed for every beacon, but we allow different beacons
to have arbitrary phase variations amongst them.

We assume the agent to be located has a receiver with multiple
band-pass filters and is able to measure the power in each channel
separately (i.e., the signal vector $s$ is $C$-dimensional). We assume
that the power of each beacon's signal drops as a function of distance
and the number of obstructions (e.g., walls) in the line-of-sight
between the agent and the beacon. The measured power in each channel
at the receiver is then based on the amplitude of the super-position
of signals from all beacons transmitting on that channel. This
super-position is a source of interference, since individual beacons
have arbitrary phase. We also assume that there is some measurement
noise at the receiver.

We assume all beacons transmit at power $P_0$, and model the power of
the attenuated signal received from beacon $l$ at location $v$ as
\begin{equation}
  \label{eq:plv}
  P_l(v) = P_0~r_{l:v}^{-\zeta}~\beta^{o_{l:v}},
\end{equation}
where $r_{l:v}$ is the distance
between $v$ and the beacon location $l$, $o_{l:v}$ is the number
of obstructions intersecting the line-of-sight between them, and $\zeta$ and $\beta$ are scalar parameters. The measured
power $s=\ee(v,\{I_l\})$ in each channel at the receiver is then
modeled as
\begin{equation}
  \label{eq:meas}
  s^c = \textrm{min} \left(\tau, \left[\epsilon_1 + \sum_l I_l^{c+1}\sqrt{P_l(v)}\cos \phi_l\right]^2 +
\left[\epsilon_2 + \sum_l I_l^{c+1}\sqrt{P_l(v)}\sin \phi_l\right]^2 \right),
\end{equation}
where $\phi_l$ is the phase of beacon $l$, $\epsilon_1$ and $\epsilon_2$ correspond to sensor noise, and $\tau$ is a threshold that we use to clip the signal to model sensor saturation.
At each invocation of the environment function, we randomly sample the
phases $\{\phi_l\}$ from a uniform distribution between $[0,2\pi)$,
and noise terms $\epsilon_1$ and $\epsilon_2$ from a zero-mean Gaussian distribution with
variance $\sigma_z^2$.

During training, the distributional version of the
environment function $\te(\cdot)$ is constructed simply by replacing $I_l$ with
$\ti_l$ in Equation~\ref{eq:meas}. For regularization, we use a term that
penalizes the total number of beacons with a weight
$\lambda$:
\begin{equation}
  \label{eq:regdef}
  R\left(\{\ti_l\}\right) = \lambda \sum_l \ti_l^1.
\end{equation}

\section{Experiments}

In this section, we evaluate our method through a series of simulation-based experiments on three different environment maps.
We first show that our inference network is effective at localization given fixed beacon placements by comparing to a standard nearest-neighbors method, which infers location estimates from a database of RSS-location pairs.
Second, we compare joint optimization against only optimizing location inference with several hand-designed beacon allocation strategies.
We then analyze the performance when learning the beacon allocation and inference network jointly, demonstrating that our method learns a distribution and inference strategy that enable high localization accuracy for all three environments. We end by analyzing the effects of different degrees of regularization and different numbers of available channels, as well as the variation in the learned beacon distributions based on different settings for the parameters of the environment function $\ee(\cdot)$.

\subsection{Experiment Setup}

We conduct our experiments on three environment maps,
which correspond to floor plans (of size $1\times 0.7$ map units)
with walls that serve as obstructions. For each map, we arrange $L=625$
possible beacon locations in a $25 \times 25$ evenly spaced grid. We
consider configurations with values of $C=4,8,$ and $16$ RF-channels.

Our experiments use the environment model defined in Equation~\ref{eq:plv}
with $P_0 = 6.25\times 10^{-4}$, $\zeta=2.0$ (where locations are in
map units), and $\beta = e^{-1.0}$, with sensor noise variance
$\sigma_z^2=10^{-4}$. The sensor measurements are saturated at a
threshold $\tau=1.0$.
We use parameters $\alpha_0 = 1$ and $\gamma = 1.25\times10^{-9}$ for the quadratic temperature scheme (Equation~\ref{eq:reg_schedule}).
These values were chosen empirically so that the beacon selection vectors
$\tilde{I}_l$ converge at the same pace as it takes the inference
network to learn (as observed while training on a fixed beacon
distribution).  After $900$k iterations, we switch the softmax (Equation~\ref{eq:wtoi})
to an ``arg-max'', effectively setting $\alpha$ to infinity and fixing
the beacon placement, and then continue training the inference network.

The inference function $f(\cdot)$ is parameterized
as a $13$-layer feed-forward neural network. Our architecture consists
of $6$ blocks of $2$ fully-connected layers. All hidden layers contain
$1024$ units and are followed by ReLU activation. Each block is
followed by a max pooling operation applied on disjoint sets of $4$
units. After the last block, there is a final output layer with $2$
units that predicts the $(x,y)$ location coordinates.

During training, locations are randomly sampled and fed through our
environment model to the inference network in batches of $1000$. All
networks are trained by minimizing the loss defined in
Equation~\ref{eq:lossdef} for $1000$k iterations using SGD with a
learning rate of $0.01$ and momentum $0.9$, followed by an additional $100$k iterations with a learning rate of $0.001$. We also use
batch-normalization in all hidden layers.
\begin{figure*}[!t]
    \centering
    \tiny
    \subfigure[\tiny Learned: Annealed Reg. ($0.0497$)]{\rotatebox{90}{\scriptsize \bf ~~~~~~~~~~~~~Map 1}\includegraphics[width=0.245\textwidth]{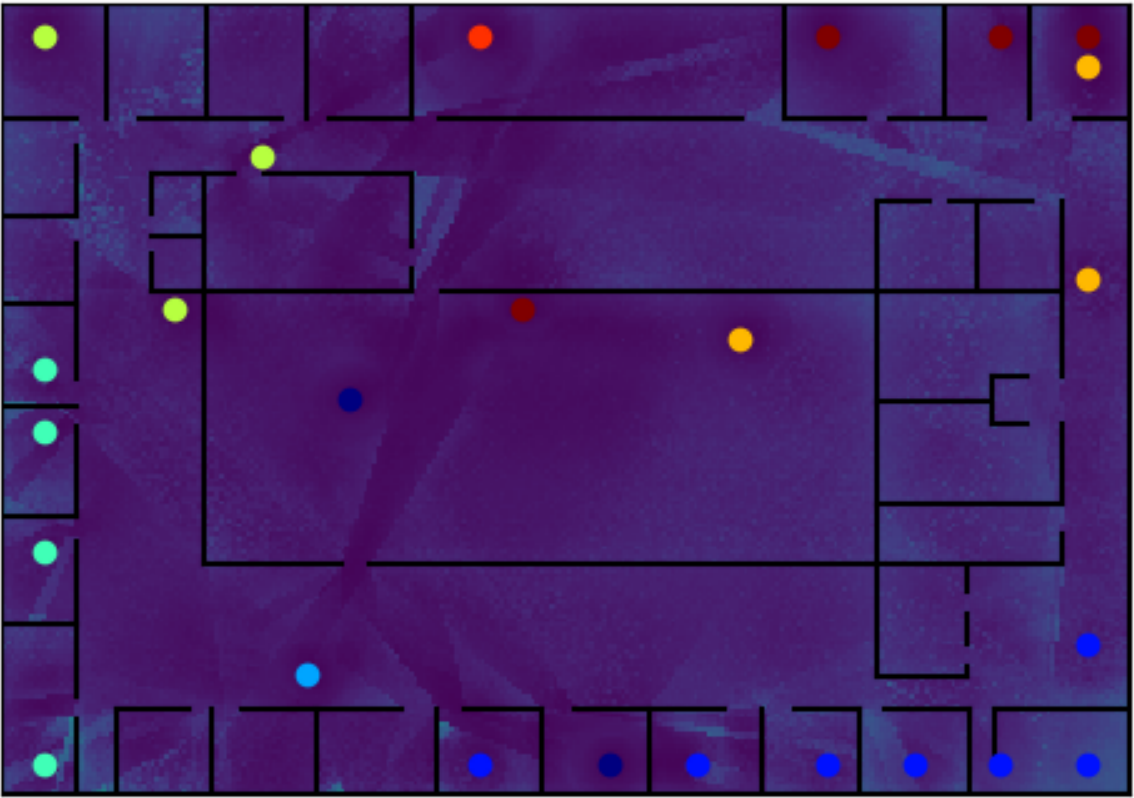}}\hfil%
    \subfigure[\tiny Learned: High Fixed Reg. ($0.0633$)]{\includegraphics[width=0.245\textwidth]{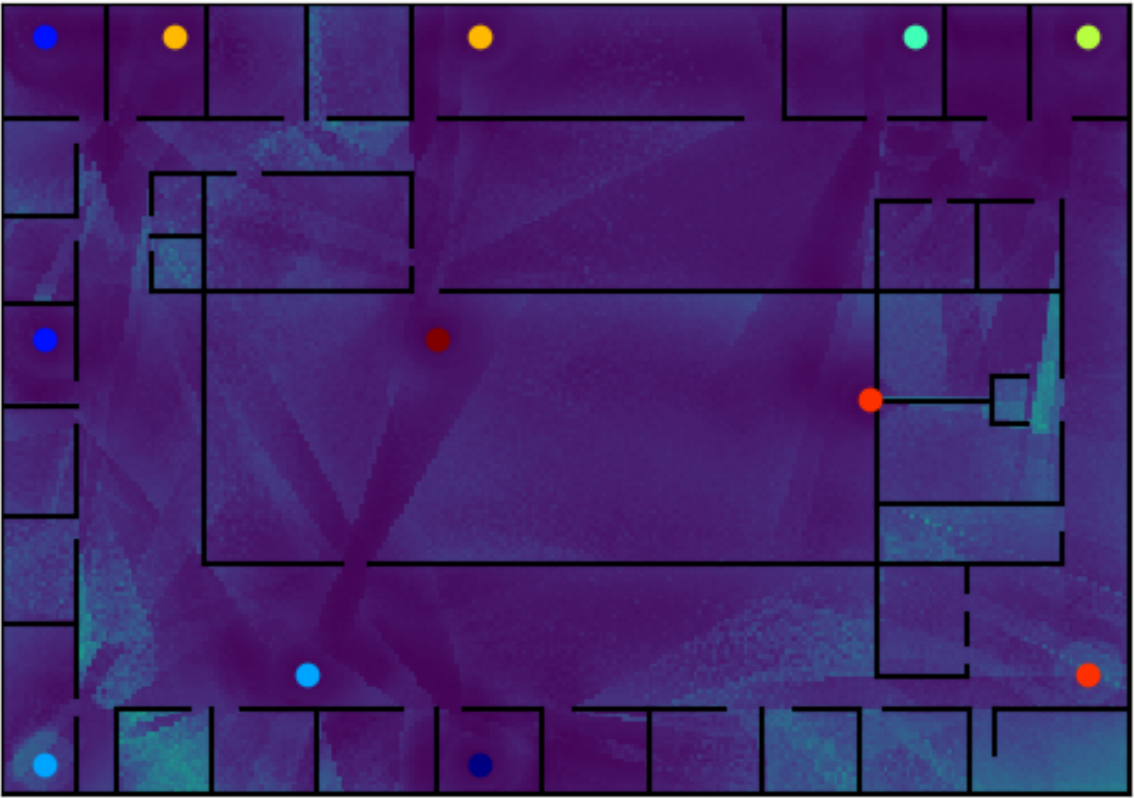}}\hfil%
    \subfigure[\tiny Learned: Low Fixed Reg. ($0.0567$)]{\includegraphics[width=0.245\textwidth]{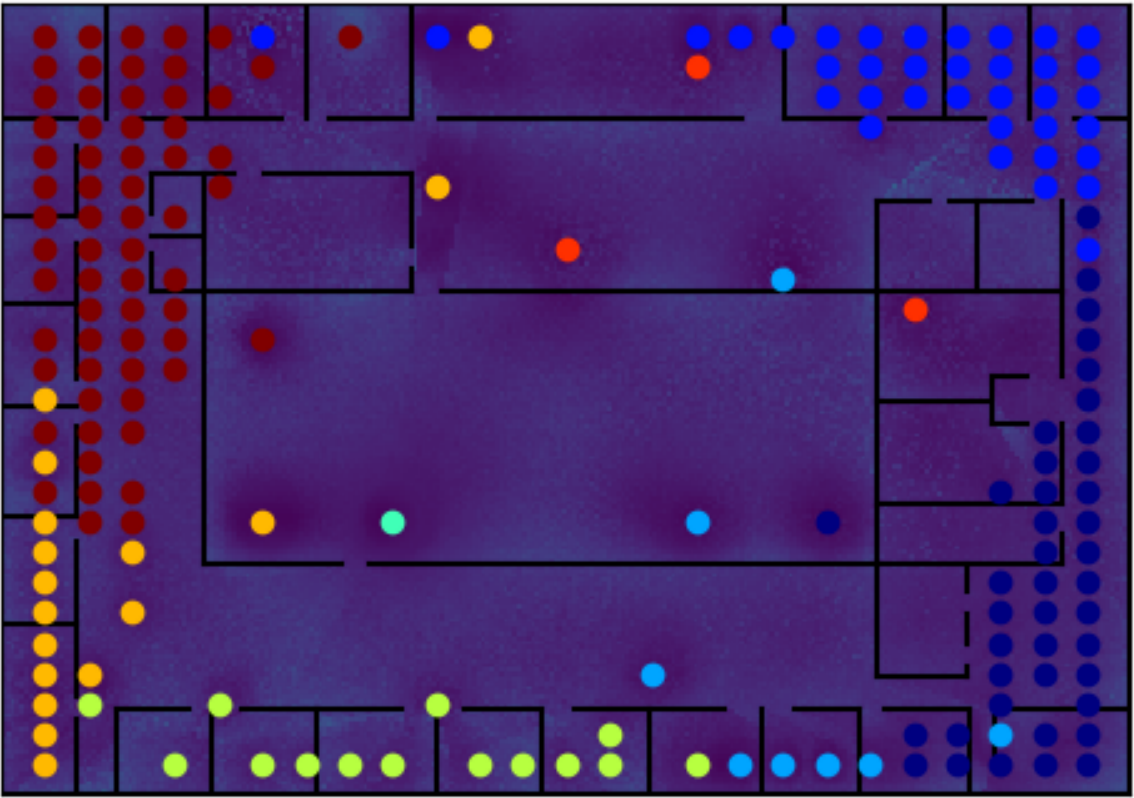}}\hfil
    \subfigure[\tiny Handcrafted A ($0.0716$)]{\includegraphics[width=0.245\textwidth]{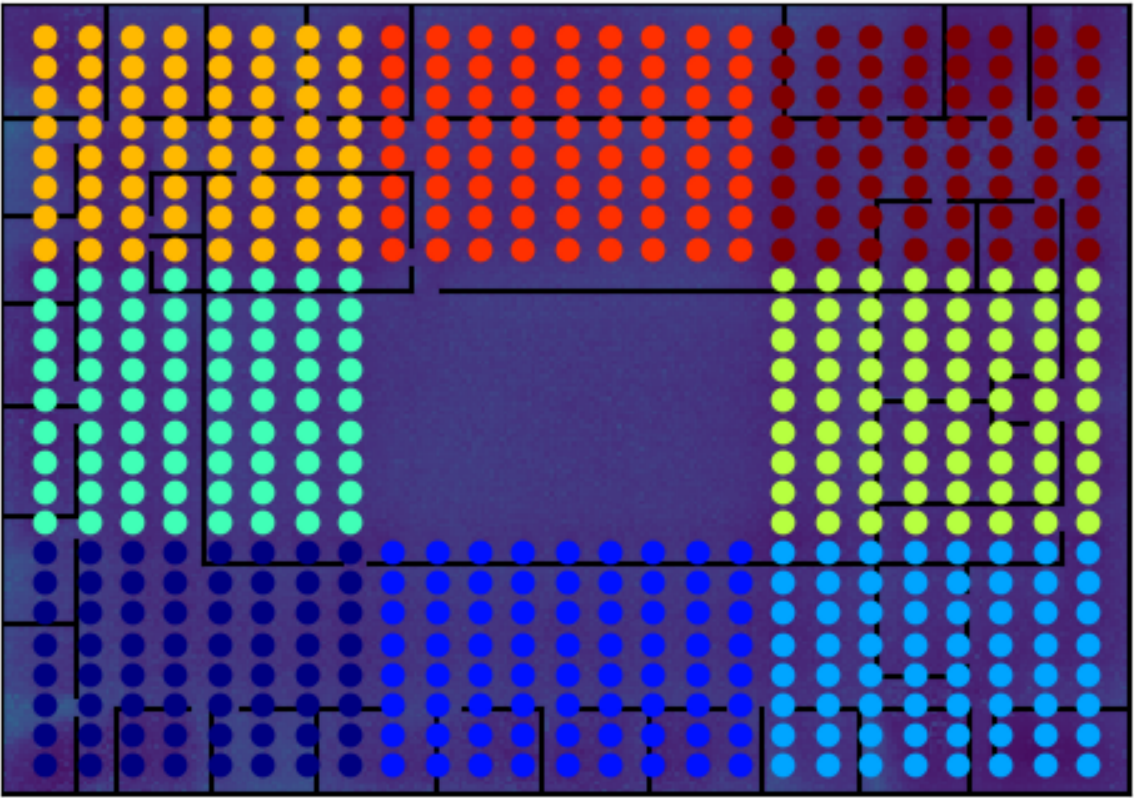}}\vspace{-0.75em}\\%
    \subfigure[\tiny Learned: Annealed Reg. ($0.0473$)]{\rotatebox{90}{\scriptsize \bf ~~~~~~~~~~~~~Map 2}\includegraphics[width=0.245\textwidth]{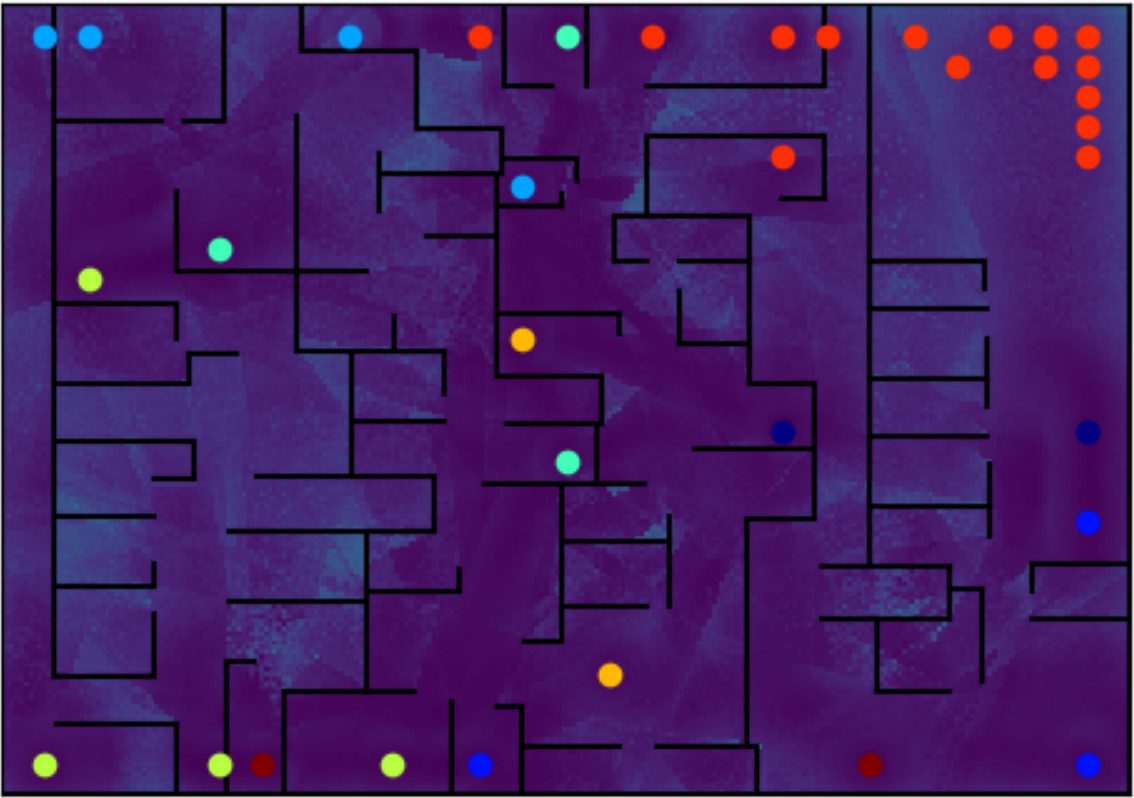}}\hfil
    \subfigure[\tiny Learned: High Fixed Reg.  ($0.0688$)]{\includegraphics[width=0.245\textwidth]{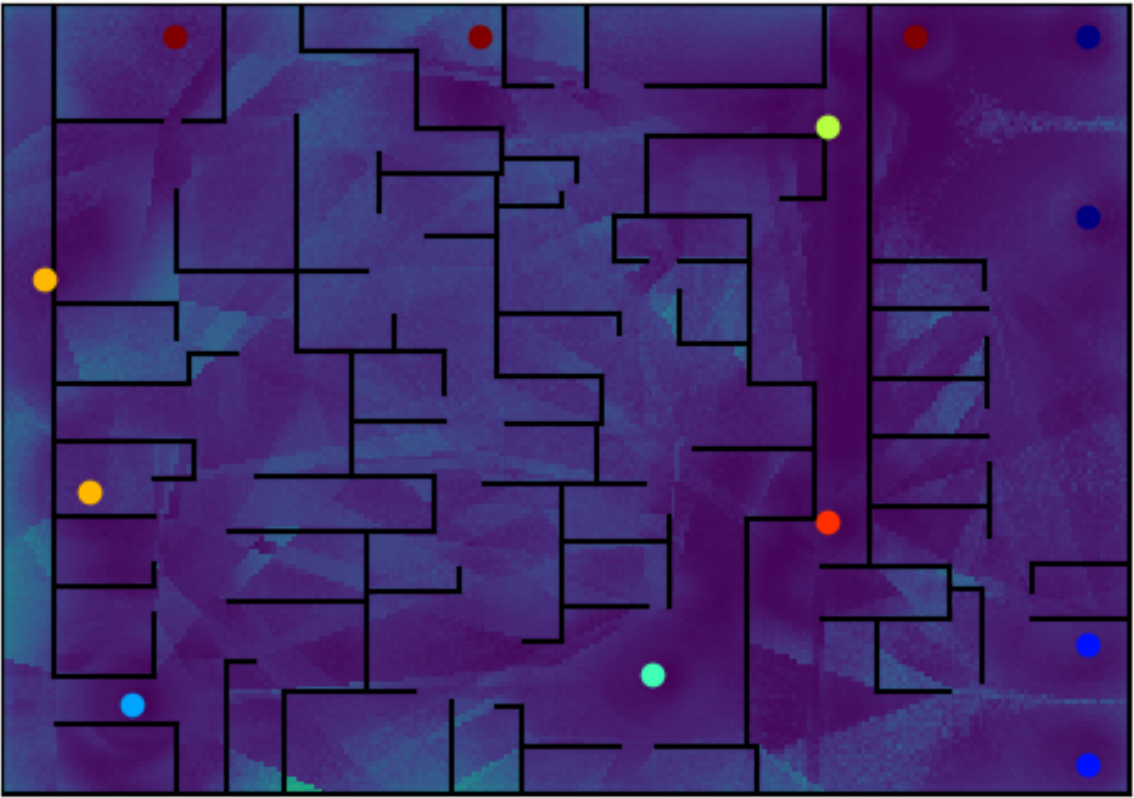}}\hfil
    \subfigure[\tiny Learned: Low Fixed Reg. ($0.04907$)]{\includegraphics[width=0.245\textwidth]{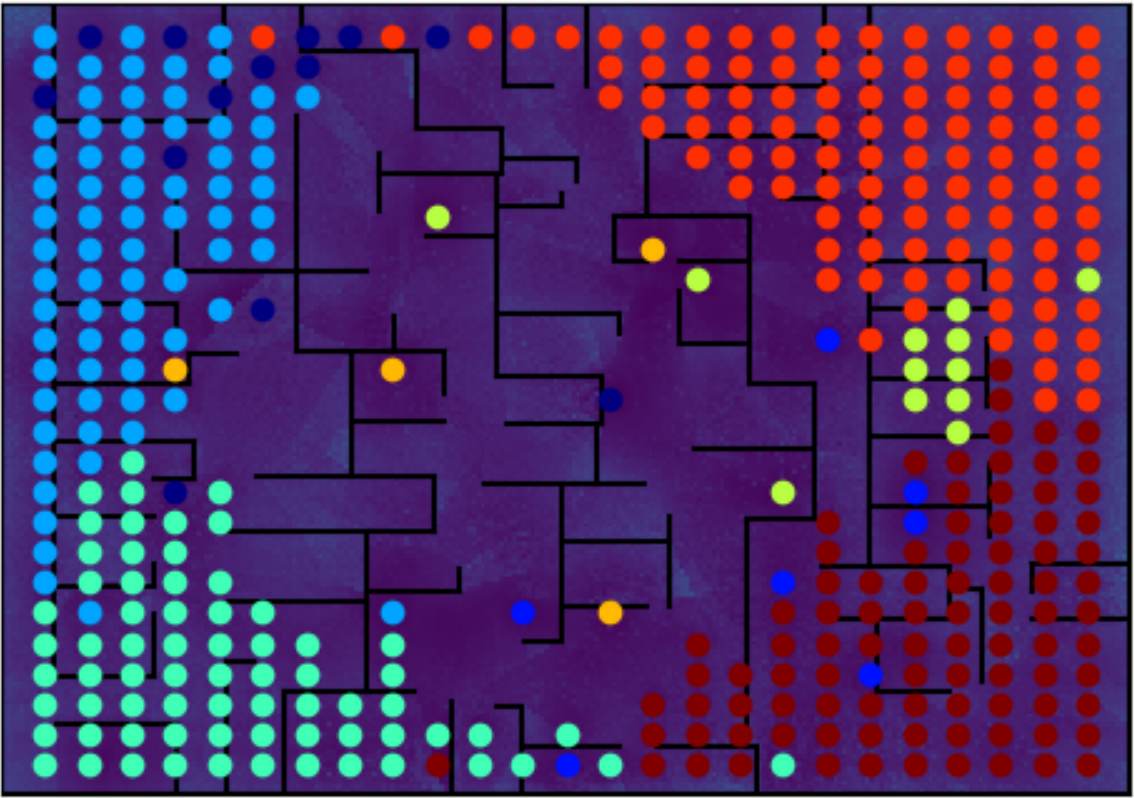}}\hfil
    \subfigure[\tiny Handcrafted B ($0.06242$)]{\includegraphics[width=0.245\textwidth]{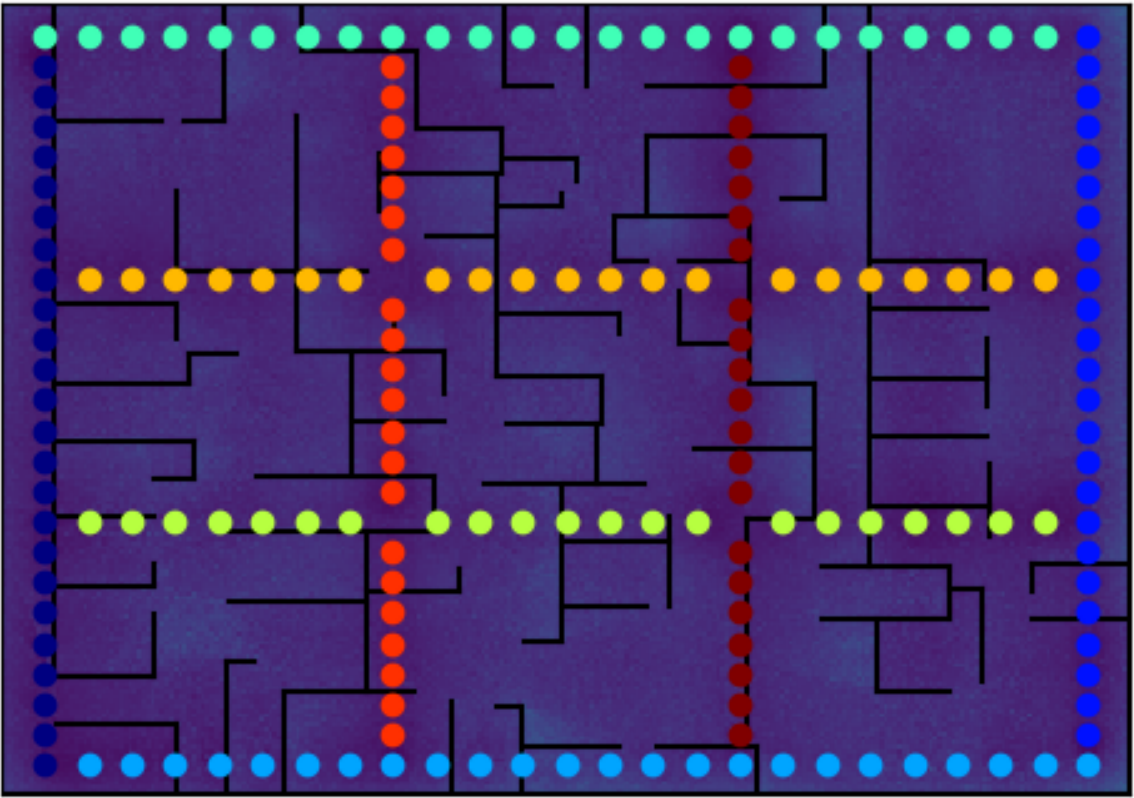}}\vspace{-0.75em}\\%
    \subfigure[\tiny Learned: Annealed Reg. ($0.0493$)]{\rotatebox{90}{\scriptsize \bf ~~~~~~~~~~~~~Map 3}\includegraphics[width=0.245\textwidth]{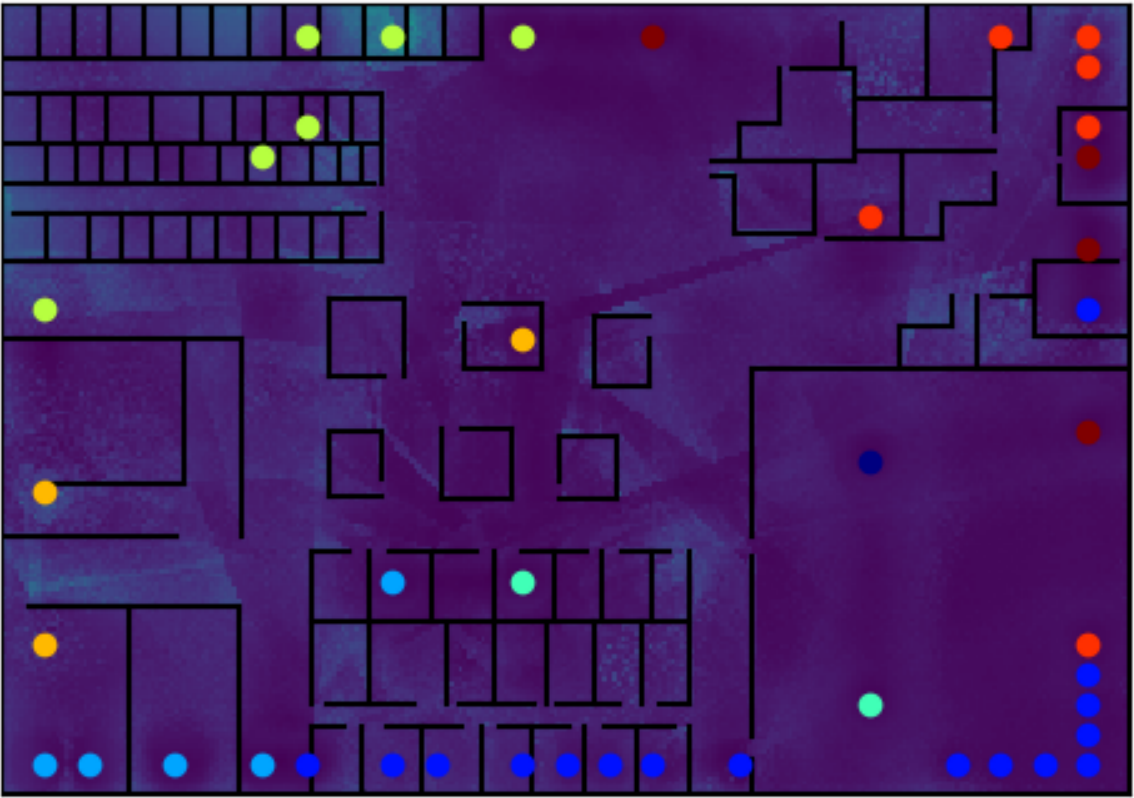}}\hfil
    \subfigure[\tiny Learned: High Fixed Reg.  ($0.0680$)]{\includegraphics[width=0.245\textwidth]{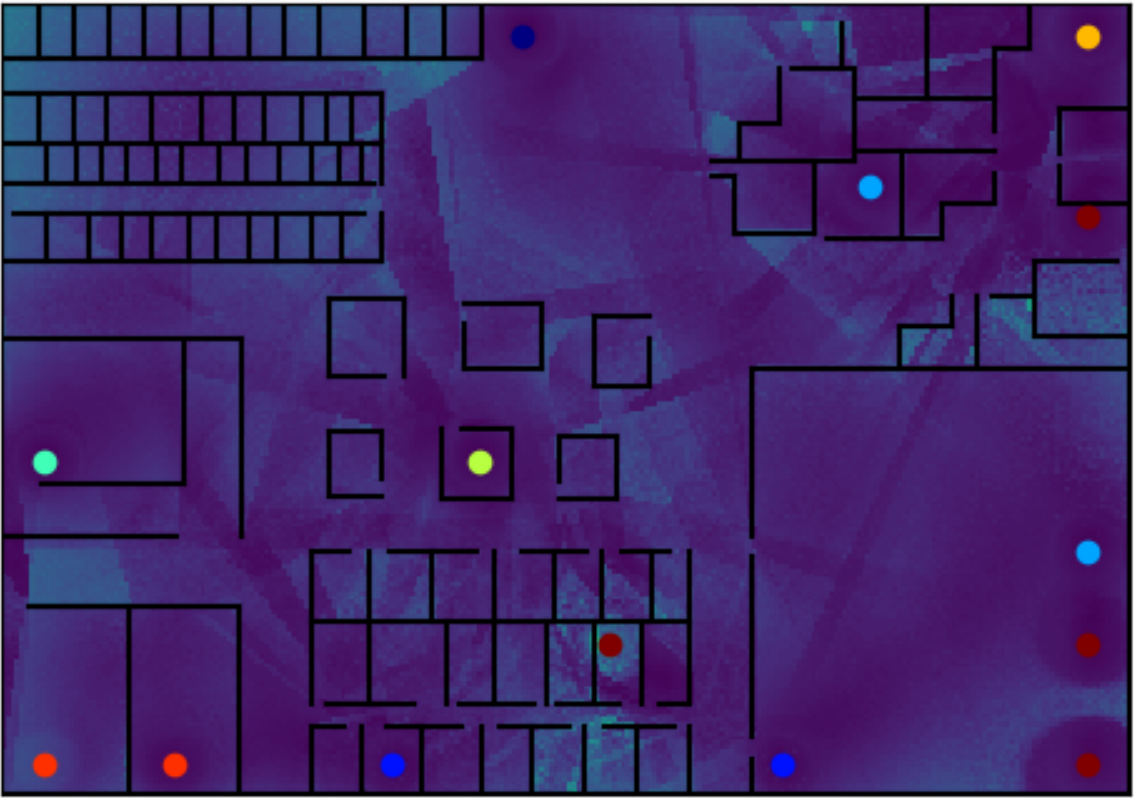}}\hfil
    \subfigure[\tiny Learned: Low Fixed Reg. ($0.0506$)]{\includegraphics[width=0.245\textwidth]{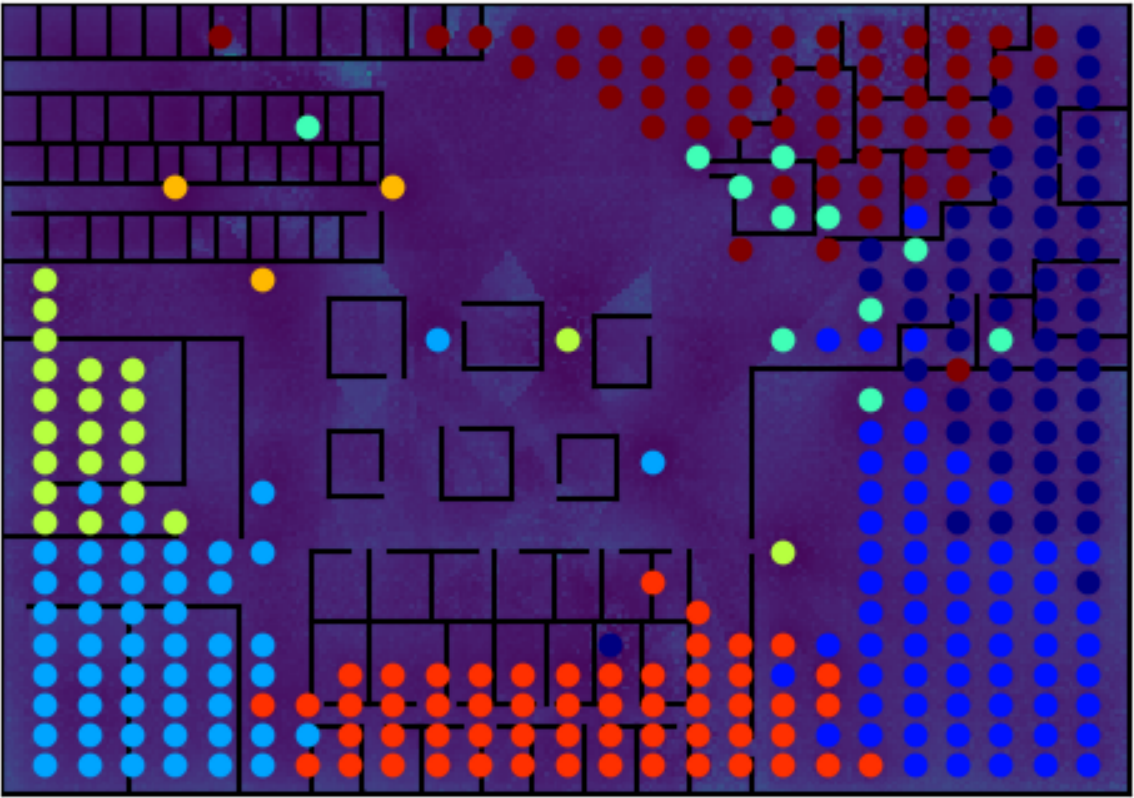}}\hfil
    \subfigure[\tiny Handcrafted B ($0.0649$)]{\includegraphics[width=0.245\textwidth]{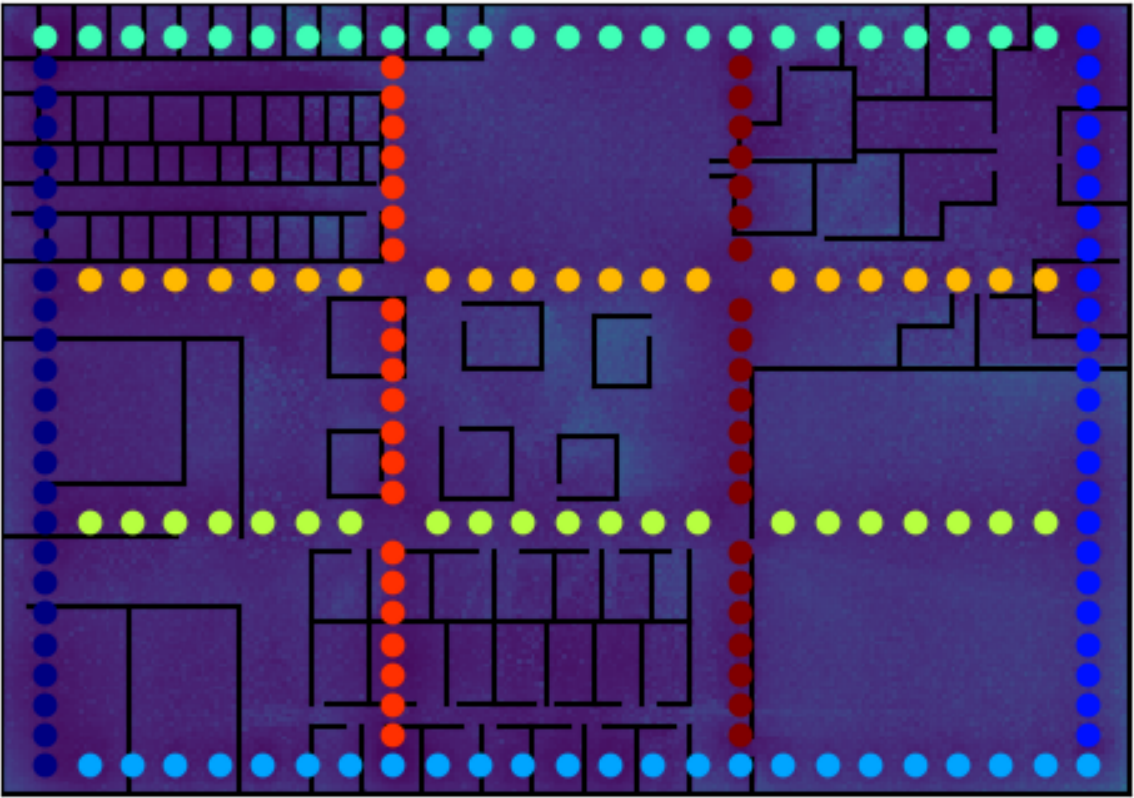}}
    \caption{Localization error maps for learned and handcrafted beacon distributions (with $8$ channels) for three environments. We show three learned distributions for each map---learned with annealed regularization, and two settings (high, low) of fixed regularization---as well as the best performing handcrafted placement. The overall RMSE is indicated below each error map (see Fig.~\ref{fig:beacons} for color mapping for errors).}
     \label{fig:map1-error}
\end{figure*}
\begin{table*}[!t]
    \centering
    \scriptsize
    \setlength{\tabcolsep}{4pt}
    \caption{Performance Comparison for Different Inference and Placement Strategies}\label{tab:overall}
    \begin{tabularx}{0.92\textwidth}{c l l c c c c c c}
        \toprule
        Map & Allocation & Inference & Beacons & RMSE & RMSE (Worst-case) & Failure Rate ($0.10$) & Failure Rate ($0.20$)\\
        \toprule
        \multirow{7}{*}{1} & Handcrafted A & kNN & $544$ & $0.0817$ & $0.1793$ & $19.1946~\%$ & $1.4662~\%$\\
        & Handcrafted B & kNN & $180$ & $0.0998$ & $0.2384$ & $26.1523~\%$ & $4.7400~\%$\\\cline{2-9}
        & Handcrafted A & Network & $544$ & $0.0716$ & $0.1537$ & $13.5691~\%$ & $0.6706~\%$\\
        & Handcrafted B & Network & $180$ & $0.0811$ & $0.1940$ & $17.1529~\%$ & $2.4632~\%$\\\cline{2-9}
        & Learned (low fixed reg.) & Network & $183$ & $0.0567$ & $0.1336$ & $\hphantom{1}6.4917~\%$ & $0.4866~\%$\\
        & Learned (high fixed reg.) & Network & $\hphantom{1}12$ & $0.0633$ & $0.1511$ & $\hphantom{1}8.0217~\%$ & $1.3686~\%$\\
        & Learned (annealed reg.) & Network & $\hphantom{1}25$ & $\mathbf{0.0497}$ & $\mathbf{0.1169}$ & $\hphantom{1}\mathbf{4.7705}~\%$ & $\mathbf{0.3706}~\%$\\
        \toprule
        \multirow{7}{*}{2} & Handcrafted A & kNN & $544$ & $0.0806$ & $0.1708$ & $18.7127~\%$ & $1.3098~\%$\\
        & Handcrafted B & kNN & $180$ & $0.0839$ & $0.2040$ & $18.3838~\%$ & $2.4336~\%$\\\cline{2-9}
        & Handcrafted A & Network & $544$ & $0.0653$ & $0.1331$ & $10.4884~\%$ & $\mathbf{0.2718~\%}$\\
        & Handcrafted B & Network & $180$ & $0.0624$ & $0.1479$ & $\hphantom{1}9.1473~\%$ & $0.6843~\%$\\\cline{2-9}
        & Learned (low fixed reg.) & Network & $371$ & $0.0491$ & $0.1154$ & $\hphantom{1}\mathbf{4.3015~\%}$ & $0.2931~\%$\\
        & Learned (high fixed reg.)& Network & $\hphantom{1}13$ & $0.0688$ & $0.1450$ & $11.9567~\%$ & $1.4807~\%$\\
        & Learned (annealed reg.)& Network & $\hphantom{1}35$ & $\mathbf{0.0473}$ & $\mathbf{0.1142}$ & $\hphantom{1}4.6226~\%$ & $0.4182~\%$\\
        \toprule
        \multirow{7}{*}{3} & Handcrafted A & kNN & $544$ & $0.0814$ & $0.1717$ & $18.7939~\%$ & $1.6503~\%$\\
        & Handcrafted B & kNN & $180$ & $0.0840$ & $0.2127$ & $16.6865~\%$ & $2.6309~\%$\\\cline{2-9}
        & Handcrafted A & Network & $544$ & $0.0670$ & $0.1432$ & $11.1586~\%$ & $0.6904~\%$\\
        & Handcrafted B & Network & $180$ & $0.0649$ & $0.1581$ & $\hphantom{1}9.7563~\%$ & $1.0512~\%$\\\cline{2-9}
        & Learned (low fixed reg.) & Network & $325$ & $0.0506$ & $0.1200$ & $\hphantom{1}\mathbf{4.2496~\%}$ & $\mathbf{0.3482~\%}$\\
        & Learned (high fixed reg.)& Network & $\hphantom{1}14$ & $0.0680$ & $0.1495$ & $11.6651~\%$ & $1.2166~\%$\\
        & Learned (annealed reg.)& Network & $\hphantom{1}43$ & $\mathbf{0.0493}$ & $\mathbf{0.1158}$ & $\hphantom{1}4.7211~\%$ & $0.6529~\%$\\
        \bottomrule
    \end{tabularx}
\end{table*}

\section{Results}

Localization performance can vary significantly within an environment. Consequently, for each beacon allocation and inference strategy, we report both average as well as worst-case performance over a dense set of locations, with multiple samples (corresponding to different random noise and interference phases) per location. We measure performance in terms of the root mean squared error (RMSE)---between estimated and true coordinates---over all samples at all locations, as well as over the worst sample at each location, which we refer to as \textit{worst-case RMSE}. We also measure the frequency with which large errors occur---defined  with respect to different thresholds ($0.1, 0.2$)---and report these as \textit{failure rates}.

Table~\ref{tab:overall} reports these metrics on the three environment maps for different versions our approach that vary the regularization settings. We have also experimented with a number of manually handcrafted distribution strategies for these maps and report the performance of the two strategies that worked best in Table~\ref{tab:overall}. For the handcrafted settings, we report the result of training a neural network for inference, as well as of $k$-nearest neighbors (kNN)-based inference (we try $k\in \{1,5,10,20\}$ and pick the setting that yields the best performance). These latter experiments show that the network-based inference performs well (better than the kNN baseline) and, therefore, that our architecture is reasonable for the task. Moreover, the results reveal that jointly optimizing beacon allocation and inference provides accuracies that exceed the  handcrafted baselines, yielding different distribution strategies with different numbers of beacons. Figure~\ref{fig:map1-error} visualizes the beacon placement and channel allocation along with the RMSE for both learned and handcrafted beacon allocations.

\subsection{The effect of regularization}
\begin{figure}[!t]
    \centering
    \includegraphics[width=0.9\linewidth]{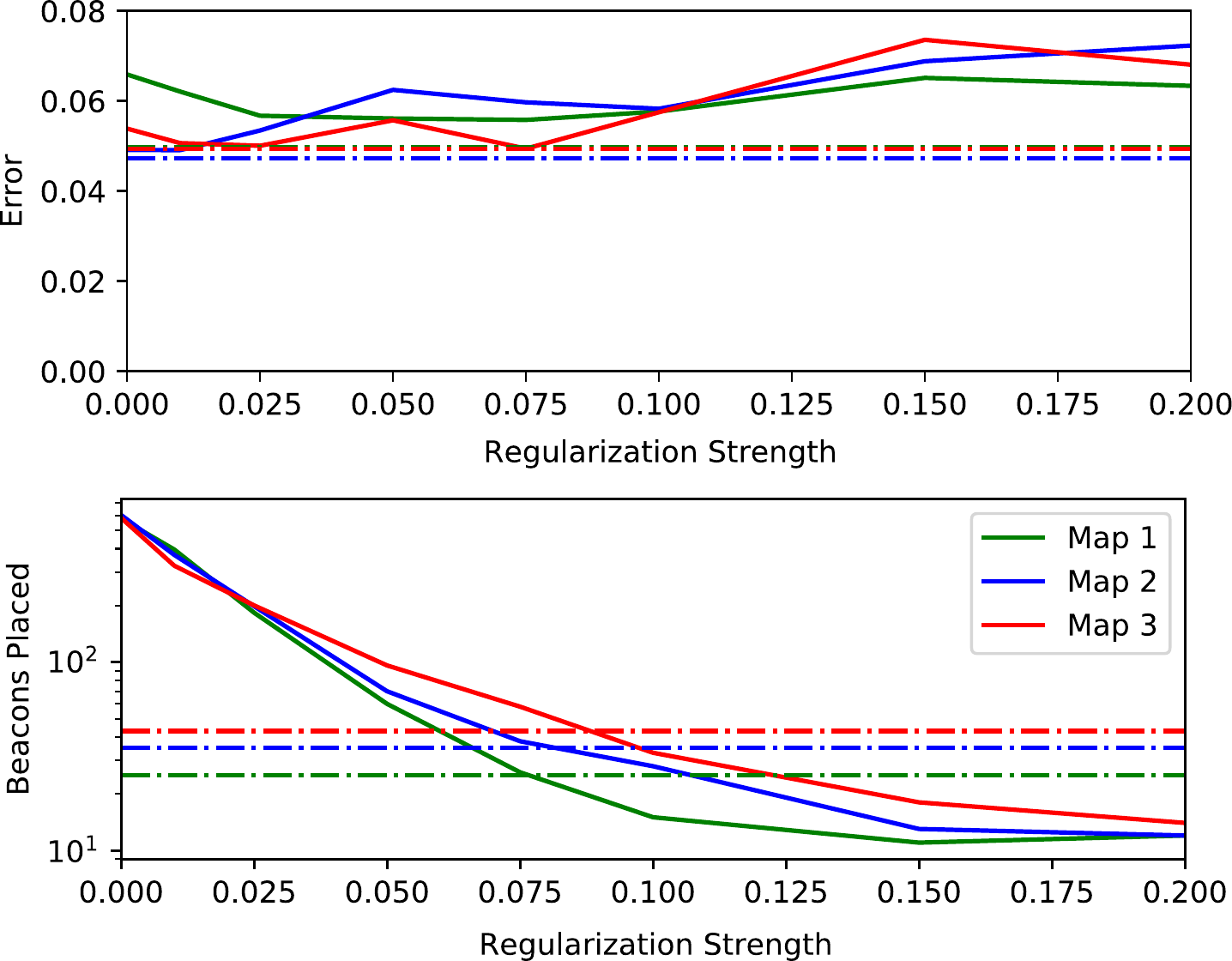}
    \caption{A plot showing the effects of regularization on mean error and the number of beacons placed. The dashed lines represent the mean error and beacons placed with annealed regularization on each map.} \label{fig:reg}
\end{figure}
Next, we report a more detailed evaluation of the regularization scheme defined in Equation~\ref{eq:regdef}, and therefore the ability of our method to allow for a trade-off between the number of beacons placed and accuracy. First, we use a constant value of $\lambda$, trying various values between $0.0$ and $0.2$. As Figure~\ref{fig:reg} shows, increased regularization leads to solutions with fewer beacons. On Map 2, we find that decreased regularization always leads to solutions with lower error. On Maps 1 and 3, however, unregularized beacon placement results in increased localization error. This suggests that regularization may also allow our model to escape bad local minima during training. Then, we experiment with an annealing scheme for $\lambda$ and find that it leads to a better performance-cost trade-off. We use a simple annealing schedule that decays $\lambda = 0.2$ by a constant factor $\eta = 0.25$ every $100$k iterations.

\subsection{Robustness to changes in signal propagation}

\begin{table}[h]
    \centering
    \setlength{\tabcolsep}{15pt}
    \caption{Behavior Under Modified Environment Settings}\label{tab:env-channels}
    \small
    \begin{tabularx}{0.8\linewidth}{l c c c}
        \toprule
        Scenario & Beacons & RMSE & RMSE (Worst-case)\\
        \midrule
        Original & $12$ & $0.0633$ & $0.1511$\\
        Low Attenuation & $\hphantom{0}8$ & $0.0426$ & $0.0910$\\
        High Noise & $27$ & $0.1400$ & $0.2907$\\
        Fewer Channels ($4$) & $11$ & $0.1290$ & $0.2581$ \\
        More Channels ($16$) & $12$ & $0.0397$ & $0.0910$ \\
        \bottomrule
    \end{tabularx}
\end{table}

Finally, we evaluate the ability of our method to automatically discover
successful placement and inference strategies for different
environmental conditions and constraints in
Table~\ref{tab:env-channels}. All results are on Map 1, with high
fixed regularization ($\lambda = 0.2$). We report results for a propagation
model with decreased attenuation at walls ($\beta = e^{-0.2}$), and
one with increased noise ($\sigma_z^2 = 2.5\times 10^{-4}$). We find
that our method adapts to these changes intuitively. Our method places
fewer beacons when the signal passes largely unattenuated through
walls and places more beacons when combating increased noise. We also
experiment with fewer ($C = 4$) and more ($C = 16$) available RF channels. As
expected, the availability of more channels allows our method to learn
a more accurate localization system. More broadly, these experiments
show that our approach can enable the automated design of location
awareness systems in diverse settings.

\section{Conclusion}

We described a novel learning-based method capable of jointly optimizing beacon allocation (both placement and channel assignment) and inference for localization tasks. Underlying our method is a neural network formulation of inference with an additional differentiable neural layer that encodes a distribution over beacon allocations. By jointly training the inference network and design layer, we automatically learn an optimal design of a location-awareness system for arbitrary environments. We evaluated our method for the task of RF-based localization and demonstrated its ability to consistently discover high-quality localization systems for a variety of environment layouts and propagation models, without expert supervision. Additionally, we presented a strategy that trades off the number of beacons placed and the achievable accuracy. While we describe our method in the context of localization, the approach generalizes to problems that involve estimating a broader class of spatial phenomena using sensor networks. A reference implementation of our algorithm is available on the project page at \url{https://home.ttic.edu/~cbschaff/nbp/}.

\flushcolsend
\clearpage 
\chapter{Optimizing Robot Design and Control} \label{sec:design_and_control}

The remainder of this thesis will shift focus from the co-optimization of sensor design and inference, to the related problem of the co-optimization of the physical design of robots along with their controllers (Figure~\ref{fig:problems}).
Similar to the dependency between beacon placement and location inference, a robot's ability to successfully interact with its environment depends both on its physical design as well as its proficiency at control. Therefore, designing a robot requires reasoning both over the mechanical elements that make up its physical structure as well the control algorithm that regulates its motion.
These physical and computational design parameters of a robot must be optimized \emph{jointly}---different physical designs enable different control strategies, and the design process involves determining the optimal combination that is best suited to the robot's target task and environment.

Consider the development of a legged robot.
Different designs will have different optimal gaits, even when the morphology is preserved. Some designs may render locomotion impossible for a target environment (e.g., a robot with heavy legs may be unable to locomote), while others may make the underlying control problem easy to solve and naturally efficient (e.g., certain bipedal designs enable passive walking~\citep{mcgeer90,goswami98,collins01}). Rather than optimizing a robot's design or gait in isolation, it is therefore beneficial to consider them together as part of a joint optimization problem.
Thus, many researchers have explored approaches that jointly reason over physical design and control~\citep{digumarti14,ha17,spielberg17}. Most recent methods are aimed at ``model-based'' approaches to control---in that they require a model of the robot dynamics or a near-ideal motion trajectory, which is chosen based on expert intuition about a specific domain and task.

Data-driven and learning-based methods, such as deep reinforcement learning, have proven effective at designing control policies for an increasing number of tasks~\citep{ibarz2021train, zhao2020sim, hua2021learning, kober2013reinforcement}. However, unlike the above methods, most learning-based approaches do not admit straightforward analyses of the effect of changes to the physical design on the training process or the performance of their policies---indeed, learning-based methods may arrive at entirely different control strategies for different designs. Thus, the only way to evaluate the quality of different physical designs is by training a controller for each---essentially treating the physical design as a ``hyper-parameter'' for optimization. However, training controllers for most applications of even reasonable complexity is time-consuming. This makes it computationally infeasible to evaluate a diverse set of designs in order to determine which is optimal, even with sophisticated methods like Bayesian optimization to inform the set of designs to explore.
\begin{figure}[!t]
    \centering
    \includegraphics[width=\textwidth]{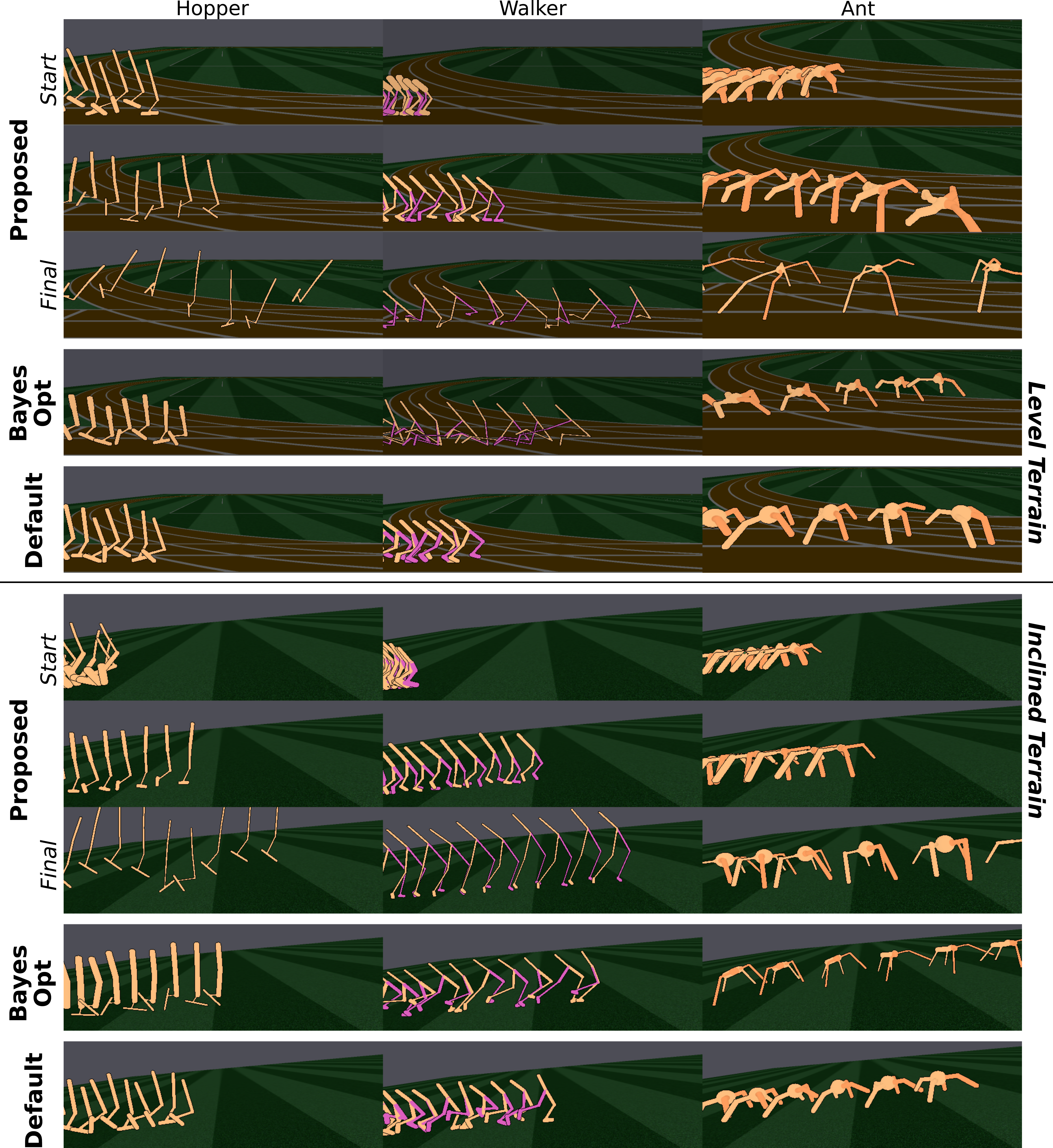}
    \caption{Our algorithm learns a robot’s physical design jointly with the control policy to optimize performance for a given task and environment. Here,
we visualize the evolution of robot design and the corresponding learned control policy (gait) for three robot morphologies tasked with
locomoting over level terrain. We also show the default Roboschool designs with their learned gaits along with those learned
using Bayesian optimization. Images are captured at fixed time intervals, so designs appearing farther to the right represent faster motion.}
    \label{fig:design_and_control_results}
\end{figure}

In this work, we seek to alleviate these limitations by introducing an efficient algorithm that jointly optimizes over both physical design and control. Our approach maintains a distribution over designs and uses reinforcement learning to optimize a neural network control policy to maximize expected reward over the design distribution. We give the controller access to design parameters to allow it to tailor its policy to each design in the distribution. Throughout training, we shift the distribution towards higher performing designs and continue to train the controller to track the design distribution. In this way, our approach converges to a design and control policy that are jointly optimal. Figure~\ref{fig:design_and_control_results} visualizes this evolution for different robot morphologies.

Our approach can be applied to arbitrary morphologies, tasks, and environments, and explores the joint space of design and control in a purely data-driven fashion. We evaluate our method in the context of legged locomotion, parameterizing the length and radius of links for several different robot morphologies. Experimental results show that starting from random initializations, our approach consistently finds novel designs and gaits that exceed the performance of manually designed agents and other optimization baselines, across different morphologies and environments.%

\section{Related Work}

Co-design of physical structure and control has a long history in robotics research. 
A large number existing approaches~\citep{park94,pil96,de02,reyer02,paul06,villarreal13} are model-based---in that they rely on having a model of environment dynamics to solve for both the physical design and the control policy. Many focus on co-design with specific types of controllers, e.g., \citet{ha17} model both design and motion parameters via a set of implicit functions that express robot dynamics and actuation limits with a desired trajectory in mind, and carry out optimization on linearized approximations of these functions. Others are designed for settings in which control is formulated as a trajectory optimization problem~\cite{mordatch15, dai16, posa16, griffin16}. For example, \citet{spielberg17} solve for the design of articulated robots jointly with trajectory parameters (e.g., contact forces and torques), with the requirement that the problem be initialized within a small neighborhood of a feasible solution, and that the user provide an estimate of the robot configuration at each time step. Also worth noting are methods that find robot designs to meet specified task requirements---e.g., given user demonstrations of desired behaviors, \citet{coros13} learn optimal kinematic linkages capable of reproducing these motions, while \citet{mehta14} synthesize electro-mechanical designs in a compositional fashion based upon a complete, user-specified structural specification of the robot. Related, \citet{censi17} describes a theoretical framework that allows one to select discrete robot parts according to functional constraints, but does not reason over geometry or motion.

Another class of approaches to this problem are based on evolutionary search (ES) over a population of candidate designs~\citep{sims1994,lipson00,paul01,watson2002embodied,pfeifer2006body,murata2007self,wampler2009optimal,hiller2011automatic,bongard11,bongard2013evolutionary,cheney2014unshackling,brodbeck2015morphological,cheney2018scalable,pan2021emergent}.
These methods iteratively generate the design population in a stochastic manner by joining or modifying existing designs, solving for their individual controllers, and then evaluating their performance.
However, ES is prone to local minima and is sample-inefficient, in part due to the need to maintain separate control policies for each candidate design in the population. Our approach improves upon the sample-efficiency of these methods through deep reinforcement learning techniques and sharing experience across designs by training a single, universal control policy.

For many applications and domains, control policies modeled as neural networks and trained with deep reinforcement learning have emerged as successful approaches that deliver state-of-the-art performance~\citep{ibarz2021train, zhao2020sim, hua2021learning, kober2013reinforcement}.
The ability of such controllers to successfully learn complex policies directly from low-level sensory input and without any expert supervision is clearly attractive. However, being purely data-driven and trained through a complex iterative process, the resulting control policies make it non-trivial to jointly optimize these policies with physical structure.

Early approaches to the co-design of neural controllers with physical designs~\cite{sims1994,lipson00,paul01,bongard11} employ evolutionary methods, albeit to restricted settings (e.g., \citet{bongard11} only optimizes over quadrupedal foothold patterns). Meanwhile, when dealing with deep neural networks that require a significant amount of computational expense to train, Bayesian optimization~\cite{bayesopt} has emerged as a successful strategy for the optimization of external (hyper-)parameters. This provides a possible strategy for learning physical designs by treating design specifications as hyper-parameters. However, Bayesian optimization also becomes computationally expensive and infeasible when applied to high-dimensional search spaces arising out of a large number of hyper-parameters.

Motivated by our success of the joint training of neural network-based estimators with sensor parameters for beacon based localization (Chapter~\ref{sec:beacons}), we seek to extend neural co-optimization methods to the optimization of design and control.
However, since our loss and measurement functions are differentiable, past work is able to rely on gradient-based updates, both for training the location estimators and for computing gradients with respect to the beacon parameters. 
In this work, we propose a co-optimization framework that trains both the physical design and a control policy network in a reinforcement learning setting based on rewards returned by the environment.

While co-optimization methods based on reinforcement learning were largely unexplored when the work in this chapter was developed, there have since been a large number of proposed methods~\cite{ha_reinforcement_2019, luck_data-efficient_2019, chen_hardware_2020, spielberg2019learning, pathak2019learning, yuan2021transform2act}.
For example, \citet{luck_data-efficient_2019} use a soft actor-critic algorithm and use a design-conditioned Q-function to evaluate designs and \citet{yuan2021transform2act} represent the design generation process as part of the environment and train a dual-purpose design generation and control policy.

\section{Preliminaries}

The methods described in this Chapter assume a basic knowledge of reinforcement learning, policy gradients, and PPO~\cite{schulman17}. This section is meant to give the reader a brief introduction to these topics.

The problem of controlling an agent can be modeled as a continuous Markov decision process (MDP), denoted by the tuple $\{\mS,\mA,p,r, p_0\}$, where $\mS \subseteq \mathbb{R}^d$ is the state space, $\mA \subseteq \mathbb{R}^n$ the action space, $p$ the (unknown) transition model between states, $r$ the reward function, and $p_0$ the initial state distribution. An agent in state $s_t \in \mS$ at time $t$ takes action $a_t \in \mA$ and the environment returns the agent's new state $s_{t+1}$ according to the unknown transition function $p(s_{t+1} \vert s_t,a_t)$, along with the associated reward $r_t = r(s_t, a_t)$. A policy is a mapping $\pi: \mS \rightarrow \mA$ from states to actions that defines how the agent should act in every state $s \in \mS$. 
The goal of reinforcement learning is to find the optimal control policy $\pi^*$ that maximizes the expected return $\mathbb{E}_\pi[\mR_t]$, which takes into account rewards at future time-steps  $\mR_t = \sum_{i=0}^\infty \gamma^i r_{t+i}$ with a discount factor $\gamma \in [0,1)$ that lowers the value of future rewards.

For an increasing number of reinforcement learning problems, a stochastic policy $\pi_\theta(a_t \vert s_t)$ is modeled as a neural network with input $s_t$ and output $a_t$. In the case of complex continuous action spaces, policy gradient methods are commonly used to learn the parameters $\theta$ through stochastic gradient ascent on the expected return.
``Vanilla'' policy gradient methods compute an estimate of the gradient $\nabla_\theta \mathbb{E}[\mR_t]$ using a sample-based mean computed over $\nabla_\theta \log \pi_\theta(a_t \vert s_t)\mR_t$~\citep{williams92}, which yields an unbiased gradient estimate~\citep{sutton00}.
Recently, methods have been proposed to improve the stability of the learning process~\cite{schulman15,schulman17}. Notably, proximal policy optimization (PPO)~\cite{schulman17} is a first-order class of methods that alternates between sampling data from the environment and optimizing the objective
\begin{equation}
    \hat{\mathbb{E}}_t\left[\min(r_t(\theta)\hat{A}_t, \text{clip}(r_t(\theta), 1-\epsilon, 1+\epsilon)\hat{A}_t)\right],
\end{equation}
where $r_t(\theta) = \frac{\pi_\theta(a_t \vert s_t)}{\pi_{\theta_{\text{old}}}(a_t \vert s_t)}$ and $\hat{\mathbb{E}}_t$ represents the average over a limited sample set. This clipped objective has the effect of maximizing expected return by making only small steps in policy space at a time.
This yields a simple yet robust reinforcement learning algorithm that attains state-of-the-art results on a wide array of tasks~\citep{schulman17}.

\section{Approach}
In this section we first introduce the problem formulation for jointly optimizing design and control, then discuss our proposed solution.

\subsection{Problem definition}

We formulate the problem of co-optimizing design and control as a set of related reinforcement learning problems.
Given a set of robot designs $\Omega$ and a task definition, we define a \textit{design-specific} Markov decision process (MDP) for each $\omega \in \Omega$: $\mM_\omega = \textrm{MDP}(\mS_\omega,
\mA_\omega, \mP_\omega, \mR_\omega)$, where $\mS_\omega$ is the state space, $\mA_\omega$ is the action space, $\mP_\omega: \mS_\omega \times \mA_\omega \times \mS_\omega \rightarrow [0,1]$ is the transition dynamics, and $\mR_\omega: \mS_\omega \times \mA_\omega \times \mS_\omega \rightarrow \mathbb{R}$ is the reward function.
When optimizing for a single task objective, it is natural that these MPDs share some common structure, e.g., the state space only changes based on the proprioceptive information available to each design, the action spaces differ only in the number of controllable degrees of freedom, and each reward $\mR_\omega$ encodes the same task objective.
In this work, we only consider design spaces that consist of robots within a single morphology. Becuase of this, all designs in $\Omega$ have a single, shared state space $\mS$ and action space $\mA$.

Let $\pi^*_\omega: \mS \times \mA \rightarrow [0, 1]$ be the optimal policy for MDP $\mM_\omega$ and $\mathbb{E}_{\pi^*_\omega}\left[\sum_t \gamma^t r_t\right]$ be its expected return. The goal of co-optimization is to find the optimal design-controller pair $(\omega^*$,~$\pi^*_{\omega^*})$ such that:
\begin{equation}
    \label{eqn:coopt_objective}
    \omega^*, \pi^*_{\omega^*}  = \underset{\omega, \pi^*_\omega}{\text{arg max}} \; \mathbb{E}_{\pi^*_\omega}\left[\sum_t \gamma^t r_t \right].
\end{equation}

\section{Co-optimization of design and control via multi-task RL}

In this problem, we are faced with many MDPs that share common structure. Solving each MDP independently is intractable and ignores these similarities, which are critical to generalizing to new designs. %
We draw on insights from multi-task reinforcement learning~\cite{Varghese2020} to more efficiently solve for the optimal design-control pairs (Equation~\ref{eqn:coopt_objective}) by exploiting this common structure.
Similar to goal-conditioned policies, our approach learns a single \textit{design-conditioned} policy $\pi_\theta: \mS \times \mA \times \Omega \rightarrow [0, 1]$ to control all the designs in $\Omega$ for the specified task.
This policy can be trained using any RL algorithm on a mixture of data collected with designs in $\Omega$.

\begin{figure}[t]
    \centering
    \includegraphics[width=1.0\textwidth]{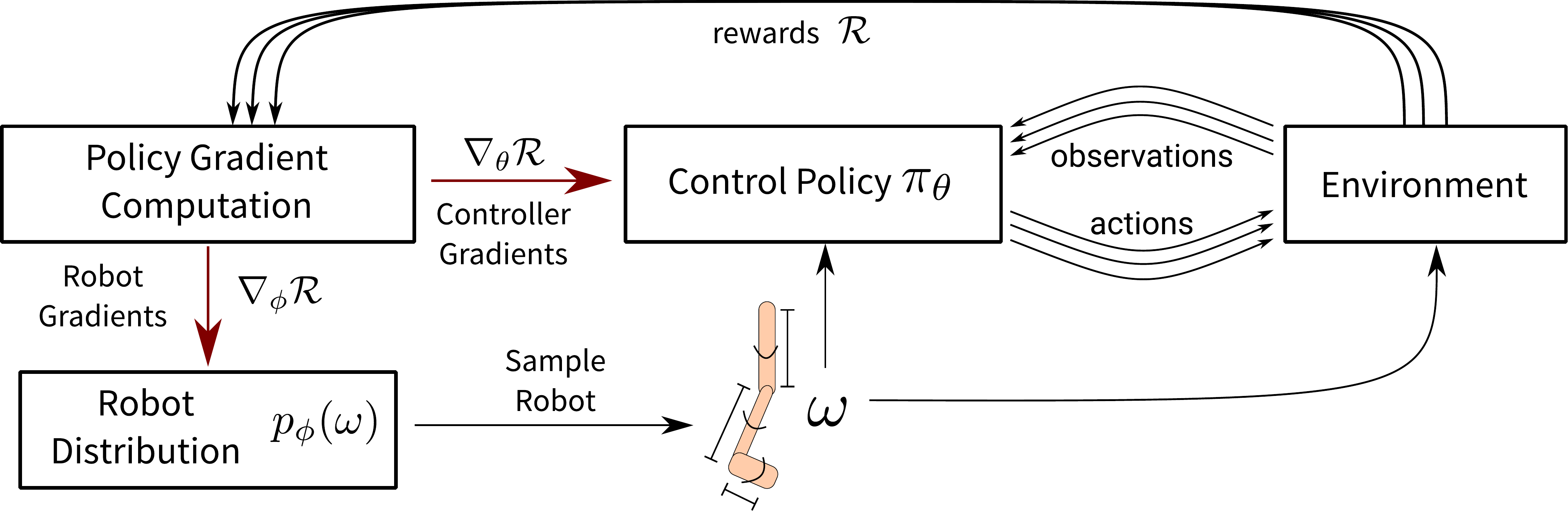}
    \caption{Our proposed algorithm maintains a distribution $p_\phi(\omega)$ over possible
physical design parameters. At each iteration of training, a specific design $\omega$
is sampled from this distribution, and an episode is run with that design and
the current control policy $\pi_\theta$ in the environment. The policy $\pi_\theta$ has access
to the parameters $\omega$ of the robot instance it is controlling in that episode,
along with observations fed back from the environment. Policy gradients
with respect to episode rewards are then used to update parameters $\theta$ and $\phi$
of both the control policy and the robot distribution, respectively.}
    \label{fig:design_and_control_alg}
\end{figure}

We propose a framework for finding the optimal design-control pair that, in addition to the policy distribution $\pi_\theta$, maintains a distribution $p_\phi(\omega)$ over possible designs $\omega\in\Omega$. The learnable parameters $\phi$ encode the framework's belief of which designs are likely to be successful. The policy function $\pi_\theta(a_t \vert s_t, \omega)$ is provided with the parameters $\omega$ of the design it is controlling and trained to be successful with not just a single design, but in expectation over the design distribution $p_\phi(\omega)$.
Moreover, since we provide the policy $\pi_\theta$ with design parameters, it is not constrained to use the same strategy for all designs and is able to tailor control to each design.

We propose an algorithm  (summarized in Algorithm~\ref{algorithm}) that carries out this optimization to maximize the expected reward obtained by the policy $\pi_\theta$ over the design distribution $p_\phi(\omega)$.
At each iteration of training, the policy is trained (using Proximal Policy Optimization (PPO)~\cite{schulman17}) with respect to gradients that maximize the expected return over designs sampled from the current design distribution $p_\phi(\omega)$. At the same time, the design distribution is updated every iteration to increase the probability density around designs that perform well when using the current learned policy $\pi_\theta$. This update is computed w.r.t. the following gradient computation:
\begin{equation}
    \label{design_update}
    \nabla_\phi \mathbb{E}_{\omega \sim p_\phi}\bigl[\mathbb{E}_{\pi_\theta}\left[\mR_t\right]\bigr] = \mathbb{E}_{\omega \sim p_\phi}\bigl[\nabla_\phi \log p_\phi(\omega)\mathbb{E}_{\pi_\theta}\left[\mR_t\right]\bigr].
\end{equation}
This update shifts the parameters of the design distribution $\phi$ to maximize the expected reward under the current policy $\pi_\theta$, and is analogous to gradient-based updates to the policy parameters~\cite{sehnke10}, except that the choice of $\phi$ now affects the transition dynamics. We parameterize the design distribution $p_\phi(\omega)$ as a Gaussian mixture model, which allows our framework to maintain multiple distinct hypotheses for designs it believes to be promising. This allows greater exploration of the physical design space $\Omega$ and helps prevent the optimization from collapsing to local optima.
\begin{algorithm}[t]
  \caption{Joint Optimization of Design and Control} \label{algorithm}
  \begin{algorithmic}
      \STATE Initialize $\pi_\theta(a \vert s,\omega)$, $p_\phi(\omega)$, $T=0$, $T_0$
      \WHILE{$T < $ BUDGET}
        \STATE Sample design $\omega \sim p_\phi$
        \STATE Control design $\omega$ with $\pi_\theta$ for $t$ timesteps, collecting trajectory $\tau = \{s_1, a_1, r_1, \ldots, s_{t-1}, a_{t-1}, r_{t-1}, s_t\}$
        \STATE Update $\theta$ using PPO and $\tau$.
        \STATE $T = T + t$
        \IF{$T > T_0$}
            \STATE Compute the average episode return $R$ of $\tau$
            \STATE Update $\phi$ using $\nabla \phi \approx \nabla \text{log} p_\phi(\omega)R$
        \ENDIF
      \ENDWHILE
  \end{algorithmic}
\end{algorithm}

At the beginning of the optimization process, the design distribution is initialized to have high entropy---with each component initialized with random means and high variance. Consequently, the policy initially learns to control a diverse set of designs. There is a warm-up period for the policy network at the beginning of training, where only the policy parameters $\theta$ are updated while the design parameters $\phi$ are kept fixed. Once the method begins to update the design distribution, the gradient updates proceed to eliminate designs that perform poorly.
At the end of training, the design $\omega$ is fixed to the mode of the design distribution $p_\phi(\omega)$, and the policy network is fine-tuned for a few more iterations with this fixed design. This procedure yields the final estimates for the design $\omega^*$ and policy $\pi^*_\theta$.

\section{Experiments}

We evaluate our framework on the problem of jointly learning the physical design and control policy of legged robots tasked with locomoting within different environments. We consider three commonly used robot morphologies included in OpenAI's Roboschool~\citep{schulman17}: the Hopper, Walker (referred to as Walker2D in Roboschool), and Ant. We optimize both with the goal of maximizing a standard reward function, and find that our approach discovers novel robot designs and gaits that outperform controllers learned for the standard Roboschool design, as well as design-controller pairs learned through random sampling and Bayesian optimization. Source code and a video highlighting our results can be found at \url{http://ttic.uchicago.edu/~cbschaff/nlimb}.

\subsection{Experiment Setup}

We consider the task of locomotion for the Hopper, Walker, and Ant morphologies on both a level and inclined ground plane (with a five degree slope). The environments are built on top of Bullet Physics, a popular open-source physics engine. We use the Roboschool default reward function, which is a weighted sum of rewards for forward progress and staying upright, and penalties for torques and for reaching joint limits. Every episode ends when a robot falls over or after a maximum number of timesteps.

We restrict our method and the baselines to optimize over robot designs that adhere to the default Roboschool morphologies. We parameterize each morphology in terms of the length and radius (e.g., mass) of each link, and the spherical body radius for the ant. We impose symmetry over both legs of the Walker, but optimize over independent design parameters for each of the four legs of the Ant. Consequently, the Hopper and Walker both have eight learnable parameters, while the Ant has twenty-five. We limit the values of each parameter to lie within a reasonable range (as a proxy for fabrication constraints). The resulting search space includes a wide variety of robots with different shapes and sizes, including many that are clearly impractical for locomotion. However, rather than providing a reduced search space guided by expert intuition, we let our framework discover and reject such designs automatically. %

The MDP state $s_t$ is comprised of joint angles and velocities, center-of-mass velocity, height of the torso, and direction to the target. For all experiments, we model the control policy $\pi_{\theta}(a_t \vert s_t, \omega)$ as a feed-forward neural network that consists of three fully-connected layers with $128$ units and $\tanh$ activation functions. A final layer maps the output to the robot action space---a vector of joint torques. For our framework, we also append the design parameters $\omega$ to the state variables before passing them to the control network.

For our framework, we represent the distribution over robot designs $p_{\phi}(\omega)$ as a GMM with eight mixture components, each with a diagonal covariance matrix. We randomly initialize the means of each component and set the variances such that the distribution spans the parameter ranges. The mixing probabilities are set to be uniform---across all eight components at the start of training, and across the remaining components after low reward components are removed every $N$ timesteps. Thus, only the component means and variances are updated based on policy gradients during training.

We train our model in parallel on eight robot samples and updates the policy $\pi_{\theta}(a_t \vert s_t, \omega)$ with PPO, for a total of $1$B environment timesteps. We train the controller for $100$M timesteps before updating the design distribution. In the case of the Ant, which takes longer to train, we train for a total of $1.5$B timesteps and start updating the design distribution after $200$M timesteps. Additionally, we evaluate and prune components every $N=100$M ($200$M for the Ant) timesteps. For the last $100$M timesteps, we choose the mode of $p_\phi$ as the optimal design $\omega^*$ and train the policy with this fixed design.

\subsection{Baselines}

In addition to reporting results from our framework (run eight times for each morphology and environment to gauge consistency), we evaluate and compare it to three baselines:

\begin{itemize}
\item \textbf{Default designs:} This baseline simply involves training a control policy for the standard, hand-crafted Roboschool designs. We use the same policy architecture as with our method and train until convergence.

\item \textbf{Random sampling:} This baseline samples designs uniformly at random within the parameter ranges, and trains a separate control policy for each sample. We use the same policy architecture as with our method and train until convergence. Since this approach is inherently more parallelizable than our method, we allow this baseline to sample designs and train control policies for three times the number of timesteps used by our method. The reward of the best design-control pair found across samples can then be compared to the performance of a single run of our experiment.

\item \textbf{Bayesian optimization:} This baseline employs Bayesian optimization (BayesOpt) to search jointly over the physical design space, using a publicly available implementation~\cite{gpyopt2016}. For each sampled design, we train a control policy until convergence and return the average episode reward to the BayesOpt routine. BayesOpt then samples a new design and the process repeats. We use the same policy architecture as our experiments and allow BayesOpt the same number of environment timesteps used by our method. However, this timestep limit only allows for roughly ten designs to be evaluated. To account for variance in outcomes, we run four copies of this baseline with different random seeds and report each outcome.

\end{itemize}

\section{Results}

\begin{figure*}[!t]
    \centering
    \includegraphics[width=\textwidth]{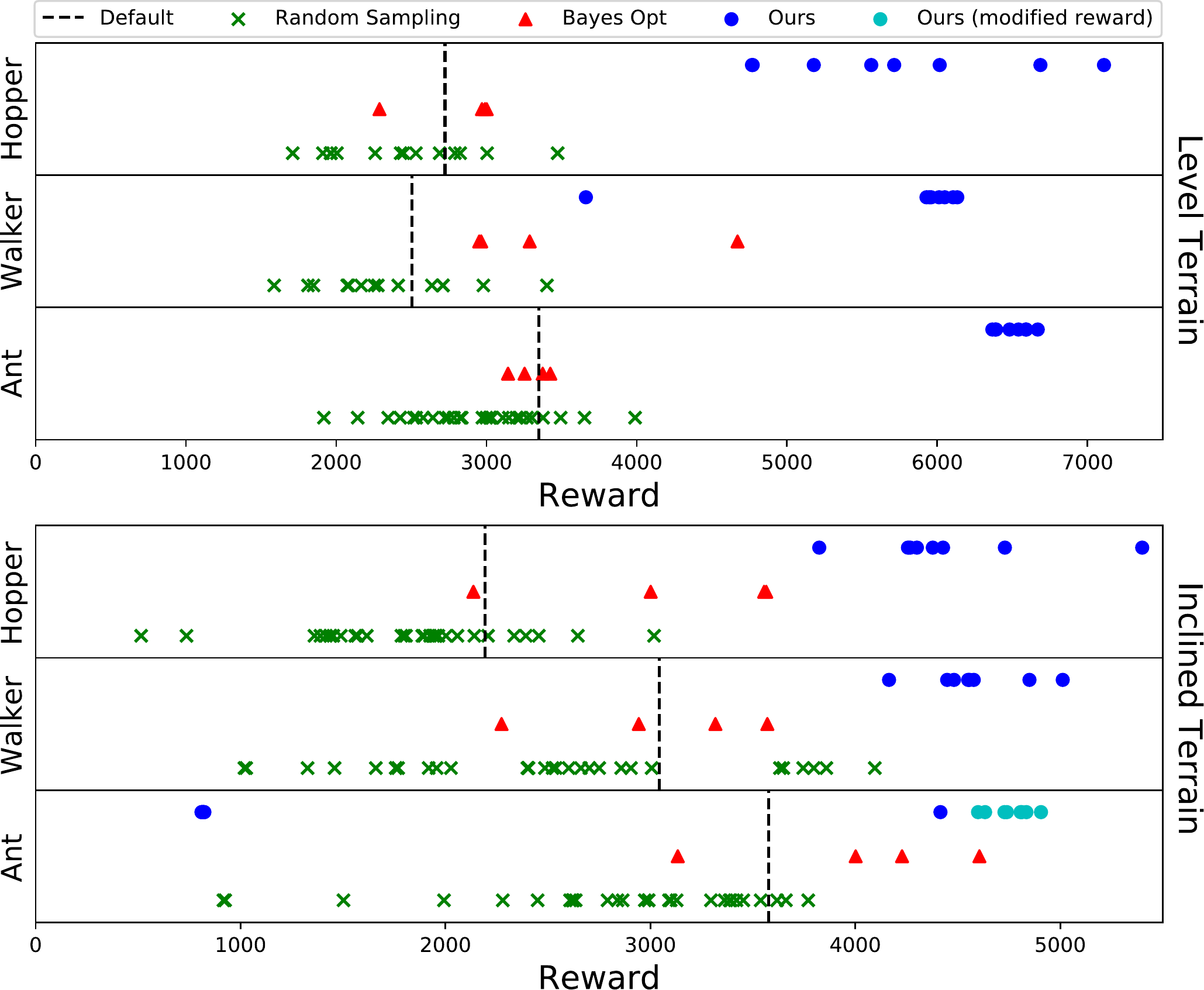}
    \caption{A comparison of the rewards achieved for different seeds with our method and the three baselines for level (left) and  inclined (right) terrains. Performance is measured as the average reward over $100$ episodes at the end of training. The cyan dots for the Ant on inclined terrain represent our performance after modifying the reward function for training, but their position indicates performance according to the original reward function.}
    \label{fig:design_and_control_rewards}
\end{figure*}

Figure~\ref{fig:design_and_control_results} presents the physical designs and gaits learned by our method and BayesOpt, as well as the gait learned for the fixed design. For BayesOpt and our method, we show the best performing seed. To illustrate the evolution of the design and control policy we also visualize results from near the beginning and middle of training. Figure~\ref{fig:design_and_control_rewards} provides a quantitative analysis of the rewards achieved by all seeds of the different methods.

\subsection{Performance on Level Terrain} We find that the joint design-policy pair found by our framework exceeds the performance of all baselines for all three morphologies on level terrain. Our method obtains these performance levels by discovering unique robot designs together with novel walking gaits. Note that our method learns these designs and control policies from \emph{random initializations}, without access to a dynamics model, expert supervision, or good initial solutions.

For the Hopper, there is some variance in performance among the seeds for our method, but every seed outperforms all seeds of the different baselines. We find that our learned Hopper designs all have long bodies and generally use one of two gaits. Designs with a heavier body have short, quick strides, while designs with a narrow, light body have large, bounding gaits. For both groups, the longer torso of the learned robot improves stability, allowing it to maintain balance while locomoting at a faster pace.

For the Walker, all but one of our seeds perform significantly better than the baselines. The seven better-performing designs from our framework all have long thin legs and small feet, and all achieve a running gait with long strides. The last seed learns a design with long feet and even longer legs, and learns a stable walking gait on its knees. %

Our method achieves consistent results for the Ant morphology, all of which do better than the baselines by a significant margin. The learned design has a small, lightweight body and thin legs. Despite not sharing parameters between legs, our method consistently finds a solution in which each leg is roughly the same size and shape. The learned gait primarily uses three legs (one on each side and one behind) to walk and generally keeps the front leg off the ground, occasionally using it for balance.

\begin{figure}[t]
    \centering
    \includegraphics[width=\linewidth]{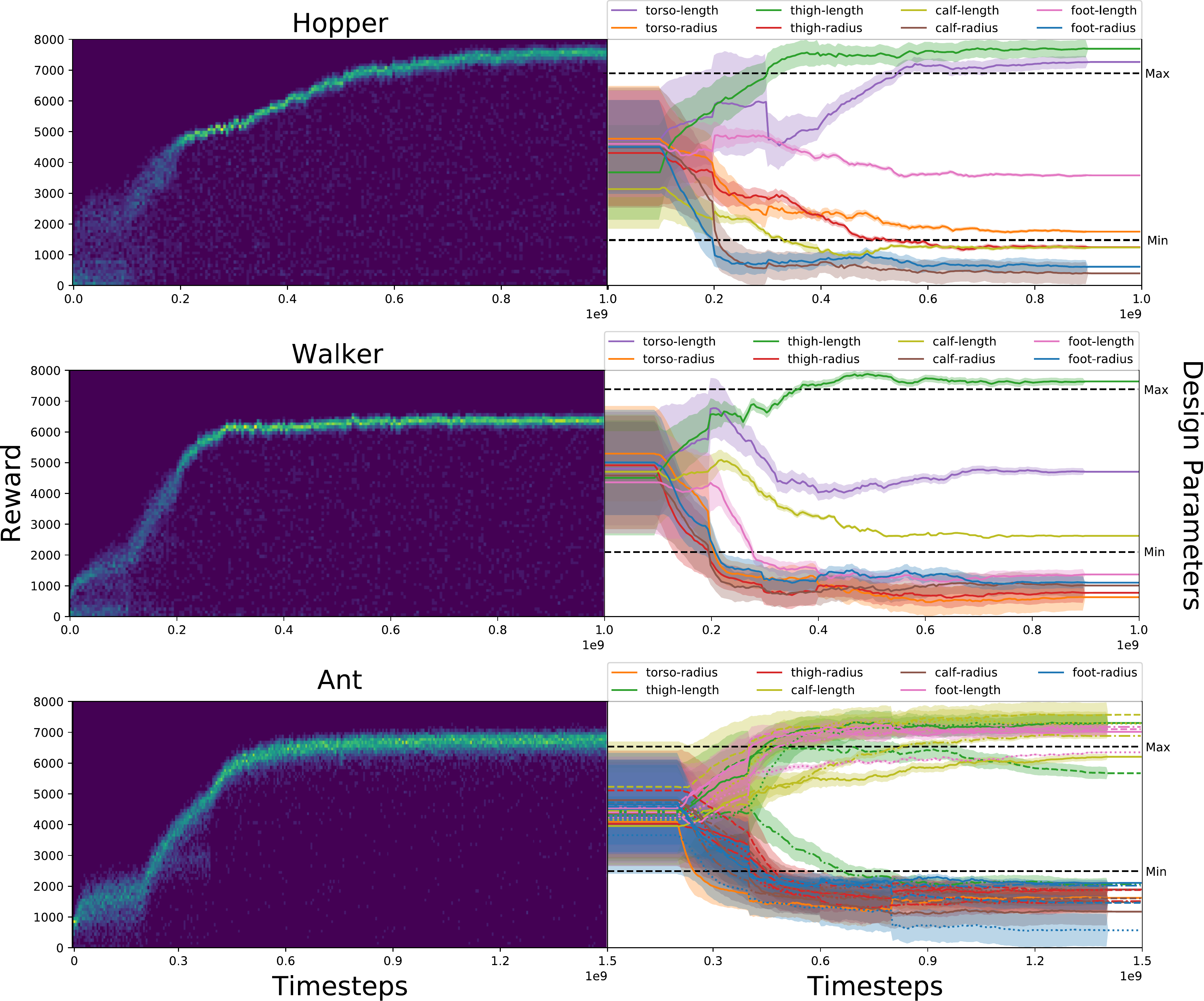}
    \caption{Histograms that show the reward evolution (top) for sampled designs of the Hopper, Walker, and Ant, along with the evolution of the mean and variance of the design distribution (bottom) during training. Every $5$\,M timesteps, we sample $100$ robots from our current design distribution and plot the distribution of rewards obtained by those robots over the course of an episode. In the plot showing design parameters for the Ant, colors represent the different parts of each leg and line styles represent individual legs.} \label{fig:histogram-params-flat}
\end{figure}

\subsection{Performance on Inclined Terrain} On the inclined terrain, our method again outperforms the baselines for the Hopper and Walker morphologies.
However, standard training with the Ant on inclined terrain represents an interesting failure case for our method. While one of our seeds outperforms all but one baseline seed, for the remaining seven, the optimization remains stuck at a local optimum with very low reward---corresponding to the Ant just balancing itself on the incline without moving forward. We believe this is the case because the control network is unable to make a significant portion of the sampled designs move forward during the initial warm-up period (when the design is fixed), and chooses the ``safer'' option of having all samples stand still. The optimization is never able to move away from this local optimum once the design distribution starts updating, moving towards large designs that balance using little energy, making locomotion even harder.

While an interesting direction for future work is finding a more general solution for avoiding local optima, we find that a simple modification to the reward function suffices in this case. During training, we reduce the reward for staying upright by half, thus further incentivizing forward motion. We include these results in Figure~\ref{fig:design_and_control_rewards} (the result included in Figure~\ref{fig:design_and_control_results} corresponds to the best seed trained with this modified reward). With this modification, all eight seeds are able to escape the standing local minima, and all outperform the baselines---with performance measured in terms of the original reward. (However, we note that a similar modification during training may also help the baselines.)

\subsection{Visualizing the Training Process}

To better understand the training process and the evolution of our learned designs, we provide a histogram of episode rewards throughout training, along with the marginal distributions for each design parameter, for different runs of our experiments (Fig.~\ref{fig:histogram-params-flat}). The histogram is generated as follows: every $5$M timesteps during training, we sample $100$ designs from the design distribution and report the episode reward of each design under the current control policy. Figure~\ref{fig:histogram-params-flat} shows that our approach maintains high variance the in design parameters early in training, then slowly converges to a narrow distribution before being fixed at the mode of the distribution at the end of training. Not surprisingly, the evolution of the reward histograms closely mimic changes to the means and variances of the design distributions. There are clear shifts in the design distributions and the reward histograms when components are evaluated and pruned.

\section{Conclusion}

We proposed a model-free algorithm that jointly optimizes over a robot's physical design and the corresponding control policy, without the need for any expert supervision. Given an arbitrary morphology, our method is able to derive novel design-control pairs that outperform Bayesian optimization and random sampling on a set of locomotion tasks.

Our findings suggest several avenues for future work. The most direct is extending the current approach to find optimized designs for a larger variety of tasks, such as locomotion in the presence of uneven terrain, obstacles, variations in friction, etc; and other domains such as manipulation. We are also interested in extending our framework to relax the assumption that the morphology is fixed. This would require a more complicated policy capable of handling different action spaces for each morphology, as well as a distribution over the discrete set of morphologies and their corresponding continuous parameters.
Another avenue for future work is to fabricate the learned design and then transfer the control policy from simulation to the real robot. Recent work by \citet{tan18} has shown this to be possible in the context of legged locomotion with the help of domain randomization. 
We explore this line of work in the context of soft robotics in Chapter~\ref{sec:soft_design}.

\flushcolsend
\clearpage 
\chapter{Optimizing over Robot Morphology}\label{sec:nlimb_paper}

\begin{figure}[h]
    \centering
    \includegraphics[width=0.9\linewidth]{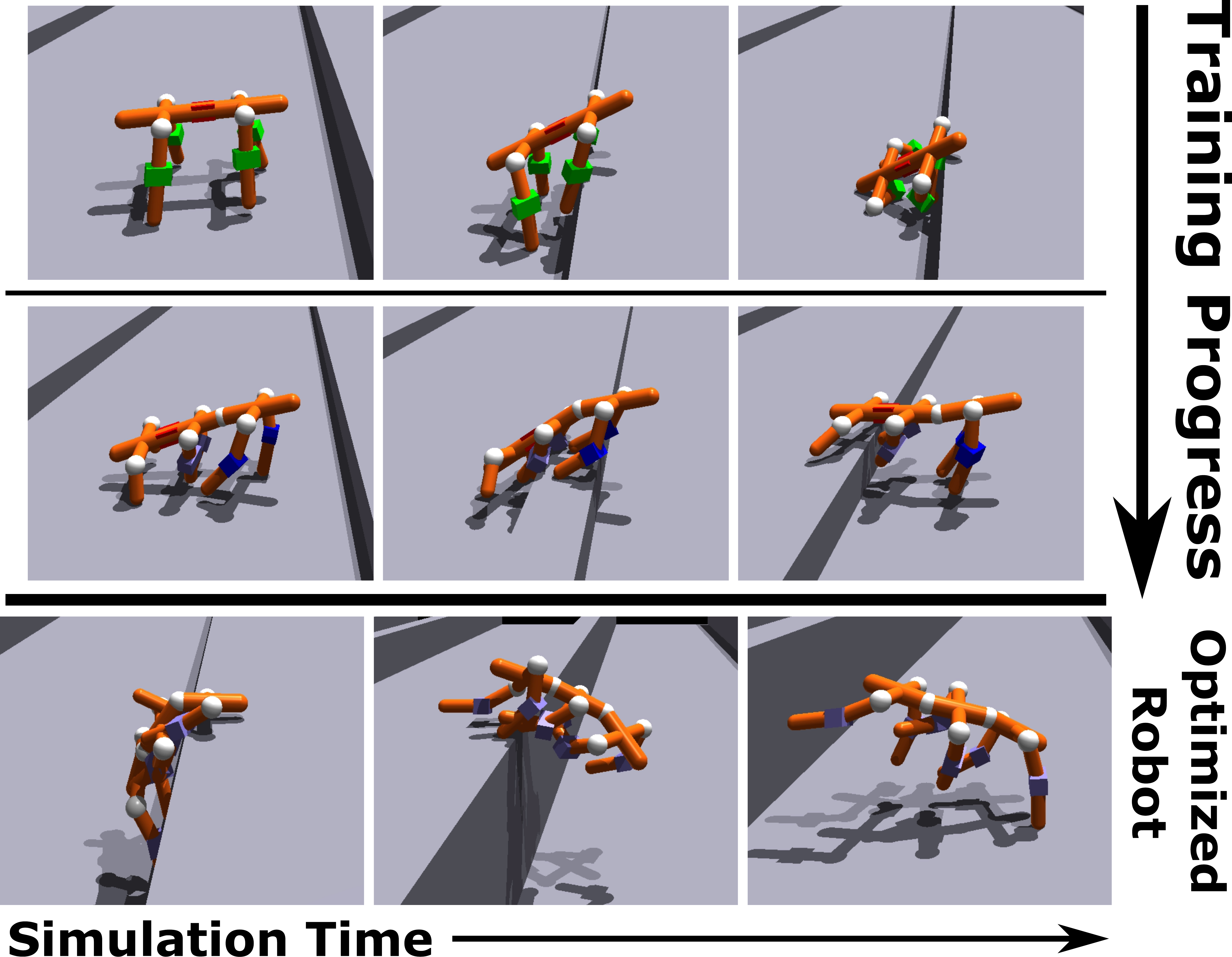}
    \caption{Our approach produces optimized robots and controllers for a given task by searching over large sets of morphologies. This figure shows the optimization of a wall-climbing robot from early in the training process (top row) to the final design (bottom row).}\label{fig:nlimb_motivation}
\end{figure}
In the previous chapter we produced optimized and high-performing robot-control pairs for a specified task, but the search over robots was restricted to a fixed set of continuous design parameters for a single, hand-picked morphology.
The morphology of a robot consists of the number and arrangement of body parts and the connections between them that define the structure of the robot, and restricting the optimization to a single morphology was a simplifying assumption that allowed us to more easily train a universal controller.
However, this assumption limits the applicability of the proposed approach because it forces the user to provide a morphology that is well suited for the task, which can be challenging.
This chapter alleviates this assumption, allowing for the optimization over discrete, morphological choices such as the number and arrangement of limbs and the controllable degrees of freedom that connect them.

However, the search over discrete morphologies and their controllers is inherently challenging. The optimization process is two-fold, requiring: 1) a search over a prohibitively large, discrete, and combinatorial space of morphologies; and 2) the evaluation of each examined morphology by solving for its optimal controller.
The second step is particularly non-trivial and computationally expensive, especially for learned controllers, which makes it challenging to examine a large set of morphologies.

We present Neural Limb Optimization (\nlimb), an efficient algorithm for computing optimal design-control pairs across a large space of morphologies (Figure~\ref{fig:nlimb_motivation}). \nlimb is a non-trivial extension to the approach presented in Chapter~\ref{sec:design_and_control}, which provides a model-free approach to co-optimization over continuous design parameters, together with a learned controller \textit{for robots with a single fixed morphology}.
Our aforementioned approach exploits the shared structure of each design to train a universal, design-conditioned controller that is able to generalize across the space of valid designs, thus avoiding the need to solve for an optimal control policy from scratch for each design.
This controller is trained in expectation over a distribution of designs that is then subsequently updated towards higher-performing designs.
Adapting these components (i.e., a universal controller and design distribution) to a large set of morphologies is challenging as it requires handling discrete and combinatorial design spaces as well as state and action spaces that vary across morphology.
In fact, developing universal controllers that generalize across morphologies has emerged as an important topic independent of morphological optimization~\cite{Huang2020, Kurin2021, gupta2022metamorph}. 

Our \nlimb architecture overcomes these challenges to jointly optimize robot design and control over a large space of morphologies through three key contributions.
First, our approach represents the space of morphologies with a context-free graph grammar.
This has two benefits: it allows for the easy incorporation of fabrication constraints and inductive biases into the symbols and expansion rules of the grammar; and it provides a way to iteratively generate designs by sampling expansion rules.
Second, this allows us to define complex, multi-modal distributions over the space of morphologies via a novel autoregressive model that recursively applies expansion rules until a complete graph has been formed.
And third, we parameterize the universal controller with a morphologically-aware transformer architecture.
We then evaluate the effectiveness of \nlimb on a series of challenging locomotion tasks, demonstrating the ability to learn complex, high-performing designs coupled with agile control policies.
For videos and code, please visit our website: \href{https://sites.google.com/ttic.edu/nlimb}{https://sites.google.com/ttic.edu/nlimb}. 
\section{Related Work} \label{sec:nlimb_related_work}

\paragraph{Co-optimization of robot design and control:} There exists a large body of work that addresses the problem of optimizing the physical design of a robot along with the corresponding control policy.
Initial research on co-optimization is concerned with identifying the discrete morphology %
of high-performing robots~\citep{sims1994, lipson00, murata2007self}. This morphology may take the form of a composition of fixed-size 3D blocks~\citep{sims1994} or deformable voxels~\citep{cheney2014unshackling, cheney2018scalable, spielberg2019learning}, or as a topology of rigid parts connected by fixed or actuated joints~\citep{hornby2003generative, desai2018interactive, ha2018computational, pathak2019learning, wang2019neural,zhao2020robogrammar,hejna2021task,xu2021multi}, %
which is the approach that we take here. %

A common approach to optimizing over discrete designs %
is to perform evolutionary search (ES) over a population of candidate designs~\citep{sims1994,lipson00,paul01,watson2002embodied,pfeifer2006body,murata2007self,wampler2009optimal,hiller2011automatic,bongard11,bongard2013evolutionary,cheney2014unshackling,brodbeck2015morphological,cheney2018scalable,pan2021emergent}. %
These methods iteratively generate the design population in a stochastic manner by joining or modifying existing designs, solving for their individual controllers, and then evaluating their performance.
\citet{hejna2021task} employ an evolutionary method to optimize a task-agnostic information-theoretic objective that ranks a population of morphologically diverse agents by their ability to reach different locations in the environment using random actions.
However, ES is prone to local minima and is sample-inefficient, in part due to the need to maintain separate control policies for each candidate design in the population. Our approach improves upon the sample-efficiency of these methods by sharing experience across designs through a single control policy.
Alternatively, several approaches improve sample efficiency by leveraging additional knowledge or assumptions about the system dynamics, typically in the context of optimizing the continuous parameters of a fixed morphology.
These approaches include trajectory optimization~\citep{spielberg17, bravo2020one}, linear approximations to the dynamics~\citep{ha17}, and leveraging differentiable simulators~\citep{xu2021end}.

Similar to our approach, several reinforcement learning-based strategies to co-optimization exist for both continuous~\cite{ha_reinforcement_2019, luck_data-efficient_2019, chen_hardware_2020} and discrete design spaces~\citep{spielberg2019learning, pathak2019learning, yuan2021transform2act}, including the work presented in Chapter~\ref{sec:design_and_control}~\cite{schaff_jointly_2019} and Chapter~\ref{sec:soft_design}~\cite{schaff2022soft}.
\citet{luck_data-efficient_2019} use a soft actor-critic algorithm and use a design-conditioned Q-function to evaluate designs. 
\citet{chen_hardware_2020} model the design space as a differentiable computational graph, which allows them to use standard gradient-based methods.  
In the context of discrete design spaces, \citet{spielberg2019learning} propose an autoencoder-based method that jointly optimizes the placement and control of a large number deformable voxels for soft-body locomotion tasks.
\citet{yuan2021transform2act} represent the design generation process as part of the environment and train a dual-purpose design generation and control policy.
\citet{pathak2019learning} propose a modular strategy whereby primitive agents self-assemble in a dynamic fashion 
to form a collective agent that is controlled using a graph neural network-based policy.
In contrast to our approach, their method produces simple assemblies with a small number of components.

\paragraph{Universal control policies:} Traditional population-based co-optimization strategies are difficult to scale due to the need to maintain separate controllers for each design within the population as it changes over time.
Recent approaches adopt a single design-aware control policy to improve optimization efficiency. Our approach from Chapter~\ref{sec:design_and_control} proposes such an approach for optimization within fixed morphology by leveraging a design-conditioned policy capable of generalizing to new designs.
This improves the efficiency of co-optimization by avoiding the need to solve for the optimal controller for each candidate design, but is only able to control designs within a fixed morphology.
More sophisticated controllers leverage graph neural networks (GNN)~\citep{scarselli2008graph} structured according to the robot morphology~\citep{wang2018nervenet, pathak2019learning, wang2019neural, Huang2020, yuan2021transform2act}.
Recently, shared controllers based on self-attention mechanisms~\citep{Kurin2021, gupta2022metamorph} have outperformed GNNs due to the ability to propagate information across nodes more effectively than the message-passing schemes of GNNs.
Motivated by the empirical success of self-attention over GNNs, we leverage transformers~\citep{vaswani2017attention} to model our controller.
\section{Co-optimization of Design and Control}

In this section, we reintroduce the problem of jointly optimizing the physical design and control of a robot in the context of a specified task, and our algorithm for solving this problem.
While these descriptions are largely the same from Chapter~\ref{sec:design_and_control}, there are some minor differences.

\subsection{Problem Definition}

We formulate the problem of co-optimizing design and control as a set of related reinforcement learning problems.
Given a set of robot designs $\Omega$ and a task definition, we define a \textit{design-specific} Markov decision process (MDP) for each $\omega \in \Omega$: $\mM_\omega = \textrm{MDP}(\mS_\omega,
\mA_\omega, \mP_\omega, \mR_\omega)$, where $\mS_\omega$ is the state space, $\mA_\omega$ is the action space, $\mP_\omega: \mS_\omega \times \mA_\omega \times \mS_\omega \rightarrow [0,1]$ is the transition dynamics, and $\mR_\omega: \mS_\omega \times \mA_\omega \times \mS_\omega \rightarrow \mathbb{R}$ is the reward function.
When optimizing for a single task objective, it is natural that these MPDs share some common structure, e.g., the state space only changes based on the proprioceptive information available to each design, the action spaces differ only in the number of controllable degrees of freedom, and each reward $\mR_\omega$ encodes the same task objective.

Let $\pi^*_\omega: \mS_\omega \times \mA_\omega \rightarrow [0, 1]$ be the optimal policy for MDP $\mM_\omega$ and $\mathbb{E}_{\pi^*_\omega}\left[\sum_t \gamma^t r_t\right]$ be its expected return. The goal of co-optimization is to find the optimal design-controller pair $(\omega^*$,~$\pi^*_{\omega^*})$ such that:
\begin{equation}
    \label{eqn:nlimb_objective}
    \omega^*, \pi^*_{\omega^*}  = \underset{\omega, \pi^*_\omega}{\text{arg max}} \; \mathbb{E}_{\pi^*_\omega}\left[\sum_t \gamma^t r_t \right].
\end{equation}

\subsection{Co-optimization via Universal and Transferable Policies}
\begin{figure}[t]
    \begin{algorithm}[H] \label{alg:nlimb_main}
        \caption{Joint Optimization of Design and Control} \label{nlimb_algorithm}
        \begin{algorithmic}[1]
            \STATE Initialize $\pi_\theta(a \vert s,\omega)$, $p_\phi(\omega)$, $T=0$, $T_0$
            \WHILE{$T < \textrm{BUDGET}$}
                \STATE Sample designs $\omega_1, \omega_2, \hdots, \omega_n \sim p_\phi$ \label{alg:nlimb_sample}
                \STATE Control $\omega_1, \omega_2, \hdots, \omega_n$ with $\pi_\theta$ for $t$ timesteps \label{alg:nlimb_control}
                \STATE Update $\theta$ using \textrm{PPO} with collected trajectories \label{alg:nlimb_ppo}
                \STATE Set timestep $T = T + nt$
                \IF {$T > T_0$}
                    \STATE Compute average episode returns $R_{\omega_1}, R_{\omega_2}, \hdots, R_{\omega_n}$
                    \STATE Update $\phi$ using $\nabla_\phi \approx \sum_{i=0}^n \nabla \text{log }p_\phi(\omega_i) R_{\omega_i}$
                \ENDIF
            \ENDWHILE
        \end{algorithmic}
    \end{algorithm}
\end{figure}

Based on Equation \ref{eqn:nlimb_objective}, the search over the design space $\Omega$ requires access to the optimal controller $\pi^*_\omega$ for each design.
Obtaining these controllers can be challenging, as it is often computationally intractable to train independent controllers for each design.
However, we can leverage the shared structure between each design to efficiently train a \textit{design-conditioned} controller: $\pi: \mS \times \mA \times \Omega \rightarrow [0, 1]$.
Ideally, this controller would serve as a proxy for optimal controllers as well as generalize zero-shot to unseen designs in the design space. Crucially, this controller must be able to interface with each state space $\mS_\omega$ and action space $\mA_\omega$, for all $\omega \in \Omega$.
Implementing such a controller has been studied in the context of co-optimization~\citep{schaff_jointly_2019, yuan2021transform2act, chen_hardware_2020} as well as the transfer of control policies across morphologies~\citep{Huang2020, Kurin2021, gupta2022metamorph}.

With such a control policy, the co-optimization procedure reduces to search over the design space $\Omega$.
In this work, we use a zero-order search algorithm based on policy gradients that optimizes a distribution over the design space towards higher performing designs.
Let $p_\phi$ be a distribution over the design $\omega \in \Omega$ parameterized by $\phi$ and $\pi_\theta$ be a design-conditioned policy with parameters $\theta$.
Our training objective is then to optimize $\pi_\theta$ and $p_\phi$ to maximize the expected return:
\begin{equation}
    \label{eqn:nlimb_train}
    \phi^*, \theta^*  = \underset{\phi, \theta}{\text{arg max}} \; \mathbb{E}_{\omega \sim p_\phi}\mathbb{E}_{\pi_\theta}\left[\sum_t \gamma^t r_t \right].
\end{equation}

The controller can then be trained using any standard RL algorithm and the design distribution can be updated with any zero-order method. Similar to past work (Chapter~\ref{sec:design_and_control}, \citet{schaff_jointly_2019}), we use policy gradients to update both the policy and the design distribution.
This leads to the following update equations:
\begin{equation}
    \label{eqn:nlimb_update}
    \nabla_\phi = \mathbb{E}_{\omega \sim p_\phi}\left[\nabla_\phi\text{log }p_\phi(\omega)\mathbb{E}_{\pi_\theta}\left[\mR_t \right] \right] \hspace{2em}
    \nabla_\theta = \mathbb{E}_{\omega \sim p_\phi}\left[\mathbb{E}_{\pi_\theta}\left[\nabla_\theta \text{log }\pi_\theta(s_t, a_t)\mR_t\right] \right],
\end{equation}
where $\mR_t$ is the expected return at time $t$.
This algorithm trains a controller to maximize performance in expectation over the design distribution $p_\phi$.
When this controller is sufficiently trained, the algorithm uses it as a proxy for the optimal controllers of each design, and subsequently updates the design distribution $p_\phi$ to favor higher performing designs. Training continues until the design distribution $p_\phi$ converges to a single design and the policy $\pi_\theta$ is optimized for that design.
See Algorithm~\ref{alg:nlimb_main} for details.
\section{\nlimb: Optimizing Design and Control over Morphologies}
\label{sec:nlimb}

While our approach described in Chapter~\ref{sec:design_and_control} successfully employ the above procedure to find optimal design-control pairs, the approach is limited to simple, fixed (i.e., pre-defined) kinematic structures and is not able to optimize across morphologies. We propose an algorithm that builds on this approach to perform morphological optimization. We describe a general approach to defining combinatorial spaces of robot designs, a policy architecture that is able to control and generalize across that space, and a way to parameterize and optimize complex distributions over the design space.

\subsection{Graph Grammars for Defining Design Spaces across Morphologies}

In this work, we explore the use of context-free graph grammars to define the space of valid morphologies.
With the end goal being to build and deploy optimized design-control pairs, fabrication constraints must be taken into account, including what parts can be used and in which ways can they be feasibly combined.
Context-free grammars can naturally capture these constraints, with terminal symbols that correspond to available parts and expansion rules that represent limits on how those parts can be combined. In this way, the grammar focuses the optimization only on feasible designs.

We define a context-free graph grammar as the tuple $(N, T, R, S)$, where $N$ denotes the set of non-terminal symbols, $T$ denotes the set of terminal symbols, $R = \{r_s^i = (s, G_{s}^i)\}$ is the set of expansion rules that map a non-terminal symbol $s \in N$ to a graph $G_s^i$, and $S$ is the starting graph. The application of a expansion rule replaces a node with non-terminal symbol $s$ with the corresponding graph $G_s^i$. Nodes within these graphs express limb information (e.g., geometry, pose, mass, inertia, etc.), and edges contain joint information (e.g., type, the kinematic transformation between limbs, max torque, gear ratio, etc.).

\begin{figure}[!t]
    \centering
    \includegraphics[width=\textwidth]{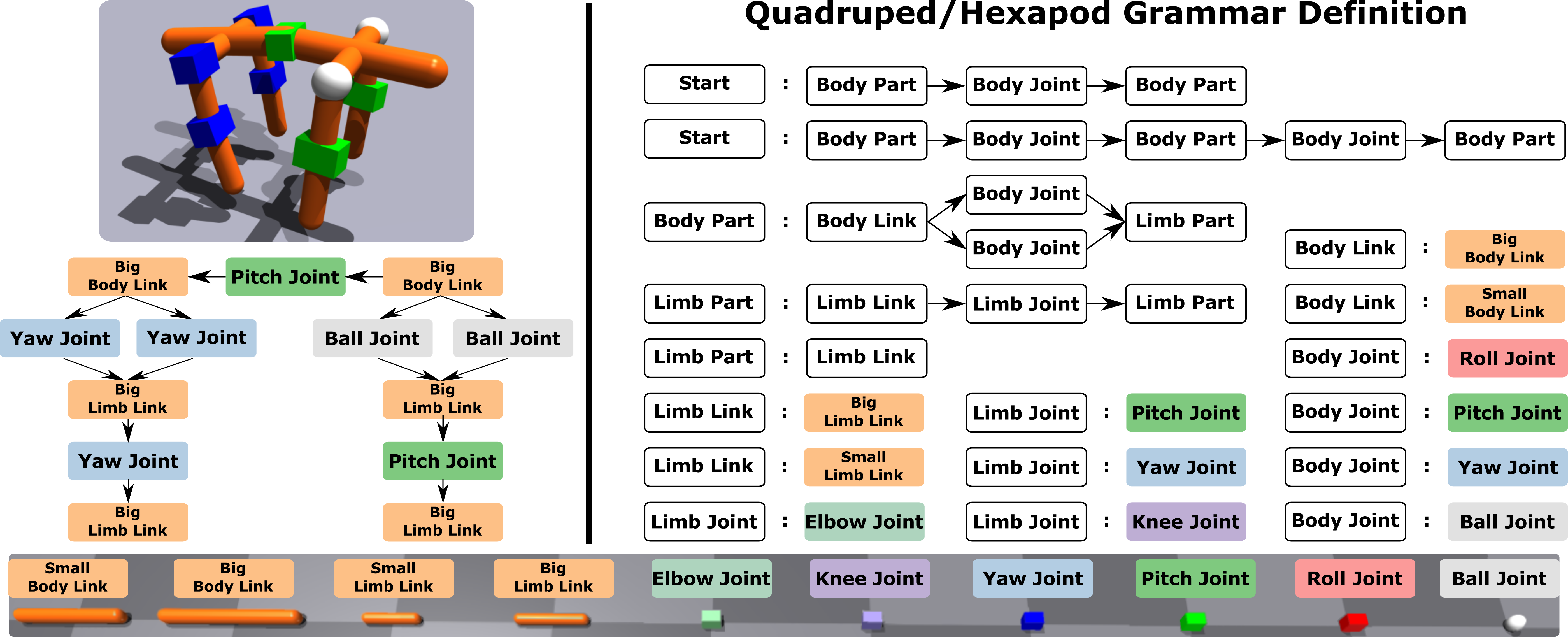}
    \caption{Our approach defines the space of valid designs using a context-free graph grammar in which terminal symbols denote robot parts and expansion rules describe how those parts can be combined. \textbf{Left}: An example design and its corresponding graph structure. \textbf{Right}: The definition of a bio-inspired grammar that produces quadruped and hexapod designs. Nonterminal symbols are denoted with white boxes, terminal symbols with colored boxes, and expansion rules are denoted with a colon. \textbf{Bottom}: Terminal symbols and their corresponding parts, where joints are colored according to their degrees-of-freedom.}\label{fig:nlimb_grammar}
\end{figure}
In our experiments, we construct a grammar (Figure~\ref{fig:nlimb_grammar}) to define a space of quadruped and hexapod robots. This grammar generates robots as a composition of simple shapes, maintains left-right symmetry, and allows for a variety of joint types and limbs with varying numbers of parts and sizes.

\subsection{Universal Controllers via Morphologically-aware Transformers}

\begin{figure}[!t]
    \centering
    \includegraphics[width=\textwidth]{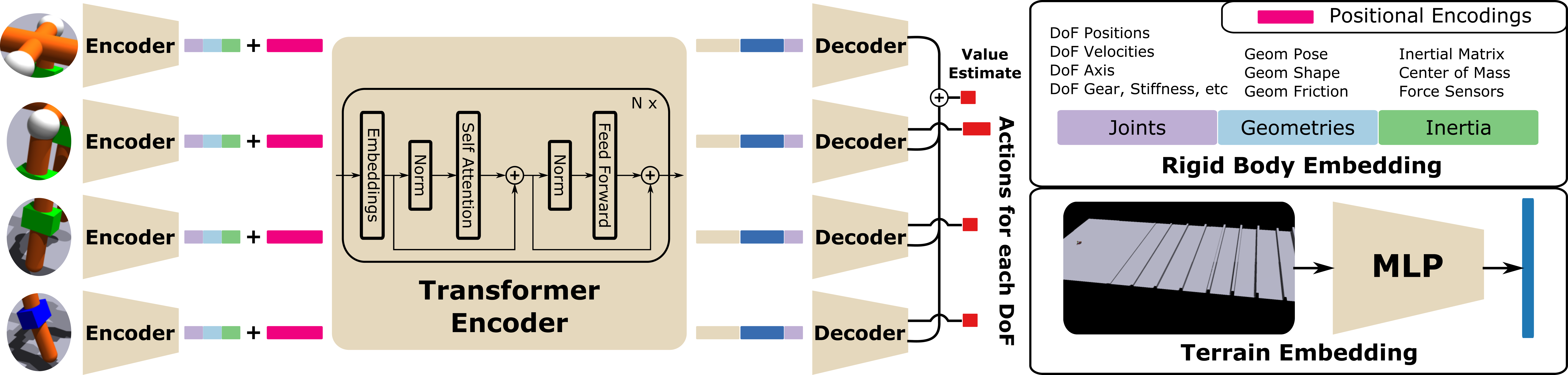}
    \caption{A rendering of our self-attention based actor-critic architecture. Rigid body pose and attributes are encoded and processed via a transformer encoder to produce actions for each degree-of-freedom and a value estimate.}\label{fig:nlimb_policy}
\end{figure}
Our co-optimization algorithm requires a universal controller that can act as a proxy for the optimal controller across the design space $\Omega$. Control policies that employ graph neural networks~\citep{wang2018nervenet,Huang2020} as well as those that use self-attention~\cite{Kurin2021, gupta2022metamorph} have proven effective at handling the varying number of rigid-bodies and controllable degrees-of-freedom across morphologies.
In this work, we use the transformer architecture~\cite{vaswani2017attention} to parameterize an actor-critic network as it was shown to outperform graph neural networks~\cite{Kurin2021}.

Figure~\ref{fig:nlimb_policy} provides a visualization of our actor-critic architecture. At each point in time, the agent receives various types of state information: local information about each rigid body and joint (e.g., pose, velocity, etc.), morphological information (e.g., shape, inertial, joint axes, etc.), and global sensory information (e.g., cameras, terrain observations, task-relevant observations, etc.). Our architecture consists of three core components: the first module encodes local and morphological information; the second processes that information through self-attention and then appends global information; and the third module decodes actions and a value estimate.

\paragraph{Rigid-body embeddings:}
We encode each robot by flattening its graph structure into a sequence of rigid-body embeddings using a depth-first traversal. Each of these embeddings contains local pose and morphological information about the rigid body, as well as each controllable degree-of-freedom that connects it to its parent. 
Information about geometries, degrees-of-freedom, and inertia are encoded separately using two-layer MLPs and concatenated to obtain the rigid-body embedding.

\paragraph{Process:}
The rigid-body embeddings are combined with positional encodings and processed with a transformer encoder.
Similar to \citet{gupta2022metamorph}, we postpone the fusion of global information, such as observations about the surrounding terrain, until after the self-attention layers to both reduce the number of transformer parameters as well as avoid diluting the low-dimensional pose and morphological information. 
Terrain observations are comprised of sparsely sampled height-field information that is encoded into a flat embedding vector using a three-layer MLP and concatenated with each processed rigid-body embedding. 

\paragraph{Decoding actions and value estimates:}
After the addition of global sensory information, we produce a value estimate by averaging the outputs of a value-decoder MLP across each rigid-body.
For each degree-of-freedom associated with a rigid body, we concatenate an embedding of that DoF before applying an action-decoder MLP that produces the mean and state-independent standard deviation of a Gaussian distribution.

\subsection{Autoregressive Models for Design Generation}

\begin{figure}[!t]
    \centering
    \includegraphics[width=\textwidth]{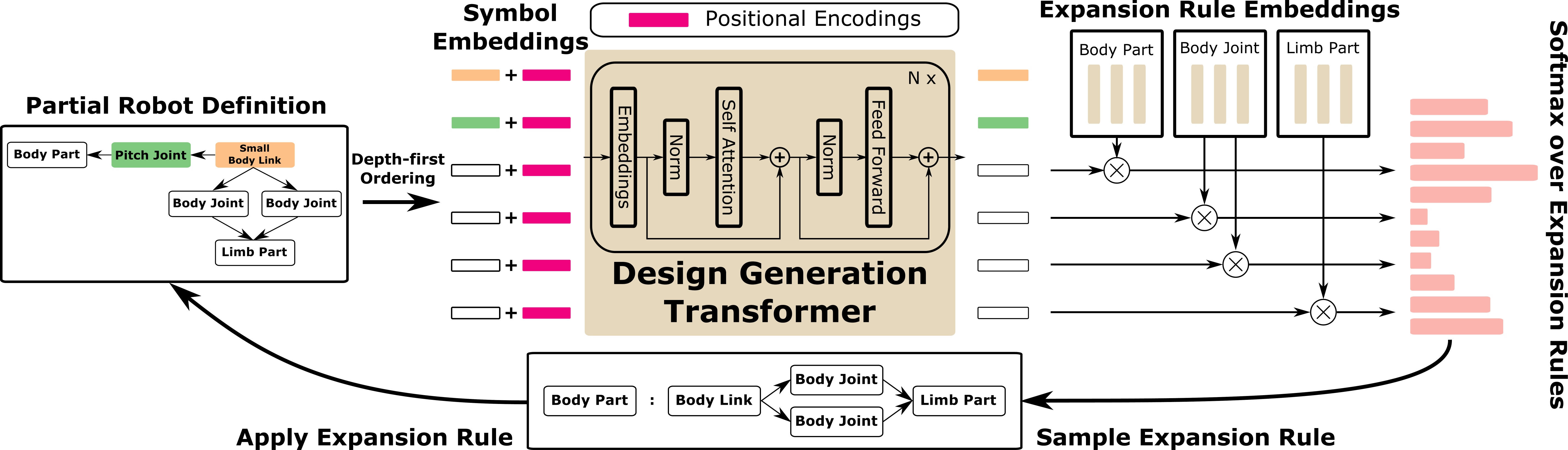}
    \caption{The design distribution is modeled via an autoregressive transformer architecture. Partial graphs are processed by the model to produce a distribution over expansion rules of the grammar. Rules are sampled and applied until no non-terminal symbols remain within the graph.}\label{fig:nlimb_design-generation}
\end{figure}
Figure~\ref{fig:nlimb_design-generation} provides an overview of our design generation model.
Given the definition of a graph grammar, designs can be generated by autoregressively sampling expansion rules until the resulting graph contains only terminal symbols.
To build such a model, we again use the transformer architecture~\cite{vaswani2017attention}.
Sampling designs in this way allows, in principle, for arbitrary distributions over the design space that can capture multi-modal behavior and complex dependencies between symbols.

At each stage of generation, the model receives as input a partial graph $G$.
This graph is then flattened to a sequence of symbols $s_1$, \dots, $s_n$ using a depth-first traversal. Those symbols are then embedded using an embedding table, combined with a positional encoding, and processed with a transformer encoder to produce representations $h_{s_1}, \ldots, h_{s_n}$.
We decode logits for each expansion rule by computing a dot-product between each representation $h_{s_i}$ and learnable embeddings corresponding to each expansion rule $r_{s_i}^j$ associated with symbol $s_i$.
These logits are combined across all symbols in the graph $G$ using a softmax operation.
An expansion rule is then sampled from this distribution and used to update the graph $G$, after which the process is repeated until $G$ contains only terminal symbols. %

\section{Experiments} \label{sec:nlimb_experiments}

\begin{figure}[t!]
    \centering
    \includegraphics[width=\linewidth]{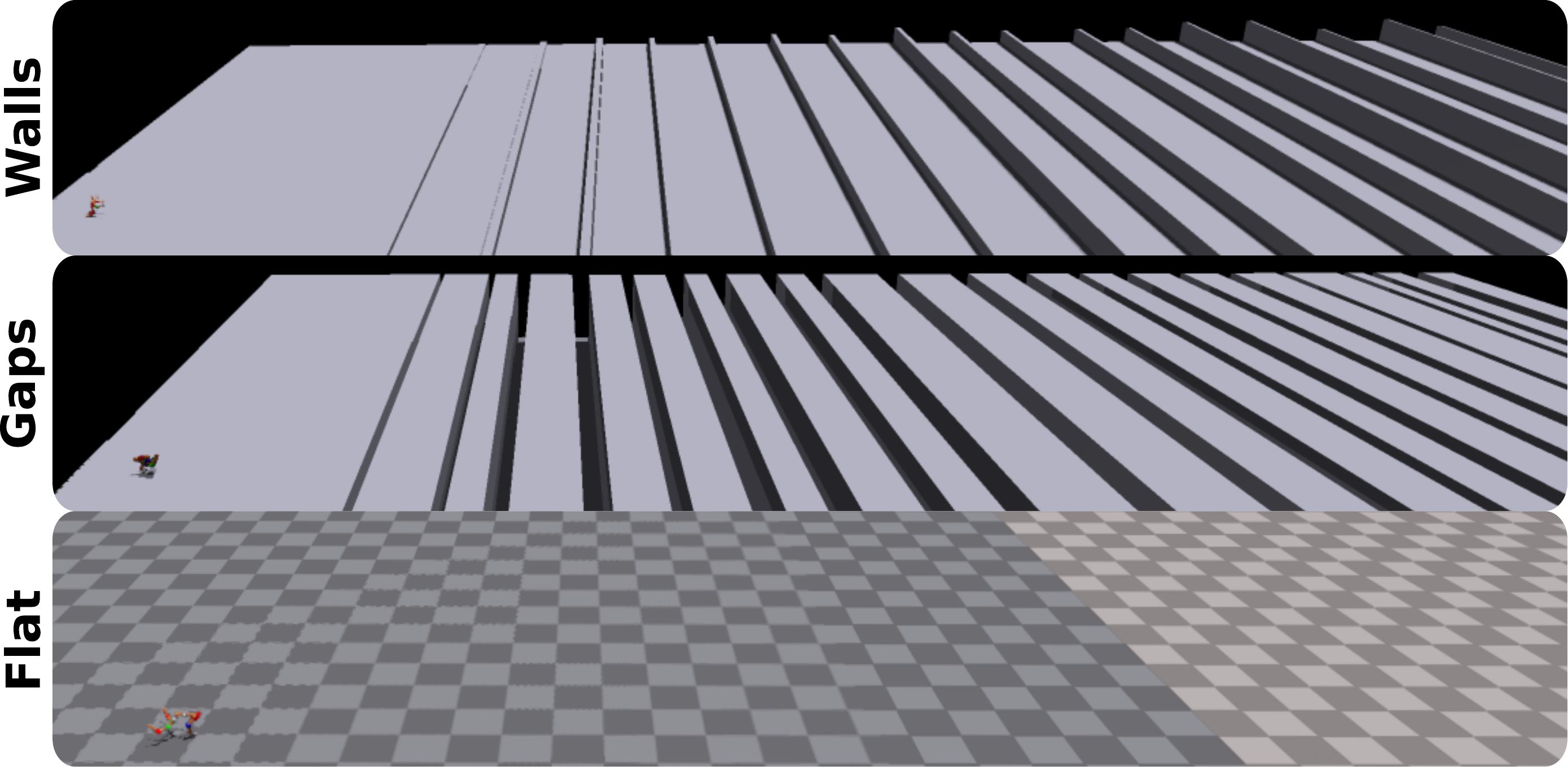}
    \caption{Images depicting the three terrains used in our experiments. In the ``Walls'' and ``Gaps'' environments, the walls (gaps) are randomly spaced with heights (widths) that increase farther from the agent's initial pose. New obstacles are resampled regularly throughout training.}
    \label{fig:nlimb_terrains}
\end{figure}

We evaluate our \nlimb framework by performing co-optimization of design and control for locomotion tasks involving a variety of terrains.
In all cases, \nlimb efficiently finds high-performing design-control pairs that outperform baselines.

\begin{figure}[t!]
    \centering
    \includegraphics[width=\linewidth]{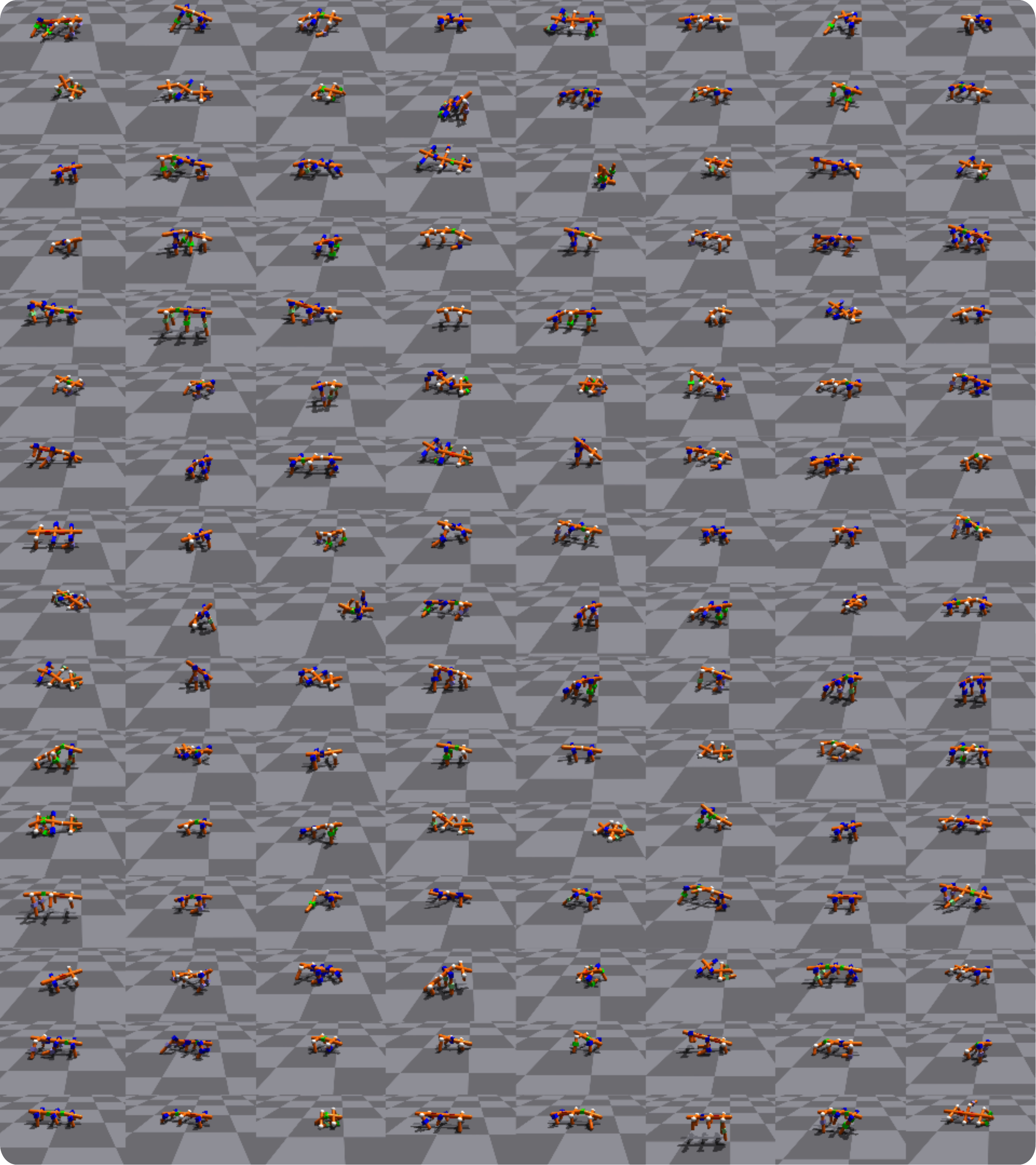}
    \caption{A visualization of randomly sampled designs from the hexapod grammar. The grammar contains millions of combinations of limb and joint parameters.}
    \label{fig:nlimb_design_space_viz}
\end{figure}

\subsection{Experiment Details}
We optimize and evaluate design-control pairs using the IsaacGym GPU-based physics simulator~\cite{makoviychuk2021isaac}, allowing us to train with thousands of parallel environments on a single GPU.
The locomotion task rewards each robot for making as much forward progress as possible within a fixed amount of time, with penalties for energy consumption, large actuation torques, and reaching joint limits.
Episodes are terminated early if the robot has fallen over.
We consider three different terrain types (Figure~\ref{fig:nlimb_terrains}): (i) flat terrain (``Flat''); (ii) a terrain with randomly placed gaps of random widths that the robot must cross (``Gaps''); and (iii) a terrain with randomly placed walls of different heights that the robot must climb over (``Walls''). Obstacles are regularly resampled throughout training.
For these experiments, we use the quadruped and hexapod grammar defined in Section~\ref{sec:nlimb}. Figure~\ref{fig:nlimb_design_space_viz} provides a visualization of randomly sampled designs from this grammar. 
We run the \nlimb algorithm on a single NVIDIA A4000 GPU for $1$B timesteps (roughly two days).
We use Proximal Policy Optimization~\cite{schulman17} to train both the controller and design distribution. For more details and hyperparameters, see Section~\ref{app:hyps}.

\subsection{Baselines}
Similar to previous work in co-optimization including our own~\citep{schaff_jointly_2019,zhao2020robogrammar,hejna2021task}, we compare our framework against a decoupled approach that randomly samples designs from the hexapod grammar and trains controllers for each design individually.
To ensure a fair comparison with our approach, we provide the baseline with the same computational budget as \nlimb. 

\begin{figure}[!t]
    \centering
    \includegraphics[width=\linewidth]{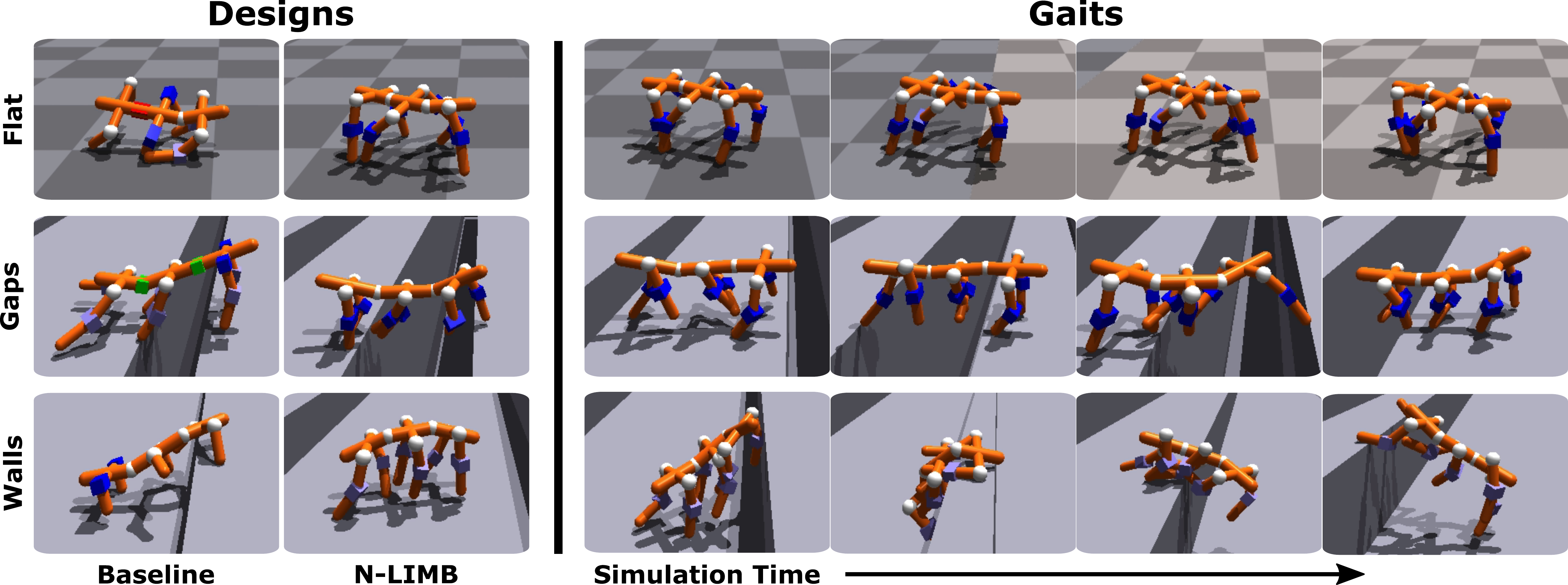}
    \caption{\textbf{Left:} A comparison between the designs learned by \nlimb and those of the random search baseline for the three terrain types. \textbf{Right:} A visualization of the learned gaits. The colored markers denote different types of degrees of freedom between rigid bodies, with white spheres denoting spherical joints and colored boxes denoting hinge joints with varying axes.}
    \label{fig:nlimb_designs-gaits}
\end{figure}

\subsection{Results}
Across all three terrains, we find that \nlimb is able to outperform the random search baseline by a large margin.
Figure~\ref{fig:nlimb_designs-gaits} shows the best designs found by \nlimb and the baseline, as well as the learned gaits while traversing difficult sections of each of the three terrain types. While \nlimb initially favors quadruped designs in some cases (Fig.~\ref{fig:nlimb_motivation}), it converges to hexapod designs for all three terrain types. \nlimb chooses to connect the body links with ball joints, suggesting that additional degrees-of-freedom outweighs a reduction in available torque and narrower joint limits.
The choice of a hexapod design may be a result of the fact that achieving longer bodies with our grammar requires a design with six legs.
Indeed, on the Gaps terrain (Fig.~\ref{fig:nlimb_designs-gaits}, middle row), \nlimb selects a design that has the largest body available within the grammar through the use of larger body links, which is necessary to cross wider gaps.
The baseline similarly identifies a long-bodied hexapod, but it connects the body links with fewer degrees-of-freedom, reducing its flexibility.
On the Walls terrain, however, the optimal design (Fig.~\ref{fig:nlimb_designs-gaits}, bottom row) has a smaller body and connects limb links with ``knee'' joints that can bend back $180^\circ$, allowing the robot to fold its limbs while cresting or pushing off the walls. In contrast, the optimal baseline design has legs with a single small link, which limits the height of the walls that the baseline robot can climb over. For locomotion on the Flat terrain, \nlimb converges to a design identical to that of the Gaps terrain, except that the torso is comprised of three small (vs.\ large) body links, trading off the need to span wide gaps for the availability of lower-torque bounding gates.
The baseline identifies a hexapod with single-link legs and replaces some of the ball joints with pitch and roll joints.

\begin{figure}[!t]
    \centering
    \includegraphics[width=\linewidth]{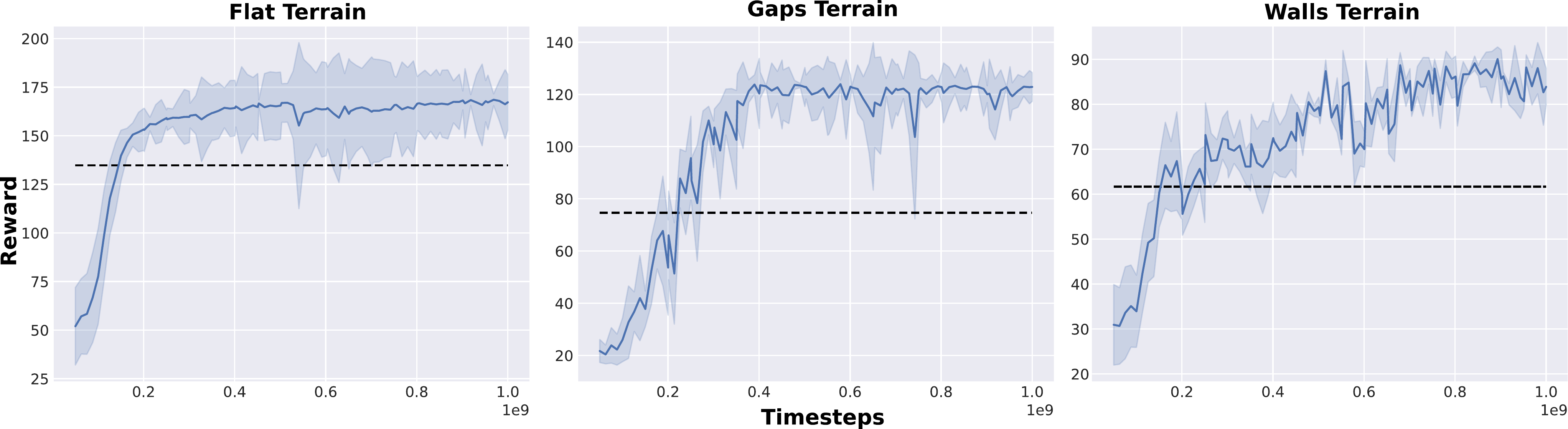}
    \caption{Plots that show the reward (mean and standard deviation) of designs sampled from the design distribution over the course of training for the three terrain types. Dashed horizontal lines denote the rewards of the highest performing design-control pairs for the baseline algorithm.} \label{fig:nlimb_reward-evolution}
\end{figure}
Figure~\ref{fig:nlimb_reward-evolution} visualizes the quantitative performance of \nlimb compared to the baseline. 
It displays the mean and standard deviation of returns obtained by designs sampled from the design distribution throughout training for each of the three terrain types. Early on, the ability for the policy to control the set of morphologies  drawn from the design distribution is limited, resulting in poor performance for most designs. However, as the controller improves and \nlimb updates the design distribution accordingly, we see that the algorithm  yields designs paired with the universal control policy that quickly outperform the baseline on all three terrains.

\subsection{Generalization of the Universal Controller}
\begin{figure}[t]
    \centering
    \includegraphics[width=0.7\linewidth]{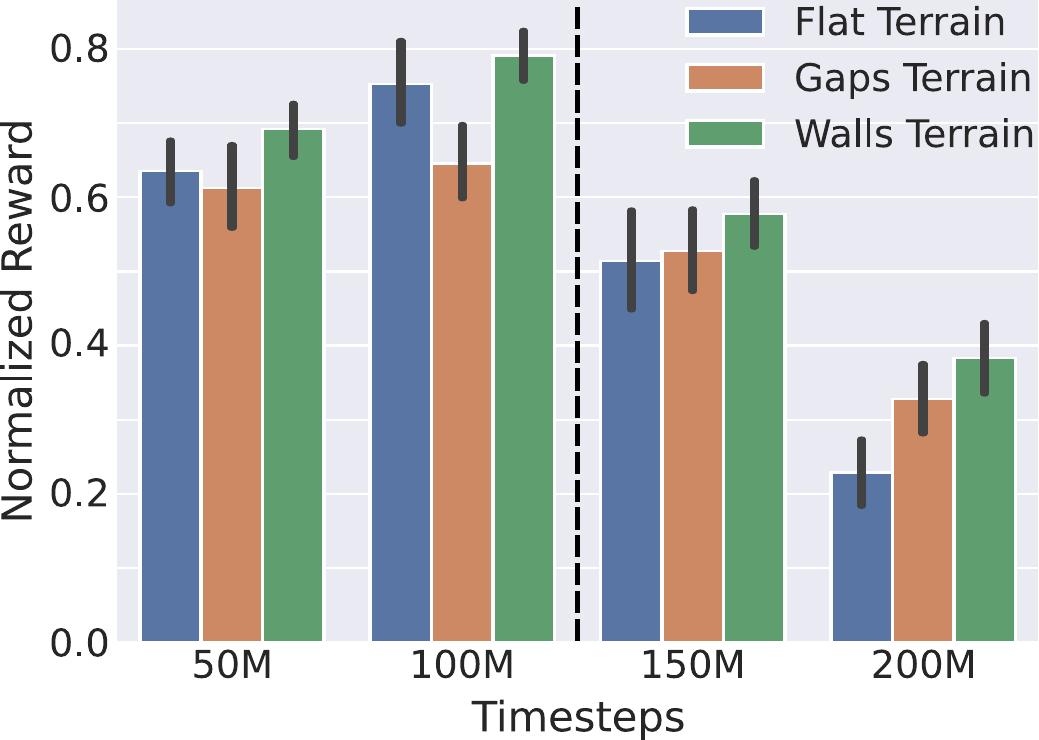}
    \caption{A visualization of the generalizability of the universal \nlimb controller as a fraction of the reward achieved using controllers trained separately on designs sampled from the initial design distribution. In all three domains, the relative reward improves until, after $100$M timesteps (dashed line), we begin to update the design distribution, upon which the \nlimb policy learns to specialize to higher-performing designs.}\label{fig:nlimb_generalization}
\end{figure}
Crucial to our approach is a universal controller that provides a proxy for the optimal controller when evaluating a design.
If this controller is unable to provide a reasonable estimate of a design's optimal performance, the design distribution may incorrectly focus the search around suboptimal designs.
To verify that our controller provides a reasonable proxy, we compare it against controllers trained individually on each design sampled from the initial design distribution.
Figure~\ref{fig:nlimb_generalization} shows the performance of the universal controller as a fraction of the reward of the individually trained controllers at four checkpoints during training.
We keep the design distribution fixed for the first $100$M steps of training, during which the universal controller impressively obtains up to $80$\% of the performance of individually trained controllers.
Once we begin updating the design distribution (after $100$M timesteps), we see that performance expectedly declines as the controller specializes to the shifting design distribution.
\section{Conclusion} \label{sec:nlimb_conclusion}

We presented \nlimb, an efficient and effective approach to co-optimizing robots and their controllers across large, combinatorial sets of morphologies.
\nlimb formulates the set of valid morphologies as a context-free graph grammar, allowing users to easily incorporate fabrication constraints and inductive biases, thereby focusing search on realizable robots.
Given such a grammar, we parameterize a distribution over the design space using a novel autoregressive model that recursively samples the expansion rules of the grammar until a completed robot is formed.
The optimization is carried out by training a universal controller in expectation over the design distribution, while simultaneously shifting that distribution towards higher-performing designs.
In this way, the optimization process converges to a design-control pair that is jointly optimal for the given task.
We demonstrate the potential of our approach by learning novel, high-performing design-control pairs on a variety of locomotion tasks and terrains.

Among the limitations of this work, we only show results in simulated settings. While we designed our grammar to preclude difficult-to-fabricate designs, we do not demonstrate real-world transfer of the learned design-controller pairs. Inspired by the success of sim-to-real transfer for control~\citep{tan18, zhao2020sim} and co-optimization~\cite{schaff2022soft}, an interesting dircetion for future work is to investigate the real-world performance of the resulting designs. Further, we place bio-inspired constraints on our grammar to avoid unreasonable designs and, in turn, to improve the efficiency of optimization. These rules may be overly conservative, preventing our method from learning designs that while being atypical, are well-suited to the task.

\clearpage %
\section{Additional Details}

This section covers additional visualizations and implementation details relevant to reproduce the experiments in this chapter.

\subsection{Network Details}
Our actor-critic network consists of a transformer encoder with one layer of self-attention, a model dimension of $256$, and $4$ attention heads. The input to the encoder is a sequence of rigid-body embeddings created from a concatenation of geometric, inertial, and joint embeddings.
We use a two-layer MLP to encode information about each geometry and subsequently add these geometric encodings together. 
Similarly, we use a two-layer MLP to encode information about each degree-of-freedom and then add them together. 
The network also uses a two-layer MLP to encode inertial information.
These three embeddings are then concatenated together.
We order the rigid-body embeddings by flattening the robot's kinematic tree with a depth-first traversal, added with a learned positional encoding, and apply dropout with $p=0.1$.
We separately encode terrain information with a three-layer MLP.
After the transformer encoder, terrain information is concatenated with each token. We apply a three-layer value-decoder MLP to each token and then average the output to produce a value estimate.
To produce actions, we concatenate degree-of-freedom information with each token and then pass the result to a three-layer action-decoder network that produces the mean and (state-independent) standard deviation of a Gaussian distribution.

Our autoregressive design distribution also consists of a transformer encoder with one layer of self-attention, a model dimension of $256$, and $4$ attention heads.
We order partial graphs using a depth-first traversal, and embed symbols for each node with an embedding table, that we then add with a positional encoding and feed to the transformer encoder.
For non-terminal symbols, we decode the resulting representations with a two-layer MLP and compute the dot product with embeddings of the grammar's expansion rules to produce logits. We combine these logits across nodes, and create a categorical distribution via softmax. 
Robot generation proceeds by sampling expansion rules and updating the graph until only terminal symbols remain.

\clearpage
\subsection{Table of Hyperparameters}
\label{app:hyps}

\begin{table}[!h]
    \centering
    \begin{tabularx}{\linewidth}{Xr}
        \toprule
        Environment Hyperparameter  &  Value \\
        \midrule
        Number of environments & $2048$ \\
        Maximum episode length & $1000$ \\
        dt & $1/60$\,sec \\
        Termination height & $0.08$ \\
        Reward: forward progress weight & $3.8$ \\
        Reward: alive bonus & $0.5$ \\
        Reward: termination cost & $2.0$ \\
        Reward: energy cost scale & $0.01$ \\
        Reward: squared action cost scale & $0.035$ \\
        Reward: joints at limit cost scale & $0.2$ \\
        \midrule
        PPO Hyperparameter  &  Value \\
        \midrule
        Discount factor ($\gamma$) & $0.99$\\
        GAE $\lambda$ & $0.95$\\
        Rollout length & $128$\\
        Batch size & $16384$\\
        Epochs per rollout & $4$\\
        Clip param & $0.2$\\
        Policy loss coefficient & $1$\\
        Value loss coefficient & $3.7$\\
        Entropy bonus coefficient & $0.0$\\
        Action bounds loss coefficient & $100$\\
        Max gradient norm & $1.0$\\
        KL target & $0.04$\\
        Initial learning rate & $0.0003$ \\
        Optimizer & $\textrm{ADAMW}(\beta_1 = 0.9, \beta_2 = 0.999, \text{weigth decay} = 0.018)$ \\
        \midrule
        \textsc{N-Limb} Hyperparameter  &  Value \\
        \midrule
        Maximum timesteps & $1$B \\
        Policy warm-up period & $60$M \\
        Timesteps per design & $3000$ \\
        Update period & $12$M \\
        Batch size & $2048$\\
        Epochs per update & $2$\\
        Clip param & $0.2$\\
        Policy loss coefficient & $1$\\
        Entropy bonus coefficient & $0.02$\\
        Max gradient norm & $0.1$\\
        KL target & $0.0126$\\
        Initial learning rate & $0.001$ \\
        Optimizer & $\textrm{ADAMW}(\beta_1 = 0.9, \beta_2 = 0.999, \text{weight decay} = 0.018)$\\
        \bottomrule%
    \end{tabularx}%
    \caption{Environment, PPO, and \textsc{N-Limb} Hyperparameters used in the experiments of this chapter.}\label{tab:nlimb_sac_param}%
\end{table}%

\chapter{Optimizing Design and Control for Soft Robotics} \label{sec:soft_design}

The deformable nature soft robots enables designs that respond to contact or control inputs in sophisticated ways, with behaviors that have proven effective across a variety of domains.
The design of such systems is tightly coupled with the policy that controls their motion, giving rise to a form of ``mechanical intelligence''~\cite{rus2015design} in which materials and mechanisms respond to their environment in useful ways that augment functionality, e.g., conforming to an object to create a better grasp or storing elastic energy to improve the efficiency and power of a walking gait. Therefore, methods that jointly optimize both the robot's physical design and its control policy provide a promising approach to discovering mechanically intelligent soft robots.
However, while researchers have extensively explored the problem of joint design-control optimization in the context of rigid robots~\cite{sims1994, park94, paul01, paul06, spielberg17, seo2019modular, digumarti_concurrent_2014, ha17, zhao2020robogrammar, schaff_jointly_2019, ha_reinforcement_2019, chen_hardware_2020, pathak2019learning}, relatively little work exists for soft robotics.

This work provides a complete framework for the simulation, co-optimization, and sim-to-real transfer of the design and control of soft robots. Integral to this framework, we propose a co-optimization algorithm
that utilizes multi-task deep reinforcement learning to generate a design-aware policy capable of generalizing across the space of designs. The algorithm exploits this policy to quickly focus its search on high-performing designs. 

An important prerequisite for co-optimization is a simulator that is both fast enough to explore a large set of designs and control strategies and accurate enough to ensure that the learned robots are physically realizable and capable of sim-to-real transfer.
However, modelling soft bodies is both challenging and computationally intensive.
The best way to simulate soft bodies for robotics is an open question, and the few co-optimization approaches for soft robotics suggest different simulation strategies~\cite{Hu19,spielberg_co-learning_2021,Hiller2014}.
However, these simulators have varying degrees of realism and their ability to produce soft robots that cross the reality gap is unclear.
Therefore, in addition to designing a co-optimization algorithm, new simulation approaches are also required to improve computational efficiency and transferability to the real world.

We employ finite element analysis (FEA), which is the defacto standard for simulating deformable materials with a high degree of accuracy. In order to improve the computational complexity of FEA-based simualation while preserving its accuracy, we extend the recent work of \citet{Goury18} that proposes a model order reduction technique for soft robotics in the open-source FEA simulation framework SOFA~\cite{Faure2012, Coevoet17}.
Their reduction technique has a large initial computational cost, but then allows for simulating a fixed robot with a computational efficiently sufficient for learning-based methods, while maintaining physical realism.
However, co-optimization requires a search over many unique designs and reducing each one is computationally infeasible.
To overcome this, we propose a reconfigurable reduction framework that reduces a set of composable parts that can then be combined to create reduced order models of soft robots with varying morphologies.

\begin{figure}[!t]
    \centering
    \includegraphics[width=1.0       \textwidth]{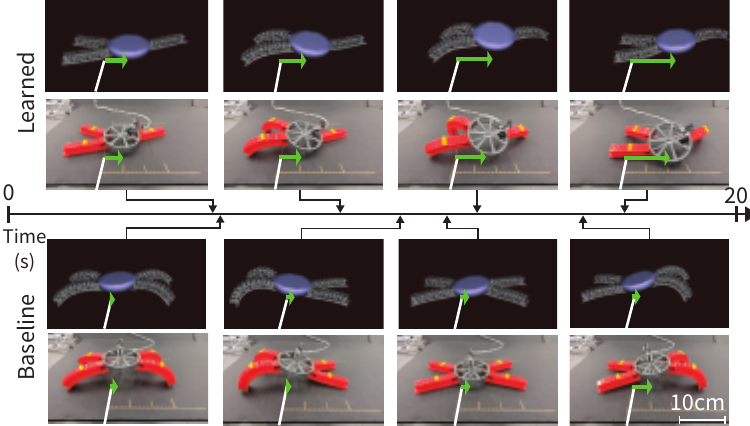}
    \caption{Our framework jointly learns the design and control of crawling soft robots (top) that outperform an expert-designed baseline (bottom). While trained exclusively in simulation, our learned robots are capable of zero-shot sim-to-real transfer, with the optimal design moving more than $2 \times$ faster than the baseline in the real world.}\label{fig:frames}
\end{figure}
While our approach is general, we focus our study around the easily manufacturable PneuNet actuator~\cite{mosadegh2014pneumatic}, which has previously been used to create robots capable of walking and crawling gaits~\cite{gamus2020understanding, shepherd_multigait_2011}.
We experimentally validate our proposed approach by learning combinations of PneuNets and their controllers that together lead to faster gaits, and demonstrate the ability to successfully transfer optimized design-control pairs to reality (Figure~\ref{fig:frames}).

Our work contributes a complete simulation and optimization framework for the joint design and control of soft robots capable of zero-shot sim-to-real transfer. Specifically, this includes:
\begin{itemize}
    \item a model-free algorithm for optimizing the blended design and control spaces of soft robots;
    \item a framework for creating reconfigurable reduced-order soft robot models that improve computational efficiency and enable the use of learning techniques;
    \item the discovery of pneumatically actuated soft robots that outperform a standard expert-designed crawling robot in simulation and reality.
\end{itemize}
See our webpage\footnote{\href{https://sites.google.com/ttic.edu/evolving-soft-robots}{https://sites.google.com/ttic.edu/evolving-soft-robots}} for code and videos of our results.

\section{Related Work}

The problem of jointly optimizing a rigid robot's physical structure along with its control has a long history in robotics research. Early work employs evolutionary methods to optimize the robot's design along with its (often neural) controller~\cite{lipson00, paul01, murata2007self, bongard11}. Another common approach is to assume access to a parameterized model of the robot's dynamics and to then optimize these parameters together with those of control (or motion)~\cite{paul06, villarreal13, ha17, spielberg17, geilinger18, taylor2019optimal, bravo2020one}. Bolstered by the availability of efficient high-fidelity physical simulators, joint optimization methods based on reinforcement
learning are able to learn capable rigid-body design-controller pairs without prior knowledge of the dynamics~\cite{schaff_jointly_2019, pathak2019learning, ha_reinforcement_2019, whitman2021learning}.

Compared to rigid robotics, jointly optimizing the design and control of soft robots is less explored. Of the work that exists, the large majority focus exclusively on simulation. Many approaches reason over design and control spaces that include a mix of discrete and continuous parameters (e.g.,
voxel-based soft robots (VSRs)~\cite{talamini2019evolutionary} are composed of discrete voxels, but the input frequency to each voxel is considered to be continuous).
\citet{spielberg_co-learning_2021} propose an autoencoder-based method that is able to optimize the placement of a large number of such voxels for simulated locomotion tasks, with fewer iterations than other approaches.
\citet{cheney2014unshackling} use an evolutionary neural strategy to develop designs for VSRs that locomote in simulation.
\citet{kriegman2019automated} describe an approach to deforming the structure of VSRs subject to damage such that the original control policy remains valid. \citet{ma2021diffaqua} use a material point method-based simulation and gradient-based optimization methods to co-optimize the shape and control of simulated swimming robots. \citet{deimel2017automated} use particle filter-based optimization to co-optimize finger angles and the grasp strategy of a soft gripper.
The success of these methods in simulation is encouraging for soft roboticists, and recent simulation-based benchmarks allow for a rigorous comparison of co-optimization methods~\cite{collins2021review,bhatia2021evolution}. However, existing work provides a limited evaluation of the physical design-control pairs, and so little is known about their ability to transfer to the real world. Indeed, experiments on voxel-based soft robots reveal that their behavior in simulation can differ significantly from reality~\cite{kriegman2020scalable}. 

One notable exception, \citet{morzadec2019toward} experimentally verify an optimized soft robotic joint, showing how shape optimized using a finite element analysis-based simulator~\cite{Coevoet17} translates to improvements in a real-world soft robotic leg, however they do not consider optimizing the controller. Another exception is recent work that integrates a pneumatic-based passive controller into the robot's design to achieve a forward walking gait~\cite{drotman_electronics-free_2021}, providing an example of how soft robots can have unclear boundaries between design and control.

Meanwhile, individual design and control methods continue to be key areas of research in soft robotics~\cite{rus2015design}. 
There exist a wide variety of design concepts for soft robots~\cite{chen2020design}
such as fluidically pressurizeable soft devices~\cite{mosadegh2014pneumatic,shepherd_multigait_2011},
metamaterial-based designs~\cite{rafsanjani2018kirigami, lipton2018handedness}, and cable-driven
devices~\cite{bern2020soft}.
Soft roboticists note that existing design optimization methods for compliant nonlinear mechanisms, such as topology optimization, are challenging to use in soft robotics due to complicated soft material behavior~\cite{chen2020design}.
The diversity of the design space for soft robots further exacerbates the challenge of automating the search for optimal designs~\cite{pinskier2021bioinspiration}. 
Model-\cite{bern2019trajectory, bruder2019modeling} and learning-based~\cite{lee2020generalized, Culha-RSS-20, kim2021review} controllers have also proven successful, as have hybrid policy designs~\cite{vitanov2020shape, bern2020soft, howison_reality-assisted_2020}.
\citet{zhu2019learning} consider an origami-like robot with various design configurations that all inform policy optimization, and \citet{morimoto2021model} employ the soft actor-critic algorithm~\cite{haarnoja2018soft} for reaching tasks. Related, \citet{vikas2016design} present a modular approach to designing 3D-printed motor-tendon soft robots that can be readily fabricated, and a model-free algorithm for learning the corresponding control policy. Unlike our framework, however, they do not jointly reason over design and control.

\section{Approach} \label{sec:coopt}

\begin{figure}[!t]
    \centering
    \includegraphics[width=0.8\textwidth]{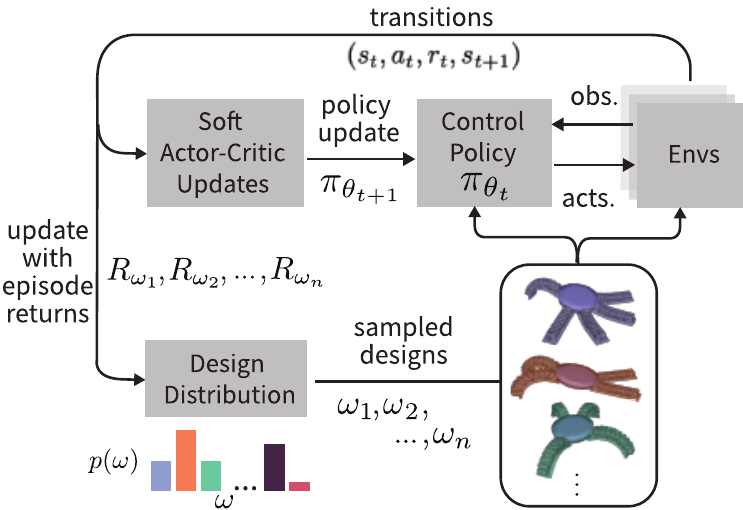}
    \caption{Our approach maintains a distribution over designs $p(\omega)$. At each iteration, the method samples a set of designs $\omega_1, \hdots, \omega_n$ and controls each using a shared, design-conditioned, policy $\pi_\theta$. We train the policy using soft actor-critic on a mixture of data from different designs, and update the design distribution based on the episode returns of the sampled designs.}\label{fig:soft_algorithm}
\end{figure}

The co-optimization problem in this setting is fundamentally the same as defined in Chapter~\ref{sec:design_and_control}.
For design space $\Omega$, we are tasked with solving a set of MDPs, $\{ \mM_\omega | \, \omega \in \Omega \}$, and selecting the design $\omega^*$ and control policy $\pi^*_\omega$ which jointly optimize the expected return for some task. To accomplish this, we extend the algorithm defined in Chapter~\ref{sec:design_and_control} by defining a design distribution and update scheme for finite design spaces.
We employ the same algorithm as in Chapter~\ref{sec:design_and_control} with a few modifications which we will now highlight.

\subsection{Optimizing with Categorical Design Distributions}

\begin{algorithm}[t!]
    \begin{algorithmic}[1]
        \STATE Initialize $\pi_\theta(a \vert s,\omega)$, $p(\omega)$, $T=0$
        \WHILE{$T$ $<$ BUDGET}
          \STATE Sample designs $\omega_1, \omega_2, \hdots, \omega_n \sim p_\phi$ \label{alg:sample}
          \STATE Control $\omega_1, \omega_2, \hdots, \omega_n$ with $\pi_\theta$ for $t$ timesteps. Add transitions to replay buffer. \label{alg:control}
          \STATE Update $\theta$ using soft actor-critic. \label{alg:sac}
          \STATE Update $R_{\omega_1}, R_{\omega_2}, \hdots, R_{\omega_n}$ with their obtained returns.
          \STATE Set timestep $T = T + nt$
          \STATE Set $\beta_T$ to match entropy target $\mathcal{H}_T$.
          \STATE Set $p(\omega) = \frac{e^{\beta_T R_\omega}}{\sum_\Omega e^{\beta_T R_\omega}}$
        \ENDWHILE
    \end{algorithmic}
    \caption{Joint Optimization of Design and Control}
    \label{alg:soft_main}
\end{algorithm}

The design distribution can be modeled and updated in a number of different ways depending on the nature of the design space. In Chapter~\ref{sec:design_and_control}, we modeled the distribution over continuous design parameters using a mixture of Gaussians and updated the parameters of this model using a policy gradient method. 
Here, we focus on discrete and practically enumerable design spaces that can be modelled with a categorical distribution.
In this setting, design performance can be explicitly tracked and gradient-based updates are not required.
Following the principle of maximum entropy, we model the design distribution $p(\omega)$ as a Gibbs distribution:
\begin{equation}
    p(\omega) = \frac{e^{\beta R_\omega}}{\sum_{\omega \in \Omega} e^{\beta R_\omega}},
\end{equation}
where $R_\omega$ is the most recent episode return obtained by design $\omega$, and $\beta$ is an inverse temperature parameter used to control entropy.
At each point in training, we set $\beta$ to maintain a decaying entropy target.
Specifically, we set $\beta = 0$ to specify a uniform distribution during an initial training period, and then decay entropy according to a linear schedule.
This schedule is akin to removing a constant fraction of designs from the search space at each step during training.
See Algorithm~\ref{alg:soft_main} for an overview of this approach.

\subsection{Application to Crawling Soft Robots}

\begin{figure}[!t]
    \centering
    \begin{minipage}{0.38\linewidth}
        \subfigure[Design Space]{\includegraphics[width=\linewidth]{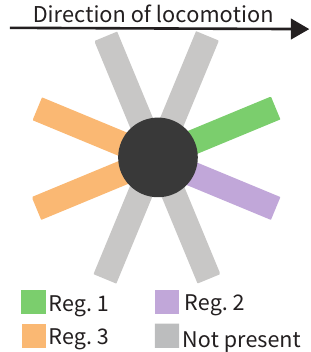}\label{fig:design_space_simple}}
    \end{minipage}\hfil
    \begin{minipage}{0.4\linewidth}
        \subfigure[Simulated Design]{\includegraphics[width=\linewidth]{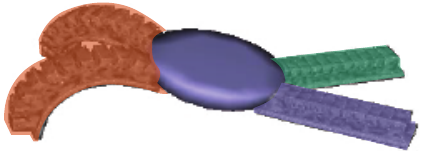}\label{fig:design_space_sim}}\\
        \subfigure[Real-world Design]{\includegraphics[width=\linewidth]{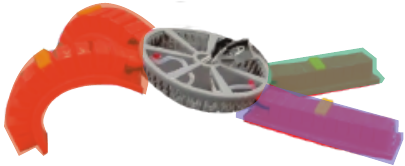}\label{fig:design_space_real}}
    \end{minipage}
    \caption{A visualization of \subref{fig:design_space_simple} our design space that consists of a disk with $N=8$ candidate locations for pneumatic actuators, each of which can be connected to one of $M=3$ pressure regulators. On the right are examples of a \subref{fig:design_space_sim} simulated and \subref{fig:design_space_real} real-world design, where colors identify the pressure regulator for each actuator. The forward direction is to the right.} \label{fig:design_space}
\end{figure}

The design space that we study here (Figure~\ref{fig:design_space}) consists of a disk with $N$ equally spaced locations where soft, pneumatic actuators can be positioned radially outward. Each actuator is connected to one of $M$ different pressure regulators. Designing the robot then amounts to choosing whether (or not) to place an actuator at each of the $N$ locations and, for each placed actuator, connecting it to one of the $M$ pressure regulators. %
Our specific implementation considers
$N=8$ candidate locations and $M=3$ regulators, and restricts the design to having between three and six actuators. This results in a total of $41202$ unique designs, which can be reduced to $6972$ by exploiting symmetry in the regulator assignments. Each actuator is a PneuNet~\cite{mosadegh2014pneumatic} that, like similar soft actuators, has been combined to achieve crawling gaits~\cite{Goury18, shepherd_multigait_2011,gamus2020understanding,vikas2016design}, providing a well-studied baseline.

While our approach is compatible with any RL algorithm, we use the standard soft actor-critic (SAC) algorithm because it offers stable and data-efficient learning dynamics.
We train an open-loop controller modeled as a feed-forward neural network for the task.
This simplification of the controller forces the design to perform ``morphological computation''~\cite{rus2015design} to enable intelligent behavior.
The policy takes as input the design parameters along with the four most-recent actions and outputs pressure targets for each regulator. 

\section{Model Order Reduction for Soft Robots} \label{sec:mor}

Simulators based on the finite-element method (FEM) that are accurate enough for sim-to-real transfer are often too computationally expensive to collect the data required for reinforcement learning.
To obtain a simulation fast enough for reinforcement learning, we perform model-order reduction (MOR).
This section presents our extension of the MOR method proposed by \citet{Goury18} to the problem of co-optimization of design and control that allows for the efficient reuse of reduced models across the space of designs.

\subsection{Reduction through Snapshot POD and Hyperreduction}\label{sec:subsec:mor_overview}

The finite element method provides an approximate numerical solution to partial differential equations (PDEs) by discretizing space into a mesh consisting of a set of finite elements.
Often, dense meshes (and subsequently, large amounts of computation) are needed to reach acceptable accuracy.

The soft actuator mesh contains nodes with position $q_{t_n}$ and velocity $v_{t_n}$ at discrete time step $t_n$. At each $t_n$, simulation requires solving a discrete form of Newton's second law~\cite{Goury18}:
\begin{equation} \label{eq:fem}
    A(q_{t_n}, v_{t_n})dv = b(q_{t_n}, v_{t_n}) + H^\top \lambda,
\end{equation}
where $dv = v_{t_{n+1}} - v_{t_n}$, $A \in \mathbb{R}^{d \times d}$ collects inertial and internal forces, $b \in \mathbb{R}^d$ contains terms from internal and external forces, and $H^T\lambda \in \mathbb{R}^d$ collects constraints (e.g., associated with contact with the floor), with $d$ being the number of degrees-of-freedom in the mesh.
When using dense meshes for accurate simulation, constructing the matrix $A$ and solving this system of equations are often the main bottleneck in FEM simulations.

We first reduce the system dimension through snapshot proper orthogonal decomposition (POD). Using the methods of \citet{Goury18}, we find a low-dimensional subspace $\Phi$ that well-approximates the space of possible motions and deformations while reducing the order through a Galerkin projection onto Equation~\ref{eq:fem}:%
\begin{equation} \label{eq:galerkin}
    \Phi^\top A(q_{t_n}, v_{t_n})\Phi d\alpha = \Phi^\top b(q_{t_n}, v_{t_n}) + \Phi^T H^\top \lambda.
\end{equation}
We achieve this by recording ``snapshots'' of the position $q_t$ of the mesh throughout a series of predefined motions that try to cover the space of common deformations.
For example, one possible set of motions is to iterate through the extremes of each actuator~\cite{Goury18}.
The method then performs a singular value decomposition and uses the left singular vectors, up to some tolerance, as the basis of $\Phi$.
The simulation then uses lower-dimensional coordinates $\alpha$, with $q_{t_n} = q_0 + \Phi\alpha_{t_n}$.

Though snapshot POD reduces the time to solve Equation~\ref{eq:fem}, it still requires computing the high-dimensional matrix $A$ at every time step. We therefore perform a hyperreduction to further approximate $A$ by predicting its entries from the contributions of a small number of elements. We use the hyperreduction method of energy conservation sampling and weighting (ECSW)~\cite{Farhat14}. For further details regarding this two-part method and a demonstration in soft robotics, we refer the reader to \citet{Goury18}.

\subsection{Modularized Reduction for Design-Reconfigurability} 

\begin{figure}[t]
    \centering
    \includegraphics[width=1.0\linewidth]{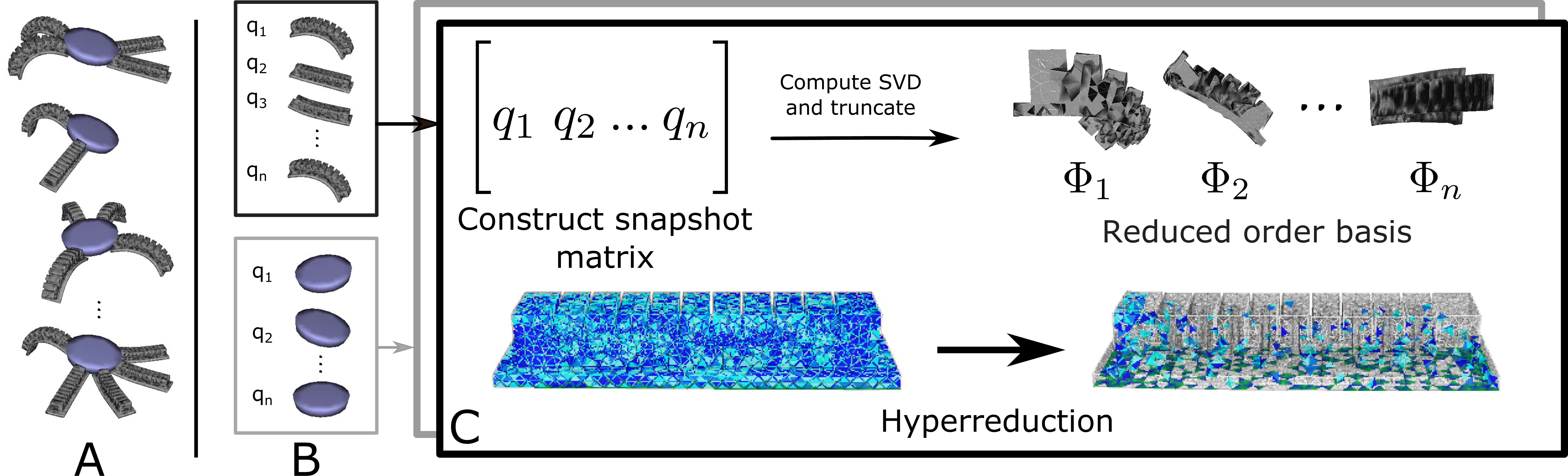}
    \caption{Our proposed technique for model order reduction that is compatible with co-optimization. \textbf{A}: Sample candidate designs and record their animations. \textbf{B}: Collect parts across designs and transform them into a common reference frame. \textbf{C}: For each part, perform a snapshot POD reduction and hyperreduction to obtain a reduced-order basis for motion, and reduced integration domain. We create reductions for new designs by combining reductions of their parts.} \label{fig:mor}
\end{figure}
The MOR method described in the previous section uses a single fixed mesh, and in this section we describe our adaptation for co-optimization (Figure~\ref{fig:mor}).
For high-dimensional soft robot design spaces, separately reducing each design is computationally intractable.
Instead, we define a modularized design space: each design $\omega \in \Omega$ is defined as a combination of a small set of fixed parts $\mathcal{P}$.
We then reduce each part in $\mathcal{P}$ using the method of Section~\ref{sec:subsec:mor_overview} independently and combine the parts in arbitrary ways to form new designs. The number of times we perform MOR is then of the same size as $\mathcal{P}$ rather than the size of $\Omega$.

MOR on the modularized design space only well-approximates the full-order model when the computed subspace $\Phi^p$ for each part $p \in \mathcal{P}$ is close to all frequently achieved deformations.
Because designs will deform in different ways, it is necessary to include `snapshots' of motions from a large set of designs to achieve high-quality reduced-order models.
Therefore, careful snapshot selection on each module in $\mathcal{P}$ is crucial and inaccuracies may be exploited during optimization to result in invalid design-control pairs.
We achieve high-quality reduced-order models for each part by collecting snapshots from a heuristically chosen subset of $\Omega$ and animating those designs by cycling through the pressure extremes of each actuator.

When constructing the reduced basis for new designs, we transform the basis $\Phi_p$ to match the initial pose $(t_i, R_i)$ of each part $p_i$ by rigidly rotating the node positions that make up each basis vector:
\begin{equation}
    \Phi^{p_i}_j = \begin{bmatrix}
        R_i \Phi^p_j[0:3] & R_i \Phi^p_j[3:6] \cdots R_i \Phi^p_j[n-3:n]%
        \end{bmatrix},
\end{equation}
where $\Phi^p_j \in \mathbb{R}^n$ is the $j^\textrm{th}$ column of $\Phi^p$ and $\Phi^p_j[k:l]$ is a slice of that vector from index $k$ to index $l$, inclusive.
We ignore the initial translation $t_i$ because translation basis vectors are included in $\Phi_p$.

\subsection{Reduction of Crawling Soft Robots}

We apply this reduction technique to our design space of crawling soft robots.
Our designs are composed of two parts: the central disk, and some number of identical PneuNets.
Therefore, the above approach allows us to perform two reductions (one for each part) as opposed to reducing each of the $6972$ designs in our design space.
We found that a sparse disk mesh was sufficiently fast and accurate for simulation and we therefore only reduce the PneuNet.

For the reduction of the PneuNet, we select a heuristic set of $256$ designs for which we collect snapshots.
Each design contains a unique subset of the eight potential PneuNet positions, and each PneuNet is controlled independently.
Similar to \citet{Goury18}, we iterate through the extremes of each actuator and record snapshots at fixed time intervals.
This can be seen as a walk through the vertices of an $n$-dimensional hypercube, where $n$ is the number of PneuNets present.
In order to verify the accuracy of our reduction, we evaluate it on a set of four test designs and animations by computing the distance between the node positions of the reduced and unreduced models. 
We perform a grid search over the two tolerances in the reduction algorithm and select the reduction that has the best time-accuracy trade off.

\section{Experiments} \label{sec:expt}

We test our approach by attempting to find a design and open-loop controller that crawl as far as possible on a flat plane in a $20$ second episode.
We define reward as the distance traveled in the (forward) $x$-direction (in cm) as measured at the center of the disk.

After performing model order reduction, we carry out FEM simulation using the SOFA Finite Element framework~\cite{Faure2012} with the soft robotics~\cite{Coevoet17} and model order reduction~\cite{Goury18} plugins.
We model the PneuNet legs (including the inflatable and constraint material) and central disk as linear elastic materials.
We assign input pressures as pressure constraints on surface meshes internal to the PneuNets.
We estimated the Young's modulus of the PneuNet material (Smooth-On DragonSkin 30) based on the published Shore hardness together with the method of~\citet{qi2003durometer}.
We model the constraint layer of the PneuNet as being linear elastic with a Young's modulus twice the magnitude of the inflatable material.
We tuned the Poisson's ratio in order to maintain numeric convergence and qualitative realism.
The friction model used is Coulomb friction.
To account for any inaccurate or unmodelled effects, we measure deformation of a single real PneuNet under fixed pressures, find the corresponding pressures that results in the same deformation of the simulated PneuNet, and fit an affine function to this data.
Pressures commanded by our learned policies are then mapped through this function to ensure a simulated response similar to that of the real PneuNets.
We find that this step greatly improves sim-to-real transfer.
\begin{figure}[t]
    \centering
    \subfigure[Baseline Design]{\includegraphics[height=1.75in]{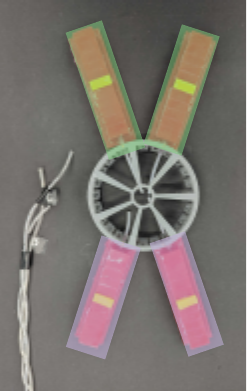}\label{fig:baseline_design}}\hfil
    \subfigure[Baseline Gait]{\includegraphics[width=0.75\linewidth]{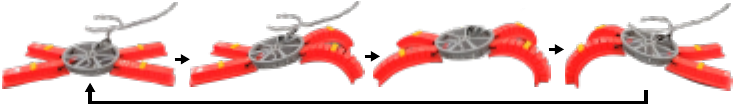}\label{fig:baseline_gait}}\\
    \subfigure[Experimental Setup]{\includegraphics[height=1.75in]{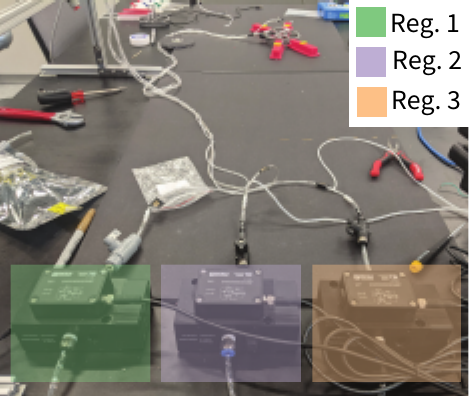}\label{fig:baseline_setup}}\hfil
    \subfigure[Reward vs.\ Phase]{\includegraphics[width=0.6\linewidth]{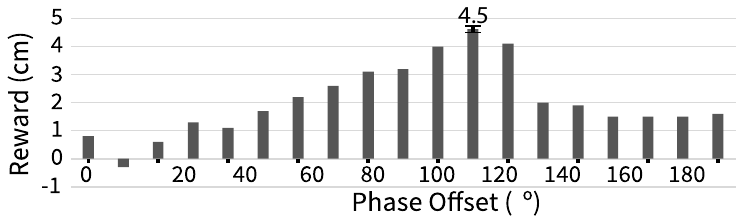}\label{fig:baseline_phase}}\\
    \caption{Our baseline consists of \subref{fig:baseline_design} an expert-designed soft robot with four legs, where the fore and hind legs are attached to pressure regulators one and two, respectively. The experimental setup consists of \subref{fig:baseline_setup} three pressure regulators, a crawling surface, and 3.15\,mm outer-diameter %
    tubing connected to the robot (in the the distance). \subref{fig:baseline_gait} Snapshots of the expert-designed gait. \subref{fig:baseline_phase} Reward (distance traveled in cm) obtained by an offset-sine gait with different phase shifts. A phase difference of $110^{\circ}$ achieves the highest reward.}
    \label{fig:gait_tuning}
\end{figure}

We built an experimental platform to evaluate the performance of our learned soft robot designs. The platform consists of a pressure chamber at 400$\pm100$ kPa in series with three pressure regulators.
Each pressure regulator is connected by a lightweight tube to the robot and is controlled by a programmable power supply. %
We created a modular assembly scheme in which any robot from the design space can be built. We 3D-printed a lightweight polymer disk that was then attached to molded, soft PneuNet legs~\cite{mosadegh2014pneumatic} fabricated from Smooth-On DragonSkin 30.

We designed a baseline design-controller pair similar to the robot used by \citet{shepherd_multigait_2011}.
The baseline (Fig.~\ref{fig:gait_tuning}) has two fore legs and two hind legs placed $45^\circ$ apart with each pair jointly controlled by a separate regulator.
Based on recent analysis of inching gaits~\cite{gamus2020understanding}, we constrain each pressure regulator to produce a sine wave of equal amplitude and period.
We achieve forward motion by imposing a phase shift between the sine waves for the fore and hind legs.
We select a pressure range of $0$\,kPa to $90$\,kPa to avoid both physical instabilities (i.e., aneurysms of the PneuNets) and numerical instabilities in the FE simulator.
We use the maximum amplitude allowed in this range of $45$\,kPa and choose a period of $4$ seconds, which is the fastest period that led to stable motion.
The optimal phase shift depends heavily on friction~\cite{gamus2020understanding} so we conducted experiments with different phase shifts between $0^{\circ}$ and $180^{\circ}$ in increments of $10^{\circ}$, and chose the value that resulted in the highest reward. Figure~\ref{fig:gait_tuning} shows the effect of phase shift on the reward.

We use $96$ parallel environments for data collection.
Each environment contains a design sampled from the design distribution $p(\omega)$ (Algorithm~\ref{alg:soft_main}, line~\ref{alg:sample}) that is controlled with the current policy for one episode (Algorithm~\ref{alg:soft_main}, line~\ref{alg:control}).
The control policy is then updated using the soft actor-critic algorithm on data from a replay buffer (Algorithm~\ref{alg:soft_main}, line~\ref{alg:sac}).
We repeat this process for $1$M environment timesteps during which we fix the design distribution to be uniform for the first $200$K timesteps, after which we linearly decay the entropy to zero at $1$M timesteps.
After training, we manufacture several of the highest-performing designs and conduct a series of experiments to evaluate the sim-to-real transfer of the learned design-control pairs.

\section{Results}\label{sec:soft_results}

\begin{figure}[!t]
    \centering
    \includegraphics[width=0.85\linewidth]{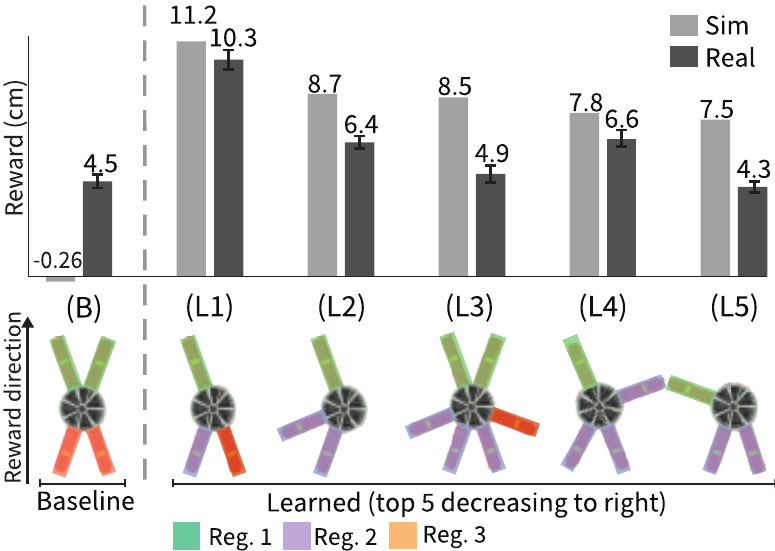}
    \caption{A comparison of the reward (distance traveled in cm) achieved in simulation and reality for the duration of the $20$\,sec episode for the baseline robot (B) and the top-five learned design-control pairs (L1--L5).} \label{fig:comparison_reward}
\end{figure}
We examined the performance of the top-five learned design-control pairs that our framework discovers, our baseline design-control pair, and their capacity for sim-to-real transfer.
We refer to the baseline as ``B'' and the learned pairs as ``L1--L5'' in order of decreasing reward.
Figure~\ref{fig:comparison_reward} visualizes these designs and compares their reward to that of the baseline in simulation and in the real world (tested over five trials).
The top four learned robots (L1--L4) outperform the baseline in both simulation and the real world, while the fifth robot (L5) performs comparably to the baseline in the physical experiments. The top learned robot, L1 outperforms the baseline (in real measurements) by a factor of $2.3$.
\begin{figure}[!th]
    \centering
    \includegraphics[width=0.9\linewidth]{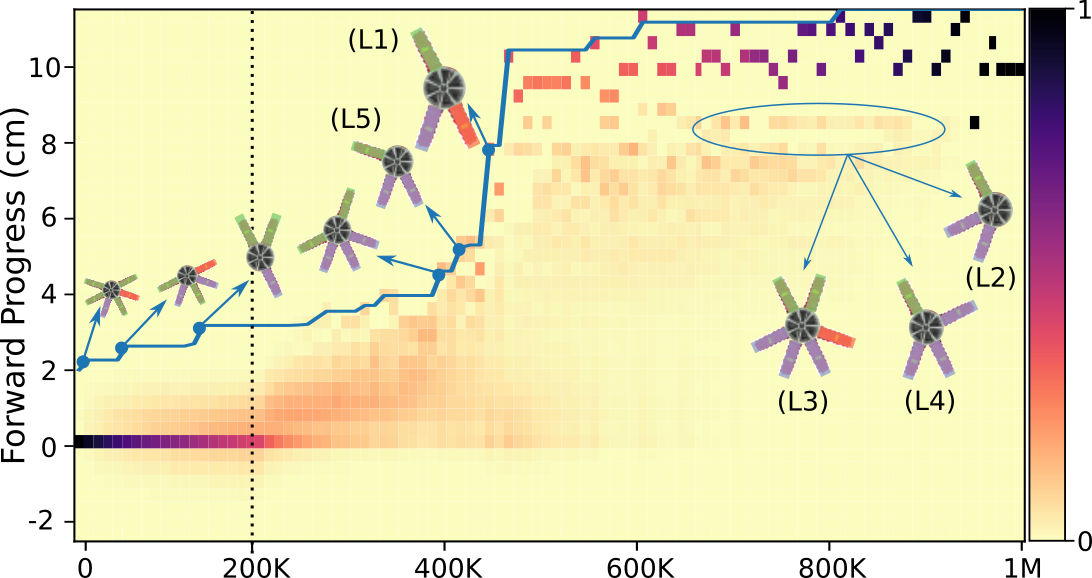}
    \caption{A histogram of rewards throughout training, weighted by the design distribution. The blue line represents the highest reward achieved at each point throughout training.} \label{fig:reward_hist}
\end{figure}
Figure~\ref{fig:reward_hist} inspects the training dynamics of our approach; it shows a histogram of rewards achieved by the design distribution throughout training.
In the beginning of training, the design distribution is constrained to be uniform and nearly every design achieves zero reward.
Starting at $200$K timesteps, the algorithm constrains the distribution with a linearly decaying entropy, after which the algorithm specializes to high performing designs (i.e., L5).
Approximately halfway through training, the algorithm converges on design L1, which achieves a reward that is several centimeters better than the next-best design-control pair.

\subsection{Qualitative Analysis of Learned Designs and Gaits}

\begin{figure}[!th]
    \centering
    \subfigure[Baseline Policy]{\includegraphics[width=0.51\linewidth]{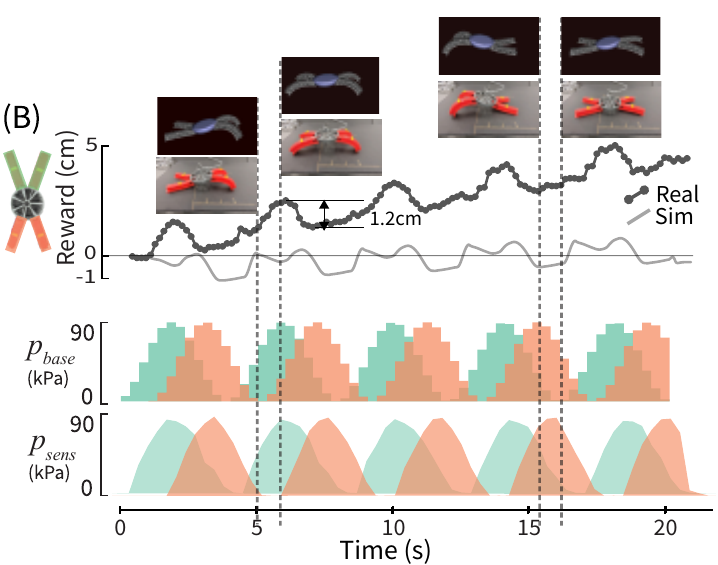}\label{fig:comparison_policy_base}}%
    \subfigure[Learned Policy with Highest Reward]{\includegraphics[width=0.51\linewidth]{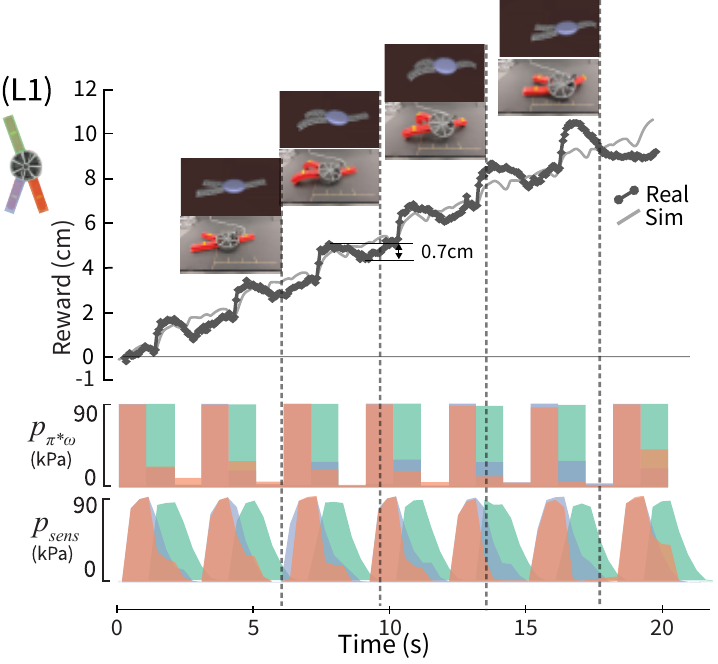}\label{fig:comparison_policy_learned}}
    \caption{A comparison between the \subref{fig:comparison_policy_base} baseline (B) and \subref{fig:comparison_policy_learned} highest-performing learned robot (L1) in terms of the reward (distance traveled in cm) achieved in simulation and reality for the duration of the $20$\,sec episode, along with the corresponding control policy in terms of commanded and sensed pressures.}\label{fig:comparison_policy}
\end{figure}
Observing the learned design-control pairs in simulation and through real-world experiments reveals that they exploit changes in frictional forces in clever ways to create forward motion. Figure~\ref{fig:comparison_policy} compares the open-loop (pressure) gaits and reward trajectories (in simulation and the real-world) over the duration of the $20$\,sec episode for the baseline and top learned design-control pairs. The baseline robot (B) uses a symmetric design that moves forward through out-of-phase actuation of the front and hind legs (Fig.~\ref{fig:comparison_policy_base}). In contrast, robot L1 uses an asymmetric design alongside out-of-phase actuation of the three attached pressure regulators. The result of this design and motion is a pivoting behavior visible in the photos within Figure~\ref{fig:comparison_policy_learned} and the supplementary video. Due to its asymmetry, the soft robot rolls to its side and the contact area of the front leg with the floor is reduced (leading to reduced frictional forces on that leg). This reduction in contact area enables the front PneuNet `leg' to slip forward instead of pushing backward. The result is that robot L1 uses forward-slipping motion without as much backward displacement per step (a.k.a.\ backsliding) in L1 ($0.7$\,cm backsliding) compared to the baseline ($1.2$\,cm backsliding).

See our webpage\footnote{\href{https://sites.google.com/ttic.edu/evolving-soft-robots}{https://sites.google.com/ttic.edu/evolving-soft-robots}} for side-by-side recordings of the learned designs and gaits in simulation and reality. %

\subsection{Sim-to-Real Transfer}

Given the goal of being able to co-optimize the design and control of physically realizable soft robots, we compare the learned robots and baseline in physical experiments.

We find that all of optimized designs and gaits make consistent forward progress, with four out of five (L1--L4) outperforming the baseline in physical measurements, and one performing comparably (L5) to the baseline.
The highest-performing robot (L1) and robot L4 have strong agreement with simulation; with the standard deviation across trials taken into account, real reward reaches within 1\,cm of the simulated reward.

\begin{figure}[t]%
    \centering
    \subfigure{\includegraphics[width=0.49\textwidth]{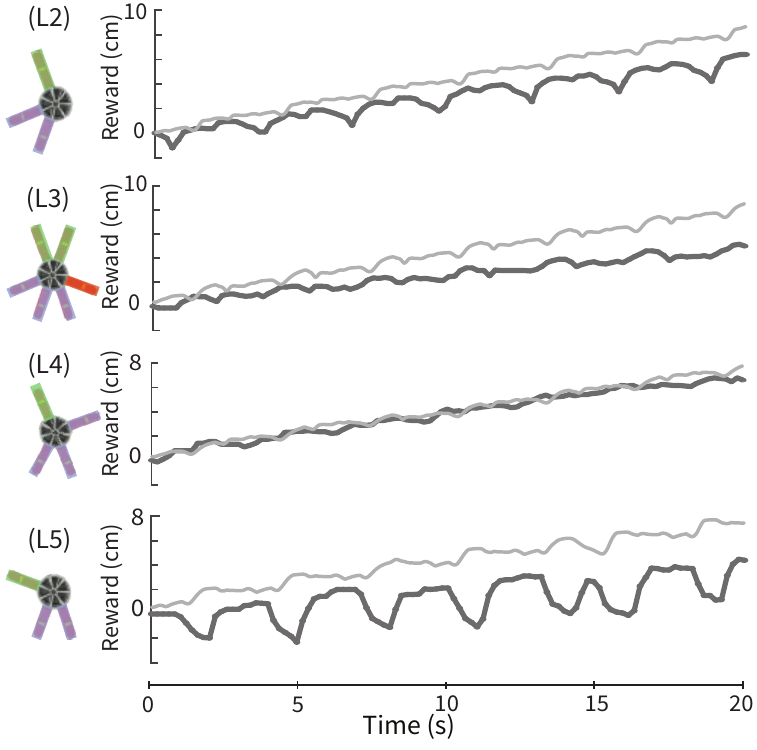}}
    \subfigure{\includegraphics[width=0.49\textwidth]{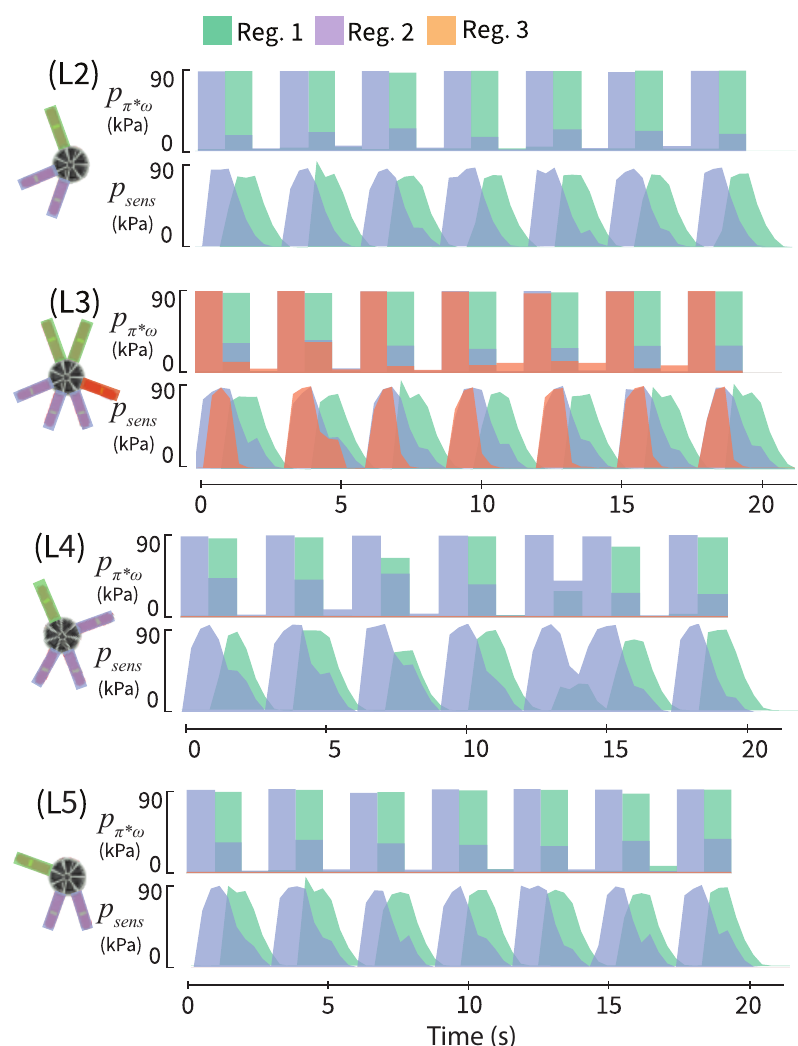}}
    \caption{This figure visualizes (left) the per-timestep rewards for learned designs L2--L5 in both simulation and reality, and (right) the commanded and sensed pressures for each gait.}\label{fig:reward_L25}
\end{figure}

Despite strong qualitative shape agreement seen in the supplementary video, L2, L3, and L5 see a pronounced drop in performance. A comparison of the simulated and real reward over time shows the main reason for this performance decrease.
Figure~\ref{fig:reward_L25} reveals that while these reward functions have qualitatively similar waveforms, the backsliding phases of the waveforms appear to have higher magnitude in reality than in simulation.
The initial phase of each of these gaits inflates the back legs, relying on the friction from the front leg(s) to hold the robot in place rather than slide backwards.
The front PneuNets slip very little in simulation, but they maintain less grip in the real world. The resulting error in the backsliding phase of each of these gaits is then accumulated with every step, resulting in less forward motion.

Figure~\ref{fig:comparison_policy_base} compares the forward progress of the simulated baseline against its real-world counterpart.
Despite detailed tuning of simulation parameters, our baseline gait achieves poor reward in simulation.
The simulated baseline gait does indeed make slow forward progress. Yet, in key segments of the gait (shown by images in Fig.~\ref{fig:comparison_policy_base}), the simulation records less progress, or backward progress, compared to the real-life measurement.
As shown by prior analysis of crawling robots~\cite{gamus2020understanding,vikas2016design}, this type of sinusoidal gait is very sensitive to frictional forces.
Because the design is symmetric, the gait relies on the subtle differences in friction to enable timing of stick-slip interactions for forward motion.
As a result, error accrual due to backward sliding that was already evident in the learned designs is of even higher magnitude here.

While there are many other differences between simulation and reality, such as dynamics associated with the pressure regulators and damping of high frequency motions, we find that the modelling errors associated with stick-slip transitions the largest effect on the transfer performance of our designs.
Stick-slip transitions are notoriously difficult to model~\cite{luo15}, and often requires smoothness approximations for numeric stability. Overall we still see mostly successful transfer for all five optimized designs.

\section{Conclusion}

This work describes a complete framework for the simulation and co-optimization of the design and control of soft robots capable of zero-shot sim-to-real transfer. We present an algorithm for co-optimization and a framework to create reconfigurable, reduced-order models for soft robotics. Experiments demonstrate that our framework learns design-gait pairs that outperform an expert-designed baseline in a soft robot locomotion task. We further characterize the successful qualitative and quantitative transfer of these learned pairs from simulation to reality.

Soft robots are compliant-bodied and mechanically intelligent. As a result, their design and control spaces are not well-separated.
For many tasks it difficult or impossible to explore these design/control spaces through prototypes or analysis alone; simulation is needed to build and evaluate designs and controllers in a tractable manner.
Our work takes a step in this direction by showing that it is possible to combine reinforcement learning techniques with finite element simulation to deliver fast and physically accurate co-optimization for soft robotics.

This work has some limitations.
Modelling errors, due to friction, linear elasticity, etc., caused us to linearly warp simulation parameters to obtain numerical stability and realism.
Even so, these errors led to a degradation of transfer performance for some designs.
Future work will include improved simulation techniques as well as investigations into domain adaptation and domain randomization methods as a means of improving sim-to-real transfer.
Given the success of this method for simple locomotion, we plan to explore adaptations of this framework to different design and control spaces as well as more complex tasks such as manipulation.%

\clearpage
\section{Additional Information}
This section provides additional details about simulation, model order reduction, training, experiments, and hardware.

\begin{table}[h!]
    \centering
    \begin{tabularx}{\linewidth}{Xr}
        \toprule
        Hyperparameter  &  Value \\
        \midrule
        Number of environments & $96$ \\
        Maximum timesteps & $1$M \\
        Buffer size & $100$K \\
        Batch size & $512$ \\
        Discount factor ($\gamma$) & $0.95$\\
        Policy learning rate & $0.0006$ \\
        Q-function learning rate & $0.0020$ \\
        Optimizer & $\textrm{ADAM}(\beta_1 = 0.9, \beta_2 = 0.999)$ \\
        Target network smoothing coefficient & $0.01$ \\
        Learning start & $10$K \\
        Timesteps per SAC update & $1$ \\
        Timesteps per sampled design & $20$ (one episode) \\
        Entropy linear decay start & $200$K \\
        Entropy linear decay end & $950$K \\
        Entropy target start & uniform \\
        Entropy target end & $0$ \\
        \bottomrule
    \end{tabularx}
    \caption{Soft Actor Critic and Co-optimization Parameters}
    \label{tab:sac_param}
\end{table}

\begin{table}[h!]
    \centering
    \begin{tabularx}{\linewidth}{Xr}
        \toprule
        Parameter & Value \\
        \midrule
        Elastic model & linear \\
        Young's modulus for PneuNet silicone & $1160$\,kPa\\
        Poisson ratio for PneuNet silicon & $0.2$ \\
        Young's modulus for PneuNet paper & $2320$\,kPa\\
        Poisson ratio for PneuNet paper & $0.49$ \\
        Young's modulus for disk & $5000$\,kPa \\
        Poisson ratio for disk & $0.3$ \\
        PneuNet mass & $105$\,g \\
        Disk mass & $75$ g \\
        Friction coefficient & $1.2$ \\
        Gravity & $9800$\,mN\\
        \bottomrule
    \end{tabularx}
    \caption{SOFA simulation parameters}
    \label{tab:sim_param}
\end{table}

\subsection{Learning and Simulation Parameters} \label{app:params}

Table~\ref{tab:sac_param} lists the hyperparameter settings that we used for co-optimization.
We represent each design as a concatenation of eight four-dimensional one-hot vectors indicating whether each PneuNet is present and which regulator it is connected to.
Our policy is open-loop and receives the last four pressure actions as input.
The actions are concatenated with the design representation to create a $44$-dimensional input to the policy and value networks.
Both the policy and value networks are four-layer feed-forward networks and each hidden layer has $[400, 200, 100]$ units, respectively.
We train on a $32$-core AMD EPYC $7502$, with the experiment taking eight days to complete.

\begin{wrapfigure}[15]{r}{0.5\textwidth}
    \centering
    \includegraphics[width=0.45\textwidth]{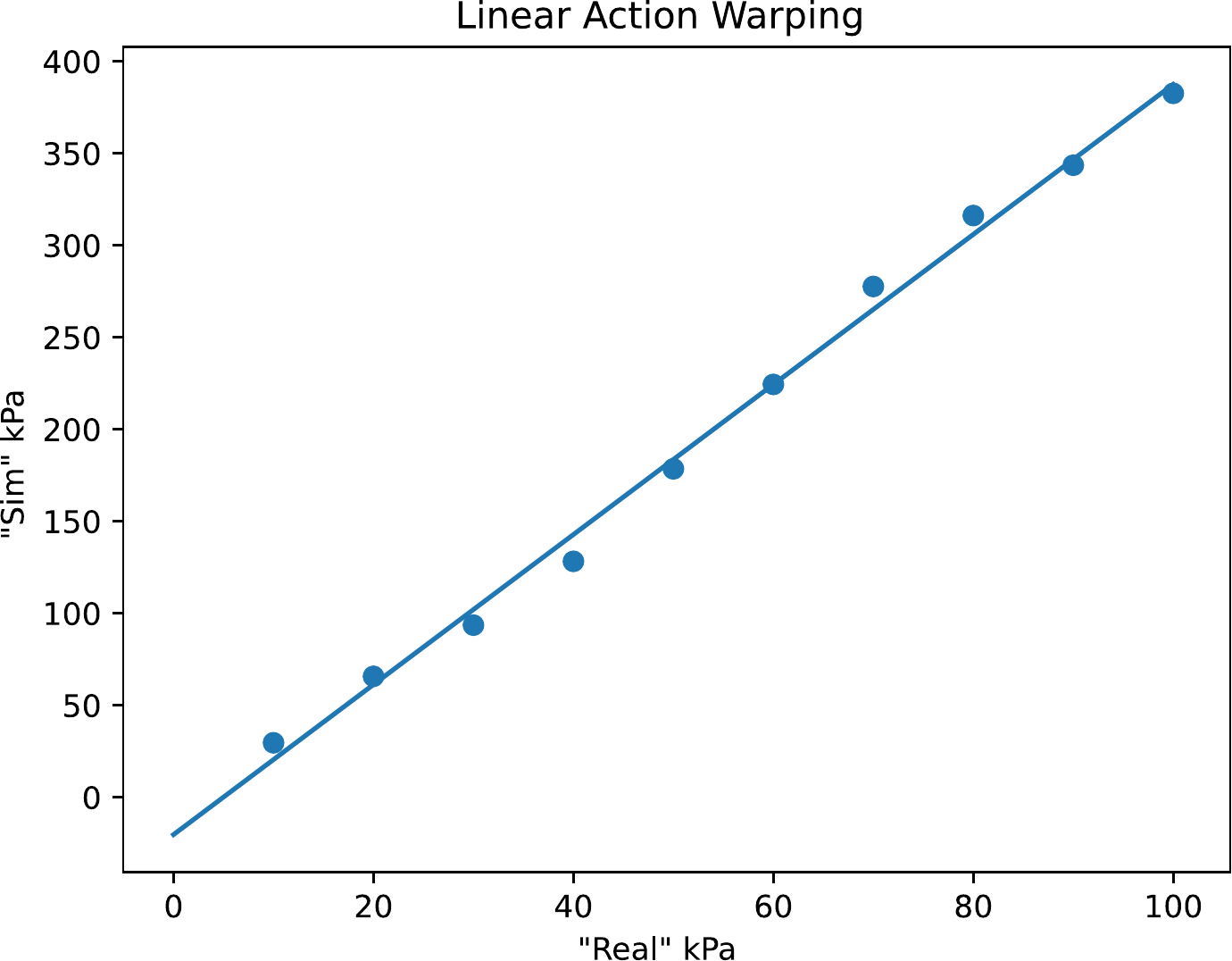}
    \caption{The linear pressure scaling used to align the behavior of a single PneuNet in simulation and reality. In simulation, we scale the pressures output by the policy prior to applying we apply them. This helps to reduce discrepancies between simulation and reality caused by modelling error.}
    \label{fig:warping}
\end{wrapfigure}

Table~\ref{tab:sim_param} provides our simulation parameters.
We use a linear elastic model and set Young's modulus for the silicone using known material parameters.
The Young's modulus for the paper and disk were chosen to be significantly stiffer that the silicone.
The Poisson ratios were set to maintain realism while avoiding numerical instabilities.
The masses were measured from their real counterparts but then scaled by a factor of $2.5$ to avoid numerical instability.
The friction coefficient was chosen to qualitatively approximate the stick-slip behavior of the baseline design when all the PneuNets are inflated and deflated simultaneously.

Figure \ref{fig:warping} shows the linear pressure scaling used to align the PneuNet bending response between simulation and reality.
To calibrate the action scaling, we record the bending of a single PneuNet under pressures from $10$\,kPa to $100$\,kPa, in increments of $10$\,kPa.
We then search for the pressures in simulation which achieve an equivalent bend and fit a linear function to the results.

\subsection{Model Order Reduction Evaluation} \label{app:mor}

\begin{figure}[!t]
    \centering
    \includegraphics[width=\textwidth]{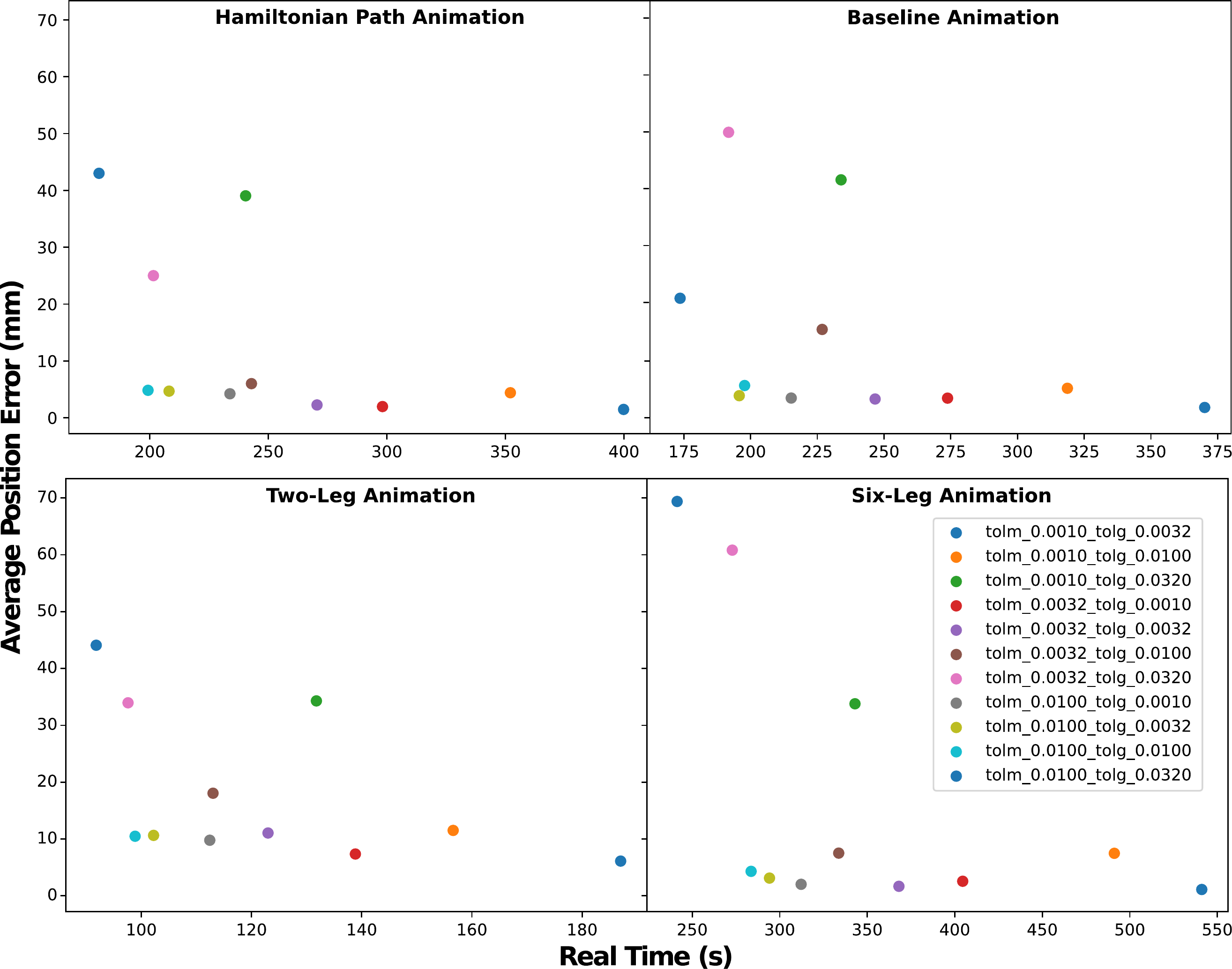}
    \caption{This plot shows the trade off between positional errors introduced by the reduction and the time to simulate each animation for a $3$x$3$ grid search over the $2$ MOR tolerances. We choose a mode tolerance of $0.0032$ and a hyperreduction tolerance of $0.0010$ (red dot) for our experiments.}
    \label{fig:mor_errors}
\end{figure}

The snapshot-POD reduction method has two tolerances that must be set, one for the error induced by the basis $\Phi$, and one for the error induced by the hyperreduction. To explore the effect of these two parameters and to verify the validity of our reduced models, create a test set of four designs and animations:
\begin{enumerate}
    \item the \textbf{Hamiltonian animation} uses the baseline four-legged design and iterates through the pressure extremes of the four PneuNets.
    \item the \textbf{baseline animation} also uses the baseline design and a gait similar to the baseline gait from our experiments.
    \item the \textbf{two leg animation} uses a design with two alternately inflating PneuNets at a $90^\circ$ angle.
    \item the \textbf{six leg animation} uses a design with six PneuNets that alternately inflate in two groups of three.
\end{enumerate}
For each design and animation, we compare the node positions of the mesh using the unredeced and reduced-order models.
We perform a coarse grid search over the two MOR tolerances and plot the simulation speed and accuracy of each reduction.
Errors are computed as the L2 distance between the node positions of the reduced and unreduced model averaged over each node in the mesh and each timestep.
Based on this analysis, we use a mode tolerance of $0.0032$ and a GIE tolerance of $0.001$ for our experiments.

\subsection{Hardware and Fabrication Details} \label{app:robotbuild}

\begin{figure}[!th]
    \centering
    \includegraphics[width=1.0\textwidth]{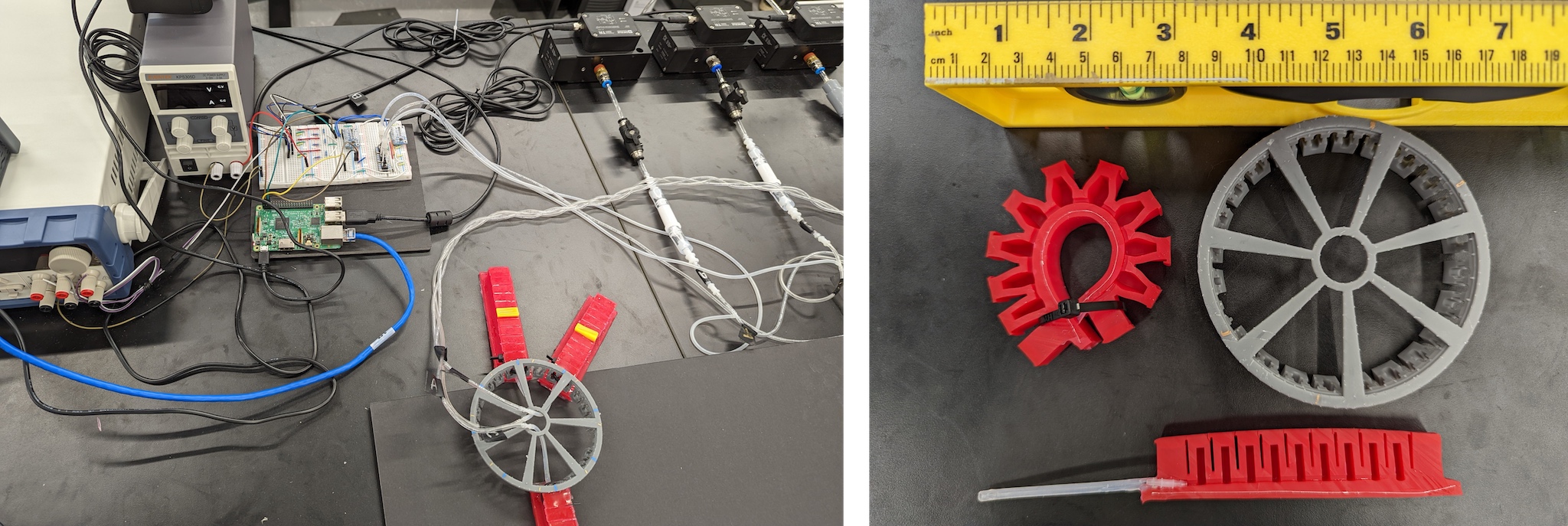}
    \caption{\textbf{Left}: A picture of our hardware set up consisting of a Raspberry Pi, three pressure regulators, two power supplies, three pressure sensor, and a breadboard connecting everything to the Pi. \textbf{Right}: A picture of our $3$D-printed disk and a molded PneuNet cut in half to display the internal structure.}
    \label{fig:robotbuild}
\end{figure}
Figure \ref{fig:robotbuild} displays our experimental platform and robot design.
We keep a pressure chamber at 400$\pm100$\,kPa attached in series with three pressure regulators (Enfield TR-010-gs).
The learned and baseline policies are executed on a Raspberry Pi where the pressure commands are converted to voltage commands and sent to a programmable power supply (BK Precision 9129B).
The power supply then sends voltages to each of the three pressure regulators through a breadboard that converts the voltages into an acceptable range.
Each pressure regulator is connected by a lightweight tube to the robot.
We connect an external pressure sensor to each regulator to verify that the correct pressure is applied.

We created a modular assembly scheme in which any robot from the design space can be built.
We 3D-printed a lightweight polymer disk (Fig.~\ref{fig:robotbuild} (right)) with uniformly distributed locations  where we can attach the PneuNet actuators while routing the pneumatic cables away from the robot (Fig.~\ref{fig:robotbuild} (left)). We 3D-printed moulds for the PneuNets based on a modification of the design given in the Soft Robotics Toolkit~\cite{holland2014soft} such that the two end prismatic segments are filled rather than hollow (Fig.~\ref{fig:robotbuild} (right)), which allows the first prismatic segment to be used as an attachment point to the disk. We fabricated the PneuNet actuators using Smooth-On DragonSkin 30 silicone. %

\flushcolsend
\clearpage 
\chapter{Conclusion}

The core research problem of this thesis is the discovery of effective and automatic strategies for optimizing embodied systems that sense or interact with the world for targeted applications.
To that end, this thesis discusses our efforts to develop efficient and general algorithms for this problem. 
In Chapter~\ref{sec:beacons}, we showed how to co-optimize sensor design and inference algorithms in the context of beacon-based localization systems.
Our approach leverages a differentiable signal propagation model and inference model to optimize beacon placement and channel allocation together with the parameters of the location prediction algorithm in an end-to-end manner for localization accuracy.
In Chapter~\ref{sec:design_and_control}, we develop an efficient reinforcement learning algorithm for co-optimizing the physical design and control of a robot.
Optimizing design and control induces a bi-level optimization problem in which the evaluation of a specific design requires solving for the optimal controls for that robot.
The key innovation of our approach is to avoid this costly search by training a universal controller that is able to quickly provide an evaluation for new designs.
In Chapter~\ref{sec:nlimb_paper}, we extend the ideas of Chapter~\ref{sec:design_and_control} to enable the optimization over robot morphologies. This introduces several challenges such as requiring a search over discrete physical parameters and handling state and action spaces that vary across morphology. Our approach uses graph-based grammars to procedurally generate designs, a novel autoregressive model to model the design distribution, and a novel transformer-based actor-critic architecture to model the universal controller.
In Chapter~\ref{sec:soft_design}, we explore the fabrication and transfer of optimized designs and their controllers.
We extend the ideas in the previous chapters to allow for the optimization of pneumatically actuated soft robots for a crawling task.
In our experiments we fabricate several optimized robots and show large performance improvements over a strong baseline in the real world.

\section{Future Work}

There are several avenues for future work including fabrication of optimized designs, setting up benchmark problems, and incorporating sensor design an perception into the optimization process.

\paragraph{Fabrication of optimized designs:}
One important direction is to further consider the fabrication of optimized designs. Chapter~\ref{sec:soft_design} shows how fabricate and transfer soft robots for a crawling task, but little work explores this for more complicated tasks/systems or rigid robots. The ultimate goal of this research is to automate the design of robots, and therefore it is necessary to carefully incorporate fabrication constraints into the algorithms and consider the misalignment between simulation and reality. Chapter~\ref{sec:nlimb_paper} describes how fabrication constraints could be incorporated through the graph grammar, but this---and the consideration of the sim-to-real gap---remains untested.

\paragraph{Benchmarks for standardizing co-optimization problems:}
Another direction for future work is the compilation of benchmark co-optimization problems to accelerate current research. Recently, there have been a plethora of interesting algorithms proposed~\cite{schaff_jointly_2019, ha_reinforcement_2019, luck_data-efficient_2019, chen_hardware_2020, yuan2021transform2act, spielberg2019learning, pathak2019learning, wang2019neural, zhao2020robogrammar}. 
However, the complexity of these algorithms and the use of different simulators, tasks, and metrics has made the comparison of these approaches challenging.
Work on future algorithmic improvements to co-optimization could greatly benefit from a shared set of tasks and metrics against which to show concrete gains. 

\paragraph{Optimizing sensor design simultaneously with control:}
While the approaches outlined in this thesis provide a general way to approach the optimization of sensors and inference, or physical components and control, our work focused on these problems independently.
However, the original work by~\citet{sims1994} showed that it is possible to simultaneously optimize robot morphology, sensor placement, and control for a variety of tasks.
Future work should seek to develop this capability within the context of learning-based methods.
One problem instance in which this is immediately useful---and has already seen some work~\cite{spielberg_co-learning_2021}---is in soft robotics, where the pose estimation of deformable bodies is a highly non-trivial problem.
In this scenario, morphology, sensor placement, pose inference, and control could be optimized end-to-end with regards to task performance. 

\printbibliography
\flushcolsend

\end{document}